\documentclass[letterpaper, 10 pt, conference]{ieeeconf}  %

\IEEEoverridecommandlockouts                              %

\usepackage{graphicx}
\usepackage{amsmath}
\usepackage{amssymb}
\usepackage{booktabs}
\usepackage{float}    %

\usepackage[dvipsnames]{xcolor,colortbl}
\usepackage{threeparttable}
\usepackage{tabularx}
\usepackage{array,multirow}
\usepackage{threeparttable}

\usepackage{textcomp}
\usepackage{bm}
\usepackage{mathrsfs}
\usepackage{times}
\usepackage{epsfig}
\usepackage{soul}

\usepackage[normalem]{ulem}
\usepackage{outlines}
\usepackage{ifthen}
\usepackage{pifont}
\usepackage[noadjust]{cite}
\usepackage{subcaption}

\usepackage{booktabs}
\def\ourmodel{\textsc{QuAD}}
\newcommand*{\ua}{$\uparrow$}
\newcommand*{\da}{$\downarrow$}

\usepackage[pagebackref=true,breaklinks=true,letterpaper=true,colorlinks,bookmarks=false]{hyperref}

\usepackage[capitalize]{cleveref}
\crefname{section}{Sec.}{Secs.}
\Crefname{section}{Section}{Sections} \Crefname{table}{Table}{Tables}
\crefname{table}{Tab.}{Tabs.}

\newcommand{\argmin}[1]{\underset{#1}{\operatorname{arg}\,\operatorname{min}}\;}
\newcommand{\real}[1]{\mathbb{R}^{#1}}

\usepackage[capitalize]{cleveref}
\crefname{section}{Sec.}{Secs.}
\Crefname{section}{Section}{Sections}
\Crefname{table}{Table}{Tables}
\crefname{table}{Tab.}{Tabs.}

\newcolumntype{s}{>{\raggedleft\arraybackslash}X}
\makeatletter
\renewcommand\paragraph{%
  \@startsection{paragraph}%
                {5}%
                {\z@}%
                {1ex \@plus1ex \@minus.2ex}%
                {0ex}%
                {\normalfont\normalsize\bfseries}}%
\makeatother

\newcolumntype{s}{>{\raggedleft\arraybackslash}X}

\title{\LARGE \bf
QuAD: Query-based Interpretable Neural Motion Planning for Autonomous Driving
}

\author{
\textbf{Sourav Biswas\thanks{\hrule \vspace{0.5em}$^{*}$Denotes equal contribution}$^{*, 1, 2}$, Sergio Casas$^{*, 1, 2}$, Quinlan Sykora$^{1, 2}$, Ben Agro$^{1, 2}$, Abbas Sadat$^{1}$, Raquel Urtasun$^{1, 2}$}\\
$^{1}$Waabi, $^{2}$University of Toronto \\
\texttt {\{sbiswas, sergio, qsykora, bagro, asadat, urtasun\}@waabi.ai}
}

\begin{document}

\maketitle
\thispagestyle{empty}
\pagestyle{empty}

\begin{abstract}

    A self-driving vehicle must understand its environment to determine the appropriate action.
    Traditional autonomy systems rely on object detection to find the agents in the scene.
    However, object detection assumes a discrete set of objects and loses information about uncertainty, so any errors compound when predicting the future behavior of those agents.
    Alternatively, dense occupancy grid maps have been utilized to understand free-space. However, predicting a grid for the entire scene is wasteful since only certain spatio-temporal regions are reachable and relevant to the self-driving vehicle.
    We present a unified, interpretable, and efficient autonomy framework that moves away from cascading modules that first perceive, then predict, and finally plan.
    Instead, we shift the paradigm to have the planner query occupancy at relevant spatio-temporal points, restricting the computation to those regions of interest.
    Exploiting this representation, we evaluate a candidate trajectory around key factors such as collision avoidance, comfort, and progress for safety
    and interpretability.
    Our approach achieves better highway driving quality than the state-of-the-art on high-fidelity closed-loop simulations.
    
\end{abstract}

\section{Introduction}

Self-driving vehicles (SDVs) strive to reach their destinations safely and comfortably by analyzing their surroundings, envisioning potential future scenarios, and using this information to determine a plan of action to carry out.
This is repeated with every new observation.

The majority of autonomy frameworks are \textit{object-based} \cite{fan2018baidu, zeng2020dsdnet, cui2021lookout, renz2022plant}, which implies detecting a discrete set of objects, typically obtained by thresholding output confidence scores from an object detector, 
predicting a small set of hypothetical future trajectories, and finally planning a safe trajectory.
However, this approach loses information about the scene from thresholding and has limited representation of future object uncertainty.
Furthermore, the expressivity of the future trajectory forecasts is limited, as keeping the number of hypotheses low is crucial for real-time inference.

Alternatively, \textit{sensor-to-plan} imitation learning frameworks avoid reasoning about individual objects 
by learning to map sensor data directly to plans \cite{codevilla2018end, zeng2019end, philion2020lift, hu2022model}. 
These learned policies are typically brittle to distributional shift since supervision is only provided in states visited by the expert during training, so the policy never learns to recover from its own mistakes \cite{ross2011reduction}.
Moreover, the decisions are not easy to interpret or explain, which is important for system validation and verification.

To tackle these limitations, \textit{occupancy-based} approaches \cite{hoermann2018dynamic, sadat2020perceive, casas2021mp3, mahjourian2022occupancy}  have proposed a more interpretable object-free paradigm. 
Occupancy describes the probability a point in space and time is occupied by any traffic participant.
This enables planners to decide on consistent and effective actions with respect to this interpretable representation, serving as an explanation of their decision.
This paradigm is also more robust to shifts in the ego state distribution due to the supervision of the intermediate representations, making it better suited for closed-loop deployment.
However, representing the occupancy in dense spatio-temporal grids over a large region of interest is resource-intensive since a high resolution is needed to attain accurate plans. %

\begin{figure}[t]
    \begin{center}
            \includegraphics[width=\columnwidth]{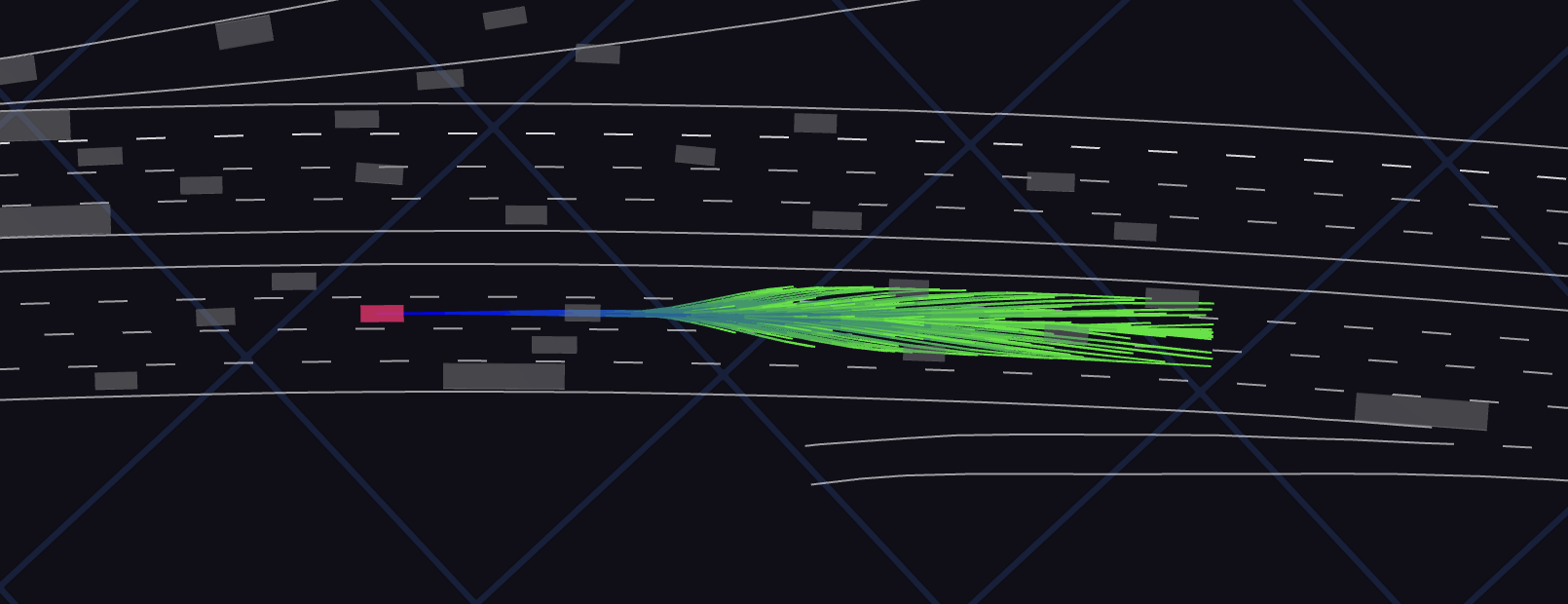}
    \end{center}
    \caption{
    The set of potential plans for the ego vehicle, with time into the future colored from blue to green.
    Our neural motion planner, \ourmodel{}, builds upon two observations: (1) the plans' reachable space is much smaller than the full spatio-temporal volume and (2) many ego states throughout the trajectories are in close proximity to each other. 
    }
    \label{fig:teaser}
\end{figure}

We propose \ourmodel{}, an interpretable, effective and efficient neural motion planner.
\ourmodel{} diverges from prior works that first perceive, then predict, and finally plan.
Instead, our unified autonomy first generates candidate trajectories respecting kinematic constraints and traffic rules, and then queries an implicit occupancy model only at spatio-temporal points needed for planning, which is used to rank the safety of the candidates.
Fig.~\ref{fig:teaser} shows an example of the candidate trajectories, which we can see only occupy a small portion of the spatio-temporal volume prior works predict.
Moreover, we note that many candidate trajectories heavily overlap, which motivates us to quantize the spatio-temporal query points to reduce redundant computation.

Through extensive evaluation we show that \ourmodel{} is able to achieve better closed loop performance in a state-of-the-art highway driving simulator while attaining better runtime than competitive baselines.

\begin{figure*}
        \begin{center}
        \includegraphics[width=1.0\textwidth]{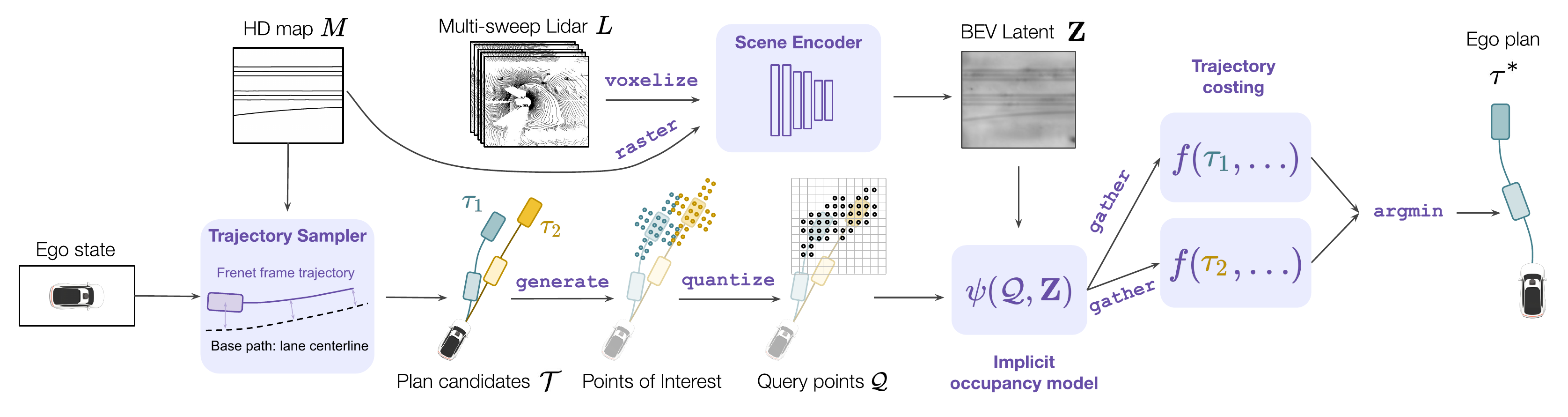}
        \end{center}
        \caption{
        \textbf{\ourmodel{} inference}. Given the ego state and the map, the trajectory sampler generates candidate plans. These plans are converted into query points that cover the relevant areas around the ego vehicle future positions.
        Leveraging multi-sweep LiDAR and HD map, a scene encoder builds a BEV latent representation which we then use to query an implicit occupancy model. Finally, we gather the occupancy relevant to each trajectory, cost them, and select the one with the lowest cost.
        }
        \label{fig:model_diagram}
\end{figure*}

\section{Related Work}

This section reviews prior approaches to end-to-end autonomy from sensor data. 

\paragraph{Object-based autonomy}
Most previous approaches have employed object-based reasoning using a three-stage pipeline: (1) object detection \cite{yang2018pixor, lang2019pointpillars} and tracking \cite{weng20203d, sharma2018beyond}, (2) trajectory prediction based on past tracks \cite{tang2019multiple, phan2019covernet, chai2019multipath, zhao2019multi}, and (3)  motion planning to decide which action the self-driving vehicle should take \cite{fan2018auto,rhinehart2018deep,sadat2019joint, rhinehart2021contingencies}.
This paradigm faces three key challenges \cite{trabelsi2021drowned,sadat2020perceive}: (i) the uncertainty from detection and tracking is not propagated to downstream prediction, (ii) the predicted future distributions tend to be overly simple in practice for computational tractability in scenes with many actors, and (iii) the planner is blind to any objects that fall below the confidence threshold used to determine whether an object exists or not in detection.
Several works \cite{luo2018faf,
casas2018intentnet, Liang_2020_CVPR, casas2020implicit, hu2023_uniad} have addressed (i) by optimizing jointly through multiple stages.
The recent work of PlanT~\cite{renz2022plant} tackles (ii) by learning a Transformer that plans a trajectory for the ego vehicle from object and route tokens coming from the detector and the map, skipping the second stage of trajectory prediction.
However, (iii) is fundamentally difficult to address in this autonomy paradigm as the core assumption is that there is a finite set of objects with a clear boundary between positive and negative instances.

\paragraph{Sensor-to-plan autonomy} 
Autonomy approaches in this family learn to map sensor data directly to plans, without any other intermediate interpretable representations.
The pioneering work of ALVINN~\cite{pomerleau1989alvinn} proposed the use of a single neural network that directly outputs driving control commands. Following the triumph of deep learning, direct control-based methods have made strides through the implementation of more expressive networks, advanced sensors, and scalable learning methods \cite{bojarski2016end,muller2018driving,codevilla2018end,hawke2019urban,hu2022model,zhang2022rethinking}.
To leverage prior knowledge about vehicle kinematics, recent methods such as NMP \cite{zeng2019end} and Lift-Splat-Shoot \cite{philion2020lift} leverage analytical trajectory samplers to propose candidate plans, reducing the learning problem to predicting the cost of the trajectories.
Despite not requiring any manual annotation for training, using only expert demonstrations as supervision tends to result in policies with stability and robustness issues, as they become highly vulnerable to distributional shift \cite{ross2011reduction,codevilla2019exploring,casas2021mp3}.
Note that some methods in this family \cite{zeng2019end,hu2022model} perform auxiliary tasks for additional supervision, but the auxiliary outputs are not used during inference by the planner and therefore no consistency between those outputs and the plans is guaranteed.
Thus, the planner's decisions cannot be explained by those auxiliary outputs.

\paragraph{Occupancy-based autonomy}
These methods predict spatio-temporal occupancy from sensor data, and plan a trajectory that is safe with respect to the predicted occupancy \textit{without considering individual actors}.
The occupancy is then used to assess the risk of a trajectory by measuring the probability of collision \cite{hoermann2018dynamic,sadat2020perceive,casas2021mp3}.
This approach has proven very effective and robust to shifts in state distribution during deployment. 
Different architectures have been proposed to predict occupancy.
P3~\cite{sadat2020perceive}, MP3~\cite{casas2021mp3}, FIERY\cite{hu2021fiery} and OccFlow~\cite{mahjourian2022occupancy} all propose to predict similar variations of 3D spatio-temporal occupancy grids using convolutional neural networks (CNN).
However, this relies on the CNN receptive field being large enough to ``transport'' the occupancy from where the evidence is in the sensor data, to where the object would be at the end of the forecasting horizon. 
Moreover, predicting a dense 3D grid is computationally demanding and unnecessary.
To tackle these two shortcomings, ImplicitO~\cite{agro2023implicito} proposes an implicit occupancy model that can be queried at continuous spatio-temporal points by leveraging deformable attention \cite{zhu2020deformable}, yet it does not propose how to make use of this model for driving.
In this work, we propose a unified autonomy framework which utilizes a state-of-the-art implicit occupancy architecture 
to obtain a strong understanding of the scene 
along with a query point quantization strategy to attain superior plans in practical runtimes without degrading driving quality.

\section{Query-based Autonomous Driving (\ourmodel{})}

The goal of our motion planner (\ourmodel{}) is to find the best trajectory plan to execute given a short history of LiDAR observations, $L$, and a high-definition (HD) map, $M$. 
As shown in \cref{fig:model_diagram}, with every new sensor observation, \ourmodel{} plans a trajectory by optimizing the following objective over a finite set of trajectory candidates $\mathcal{T}$ generated based on the current ego kinematic state and the map:  
\begin{align}
   \tau^* &= \argmin{\tau \in \mathcal{T}} J\big(\tau, \mathbf{Z}, M \big),
\end{align}
where $\mathbf{Z} \in \real{H \times W \times C}$ is a bird's eye view (BEV) latent representation of the environment extracted from a voxelized multi-sweep LiDAR and a raster map by a learned scene encoder.

We define our objective function $J$ as a linear combination of $N$ explainable costs $f_i$, weighted by learnable coefficients $w_i$
\begin{align}
   J\big(\tau, \mathbf{Z}, M \big)  &= f\big(\tau, \psi(\mathcal{Q}_\tau, \mathbf{Z}), M \big) \\   &= \sum_i^N w_i f_i\big(\tau, \psi(\mathcal{Q}^{(i)}_\tau, \mathbf{Z}), M \big).
\end{align}
The learned occupancy model $\psi$ provides flexibility, expressivity and interpretability in our planner. It models occupancy \emph{implicitly} and can be queried for occupancy at any set of continuous spatio-temporal locations $\mathcal{Q}$.
More precisely, $\psi(\mathbf{q}, \mathbf{Z})$ outputs the probability that the continuous spatio-temporal location $\mathbf{q}=(x, y, t)$ lies within an object surface in BEV. 
Given this implicit occupancy representation, our planner can understand complex environments by having the different costs $f_i$ reason about occupancy at different spatio-temporal locations $\mathcal{Q}^{(i)}_\tau$.

\subsection{Trajectory Sampling and Query Points} 

\paragraph{Trajectory Sampler}
We consider a set of candidate trajectories $\tau \in \mathcal{T}$ starting from the current ego state and going 5 seconds into the future.
The ego state consists of the SDV current location in BEV, speed, and steering. 
A trajectory $\tau$ contains a sequence of kinematic bicycle model \cite{kong2015kinematic} states for each time step in the planning horizon.
It is crucial that the set of sampled trajectories, while small enough to enable real-time computation, encompasses a range of maneuvers, including lane following, lane changes, nudges to avoid encroaching objects, hard brakes. 
To accomplish this efficiently, we opt for a sampling approach that takes into account the lane structure.
More precisely, we use the lane centerlines from the HD map as base paths and sample longitudinal and lateral profiles in Frenet frame \cite{werling2010optimal,sadat2019joint}.
As a result, the sampled trajectories align with appropriate lane-based driving, while  incorporating lateral variations.

\paragraph{Points of Interest} The goal of these points of interest is to cover the relevant areas around the ego vehicle throughout the candidate plans. For every trajectory $\tau$ and time step $t$ (sampled every 0.5s), we consider points within the ego bounding box as well as points forwards, backwards, and to the sides of the ego. For simplicity, we sample a uniform grid of points within the ego box at a certain resolution, and simply shift this grid forward/backward by the length of the ego vehicle and right/left by the width to obtain all the points of interest. This process is shown in \cref{fig:model_diagram} for the final time step of two trajectories.

\paragraph{Point Quantization} 
Since the points of interest are sampled along and around the trajectories $\mathcal{T}$, which are generated to ensure coverage of the available actions, the distance between multiple pairs of query points $||\mathbf{q}_j - \mathbf{q}_k||_2$ from different trajectories can be very small, as depicted in Fig.~\ref{fig:teaser}.
We leverage this observation to improve the efficiency of our planner by quantizing the query points with a certain spatial resolution and only querying $\psi$ with the unique set of points after quantization.
We tune the quantization resolution to maximize efficiency without sacrificing driving performance.
Empirically, we find this to reduce the number of queries by two orders of magnitude, from millions to tens of thousands.

\subsection{Implicit Occupancy Model}

Our implicit occupancy model consists of a scene encoder that provides a BEV latent representation of the environment $\mathbf{Z}$, and an implicit occupancy decoder $\psi$ that attends to the latent scene representation to predict occupancy probability at query points.
By maintaining an intermediate occupancy representation, we can make planning decisions which are interpretable and consistent with the occupancy predictions.

\paragraph{Scene Encoder} 
We use as input a sequence of LiDAR point clouds, containing the 5 latest LiDAR sweeps.
Each sweep contains a set of points with coordinates $(p_x, p_y, p_h)$, where the $(p_x, p_y)$ is the point location in the SDV coordinate frame while $p_h$ is the height over the ground.
We then voxelize \cite{yang2018hdnet} the LiDAR in BEV to obtain a 3D tensor where the different sweeps are concatenated along the height dimension.
Since the behavior of other traffic participants is highly influenced by the road topology, we make use of the prior knowledge stored in the HD map to provide important cues about the regions they might occupy and how they could move. 
More precisely, we raster the polylines representing the lane centerlines in the HD map as a BEV binary map with the same spatial resolution as the LiDAR. 
Following \cite{casas2018intentnet, agro2023implicito}, our scene encoder uses two convolutional stems for processing the voxelized  LiDAR and map raster respectively.
The resulting feature maps are concatenated along the channel dimension and passed through a lightweight Feature Pyramid Network \cite{lin2017feature} to get a fused BEV feature map $\mathbf{Z}$ containing information from both modalities at half resolution of the inputs.
Intuitively, the latent scene embeddings $\mathbf{Z}$ contain local geometry, motion and semantic descriptors from the area within the receptive field of our encoder. 

\paragraph{Implicit Occupancy Decoder} 
Leveraging the latent scene embedding $\mathbf{Z}$, our implicit occupancy decoder predicts the occupancy probabilities 
at a set of query points $\mathcal{Q} = \{\mathbf{q}_j\}_{j \in [1, |Q|]}$.
In more detail, each query point $\mathbf{q}=(x,y,t)\in\mathbb{R}^3$ denotes a spatio-temporal point in BEV at a future time $t$. 
We exploit ImplicitO~\cite{agro2023implicito}, 
a recently proposed state-of-the-art architecture for occupancy prediction. 
Given a query point, it bilinearly interpolates a latent vector at the query point BEV location $(x, y)$, and uses it to predict locations to attend using deformable attention \cite{zhu2020deformable}. 
With the attended latent vector, an MLP decoder predicts occupancy for a particular query point. 
This simple architecture has a great advantage over prior methods that predict occupancy grid maps using CNNs: 
it can attend anywhere in the BEV latent instead of being limited by the CNN's receptive field to a small region. 
This is crucial since vehicles can travel very fast, so to accurately predict the occupancy into the future (e.g., at $t=5$s), 
the model needs to find the original LiDAR evidence at $t=0$s, which may be 150-200 meters behind.
The specific query points $\mathcal{Q}$ used during inference depend on the trajectory sampler and costs explained next.

\subsection{Trajectory Costing}

In order to plan an effective trajectory, we must consider various factors of driving such as collision likelihood, traffic violations, goal location, and comfort.
To meet this desiderata in a way that the decisions are explainable, we consider a set of interpretable costs.
We split costs $f_i$ into \textit{agent-agnostic} costs and \textit{agent-aware} costs.
At a high level, agent-agnostic costs describe the comfort, rule compliance, and progress of a candidate trajectory.
Agent-aware costs evaluate the safety of the trajectories with respect to other agents using the outputs of our implicit occupancy model $\psi$ at the query point locations $\mathcal{Q}$. 
In the next paragraph we describe our agent-aware costs at a high-level. 

A \emph{collision} cost considers the maximum probability of collision for each time step $t$ of each trajectory candidate $\tau$.
Specifically, we gather the occupancy at the query points within the ego bounding box $B_\tau^t$, and take the maximum probability. 
For each trajectory, we aggregate the maximum probabilities over time steps with a cumulative sum to further penalize trajectories that collide earlier on.
A \emph{longitudinal buffer} cost penalizes trajectories with agents too close in front or behind the ego vehicle by gathering the occupancy at those locations.
We apply a linear decay to the cost based on the distance with respect to the ego.
Similarly, \emph{lateral buffer} cost penalizes trajectories that remain in close lateral proximity to other agents in the scene.

\subsection{Learning}
\label{sec:learning}
We optimize our motion planner in two stages. 
We first train the implicit occupancy model to learn to perceive and forecast.
In a second stage, we freeze the occupancy model and train the cost aggregation weights $\{w_i\}$ to imitate an expert driver.
This two-stage training maintains the interpretability of the occupancy intermediate representation and allows the cost aggregation weights to train with stable occupancy predictions.

\paragraph{Occupancy}
An advantage of having intermediate representations is that one can use much richer and denser supervision to perceive the world and understand its dynamics rather than just imitating the expert trajectory.
For each continuous query point $\mathbf{q} \in \mathcal{Q}$, occupancy is supervised with binary cross entropy loss.
Following \cite{agro2023implicito}, we train with a batch of continuous query points $\mathcal{Q}$, uniformly sampled across the spatio-temporal volume.

\paragraph{Costing} 
We train the cost aggregation such that the behavior of our planner imitates that of an expert.
Because selecting the trajectory with the minimum cost from a discrete set is not a differentiable process, we use the max-margin loss to penalize trajectories that are either unsafe or have a low cost but differ significantly from the expert driving trajectory following prior works \cite{sadat2019joint,sadat2020perceive}.
Intuitively, this loss incentivizes the expert trajectory $\tau_e$ to have a smaller cost $J$ than all other trajectories.
More precisely, our objective is

{
   \small
   \begin{align} 
      \mathcal{L}_w=\max _{\tau}\left[\Delta J_r\left(\tau, \tau_e\right)+l_{\mathrm{im}}+\sum_t\left[\Delta J_c^t\left(\tau, \tau_e\right)+l_{\mathrm{c}}^t\right]_{+}\right]_{+},
   \end{align}
}
where $\Delta J\left(\tau, \tau_e\right) = J\left(\tau_e\right)-J(\tau)$ is the difference between the cost of the expert trajectory $\tau_e$ and the candidate trajectory $\tau$;
$J_c^t$ is the collision cost at a particular time step into the future, and $J_r$ are the rest of the costs, aggregated;
$\left[\right]_+$ represents the ReLU function; 
and $l_{\mathrm{im}}$ and $l_{\mathrm{c}}^t$ are the imitation and safety margins, respectively.
Note we omit $\mathbf{Z}, M$ from $J$ for brevity.
The imitation margin is simply the distance between the trajectory waypoints in $\tau_e$ and $\tau$, and the safety margin is whether the candidate trajectory $\tau$ collides with any ground-truth object.

As a way to mitigate distribution shift from open-loop learning to closed-loop deployment, we exploit dataset aggregation \cite{ross2011reduction} by combining the initial set consisting of expert demonstrations on expert states with another dataset generated with states visited by our learned autonomy model.

\section{Experiments}

\begin{table*}[t]
    \footnotesize
    \centering
    \begin{tabularx}{\textwidth}{l sssssss ss sssss}
        \toprule
         &\textbf{Mission} &\multicolumn{7}{c}{\textbf{Safety and Compliance}} &\multicolumn{4}{c}{\textbf{Progress, Consistency and Comfort}} \\ 
        \cmidrule(l{5pt}){1-1} \cmidrule(l{5pt}){2-2} \cmidrule(l{5pt}r{5pt}){3-9} \cmidrule(l{5pt}r{5pt}){10-13}
        &GSR\ua &ECR\da &PCR\da &\multicolumn{4}{c}{MinTTC\ua} &TVR\da &Progr.\ua &L2E\da &P2P\da &Jerk\da \\ \cmidrule(l{5pt}r{5pt}){5-8}
        & & & & $p10$ & $<1s$ & $<2s$ & $<5s$ & & & & & \\\midrule
        \textsc{Expert} &91.6\% &0.0\% &2.7\% &4.64 &0.0\% &1.1\% &17.9\% &6.8\% &454.2 &0.0 &15.1 &0.30 \\
        \midrule
        \textsc{PlanT} \cite{renz2022plant} &23.7\% &10.0\% &2.9\% &3.42 &8.4\% &8.9\% &19.5\% &63.7\% &331.7 &95.0 &42.8 &1.03 \\
        \textsc{CIL} \cite{codevilla2018end} &0.0\% &16.3\% &5.6\% &1.55 &8.4\% &15.3\% &36.8\% &100.0\% &64.6 &298.2 &19.7 &1.38 \\
        \textsc{NMP} \cite{zeng2019end} &51.6\% &14.2\% &21.4\% &0.50 &10.5\% &16.3\% &62.1\% &18.4\% &337.6 &133.3 &44.2 &1.21 \\
        \textsc{P3} \cite{sadat2020perceive} &76.3\% &6.8\% &8.3\% &3.44 &4.7\% &5.3\% &29.5\% &11.6\% &\textbf{436.1} &37.5 &14.7 &0.17 \\
        \textsc{OccFlow} \cite{mahjourian2022occupancy} &60.5\% &30.5\% &28.0\% &0.40 &24.2\% &28.4\% &44.2\% &36.8\% &385.3 &55.2 &\textbf{12.8} &\textbf{0.14} \\
        \midrule
        \ourmodel{} (Ours) &\textbf{84.7\%} &\textbf{2.1\%} &\textbf{1.0\%} &\textbf{4.67} &\textbf{1.6\%} &\textbf{1.6\%} &\textbf{14.2\%} &\textbf{8.4\%} &430.6 &\textbf{36.6} &15.6 &0.29 \\
        \bottomrule
    \end{tabularx}
    \caption{\textbf{[Safety-focused set] Closed-loop simulation results}}
    \vspace{-10pt}
    \label{tab:closed-loop-safety}
\end{table*}

In this section, we first compare \ourmodel{} to state-of-the-art autonomy models, 
measuring their ability to drive in closed-loop in both safety-focused and canonical highway driving. Then, we analyze trade-off between runtime and driving quality of different methods, and ablate the importance of \ourmodel{}'s query point quantization.
Finally, we show qualitative results.

\subsection{Data}

We utilize our high-fidelity end-to-end simulator to generate datasets for open-loop evaluation and training as well as a closed-loop benchmark.
Our simulator can generate both LiDAR sensor data \cite{manivasagam2020lidarsim} as well as intelligent actor behaviors \cite{IDM,MOBIL}.
We generate several datasets and benchmarks with distinct purposes.

\paragraph{Safety-focused set} 
We generate a dataset composed of $\sim900$ safety-focused scenarios, split into $\sim700$ for training and $\sim200$ for evaluation. 
The scenarios in this set contain scripted interactions %
such as cut ins, blocked lanes, agents merging onto the highway, ego merging onto the highway through an off-ramp, and exiting through an off-ramp, aggressively slow/fast moving actors.
The ego is provided with a variety of mission routes including keep lane, lane change, or lane merge.
These scripted interactions are parameterized by highly controllable parameters such as time-to-collision, time-to-arrival to a merge point, initial speeds.
Training and validation utilize non-overlapping parameter values to ensure generalization is tested.

\paragraph{Canonical driving set} 
Contains a total of $\sim1700$ scenarios, $\sim1400$ for training and $\sim300$ for evaluation, each lasting around 20 seconds.
This set is generated from a set of randomly distributed discrete and continuous parameters which control the map 
(using a mix of simulated and real maps to get exposure to diverse curvatures, speed limits, number of lanes, and distinct topologies), 
ego starting location, initial agent conditions such as speed, location, and maximum acceleration limits to ensure a diverse distribution of scenarios, 
and a variety of vehicle types. %
In this set, the ego vehicle does not have a particular goal or mission (in contrast to safety-focused set), making the task easier. %
Similar to the safety-focused set, training and evaluation are non-overlapping by ensuring different random parameter values.

\paragraph{Open-loop vs. closed-loop} 
Both safety-focused and canonical scenario sets can be used for open-loop as well as closed-loop execution.
In open-loop, an expert planner is in charge of driving and the policy under train/test is only proposing plans that do not get executed and hence do not affect the next state of the world.
The expert is a privileged planner that has access to the internal state of the simulation, receiving ground truth current states for other actors in the world as well as their most recent future plan.
In contrast, during closed-loop simulation, the planner policy under test is driving the ego vehicle. In other words, the states the ego vehicle visits are dependent on the policy under test.
We rely on closed-loop simulation as our ultimate benchmark, since it best captures the driving performance a planner would have in the real-world, where the planner actions affect the environment.

\subsection{Experimental setup}

\paragraph{Baselines} There are three main families of baselines we consider. 
(1) \textit{Object-based autonomy} composed of separate object detection and planning modules. We compare against PlanT \cite{renz2022plant}, a recent method that leverages an object detector and then a Transformer-based planner that reasons about the detected objects together with the route.
(2) \textit{Sensor-to-plan autonomy} approaches including direct regression of the plan through Conditional Imitation Learning (CIL) \cite{codevilla2018end} as well as NMP \cite{zeng2019end}, which learns a cost-prediction network to rank samples.
(3) \textit{Occupancy-based autonomy} that leverages grid-based occupancy and a sample-based planner \cite{sadat2019joint}. We consider OccFlow \cite{mahjourian2022occupancy} and P3 \cite{sadat2020perceive} as two possible architectures to predict temporal occupancy grids.

For P3 and OccFlow we train in two stages.
In the first stage, we train the occupancy module on nominal scenarios with dense traffic.
In the second stage, we train the planner on safety focused scenarios. 
For the remaining baselines PlanT, CIL, and NMP since they are trained in a single stage 
we train the entire module on the combined set of canonical and safety training scenarios.

\paragraph{Metrics}
We consider a comprehensive set of metrics to provide a holistic view of autonomy performance.

To evaluate safety we propose the following metrics. 
\emph{Execution Collision Rate (ECR)} measures the percentage of scenarios with a collision.
\emph{Plan Collision Rate (PCR)} calculates the collisions of our planned trajectory with respect to the simulated actors' future trajectories.
\emph{Minimum Time-To-Collision (MinTTC)} computes the minimum time buffer (in seconds) --across a scenario-- before a collision occurs, taking into account the ego plan and the actor plan. 
To better illustrate the worst-case we compute percentile 10 ($p10$), and to get an idea of different levels of risk exposure we report cumulative TTC buckets ($<1s$, $<2s$, $<5s$), in decreasing order of severity.
To measure compliance, we use \emph{Traffic Violation Rate (TVR)} to show the percentage of scenarios where the ego vehicle violates a lane boundary or speed limit, goes off-road, or collides. 

On top of safety and compliance, it is also important driving makes sufficient progress towards the goal and is smooth. 
We compute \emph{progress within map (Progr.)} as the meters traveled without going off-road.
\emph{L2 distance to Expert (L2E)} measures distance from the expert demonstrations. In closed-loop, this corresponds to the distance between the executed trajectories. 
\emph{Plan-to-plan consistency (P2P)} computes the distance over the finite horizon planned trajectories for consecutive plans.
\emph{Jerk} measures the level of discomfort of the planned trajectories.
Note secondary metrics are only meaningful when primary metrics are similar, 
since avoiding collisions naturally causes higher jerk and plan-to-plan distance
thus safety and compliance must be prioritized at all times.
Finally, for self-driving vehicles it is critical to reach a target location. %
To measure this, we propose \emph{Goal Success Rate (GSR)} as the ratio of scenarios where the ego vehicle reaches the goal without colliding or violating traffic rules.

\subsection{Results}

\paragraph{Comparison against state-of-the-art}
Table~\ref{tab:closed-loop-safety} shows our main benchmark results, where we evaluate the baselines and our method in closed-loop, on the safety-focused highly-interactive scenarios.
We find our method is much safer than the baselines, as showcased by lower ECR, lower PCR as well as higher TTC. 
Moreover, it complies better with the rules of the road as showcased by our lower TVR.
On top of being safer and more compliant, \ourmodel{} achieves the highest GSR %
indicating the maneuvers we pick at a behavioral level are effective.
On the secondary metrics our model ranks high in progress, best in expert imitation, and has fairly consistent plans over time. 
While we do observe higher jerk than some baselines, our jerk is closer to that exhibited by the expert.

\begin{table}[t]
    \scriptsize
    \centering
    \setlength\tabcolsep{4pt}
    \begin{tabularx}{\columnwidth}{l r r r r r r r}
        \toprule
         &ECR\da &PCR\da &MinTTC\ua &TVR\da &Progr.\ua &L2E\da &Jerk\da \\
        &      &       &$p10$     &       &          &       & \\
        \midrule
        \textsc{Expert} &0.0\% &0.0\% &10.00 &2.2\% &317.6 &0.0 &0.40 \\
        \midrule
        \textsc{PlanT} &29.6\% &18.2\% &0.40 &86.6\% &214.6 &66.3 &1.08 \\
        \textsc{CIL} &49.0\% &28.8\% &0.40 &92.0\% &50.7 &255.1 &1.35 \\
        \textsc{NMP} &7.0\% &22.8\% &1.67 &8.3\% &214.9 &80.6 &1.20 \\
        \textsc{P3} &2.9\% &6.0\% &3.88 &4.8\% &\textbf{317.3} &36.2 &0.28 \\
        \textsc{OccFlow} &20.4\% &20.9\% &0.40 &22.9\% &306.0 &39.6 &\textbf{0.23} \\
        \midrule
        \ourmodel{} (Ours) &\textbf{0.0\%} &\textbf{0.5\%} &\textbf{4.64} &\textbf{2.2\%} &299.1 &\textbf{33.7} &0.43 \\
        \bottomrule
    \end{tabularx}
    \caption{\textbf{[Canonical set] Closed-loop simulation results}}
    \vspace{-10pt}
    \label{tab:closed-loop-canon}
\end{table}

\begin{figure}[t]
    \centering
    {\arrayrulecolor{gray}
    \setlength{\tabcolsep}{0pt}
    \renewcommand{\arraystretch}{0.5}
    \begin{tabular} {c c}
        \includegraphics[width=0.49\columnwidth]{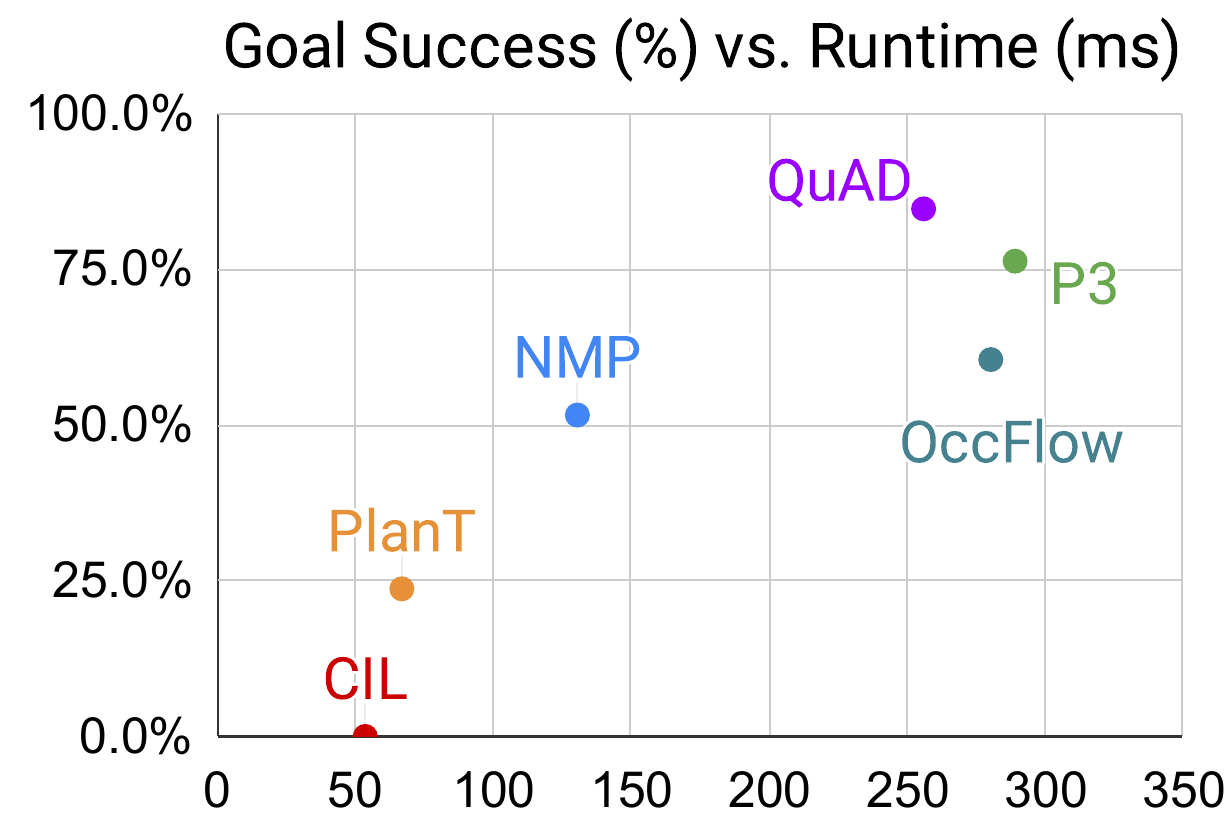} &
        \includegraphics[width=0.49\columnwidth]{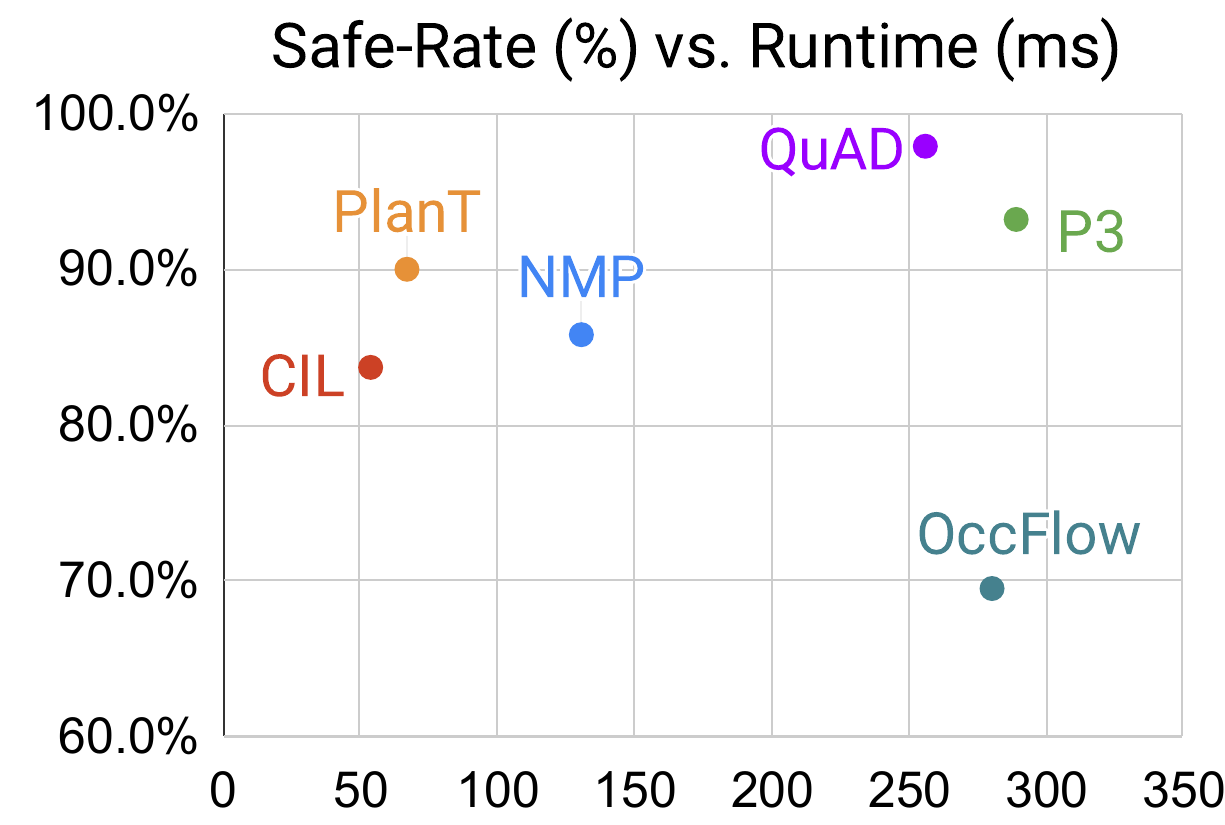}
    \end{tabular}
    }
    \caption{\textbf{Driving quality vs. runtime comparison}}
    \vspace{-10pt}
    \label{fig:runtime}
\end{figure}

Table~\ref{tab:closed-loop-canon} showcases our closed-loop results in canonical driving scenarios. 
Similar to previous results, we observe a substantial safety improvement, being able to attain very close safety and compliance performance to that of the expert. 
\ourmodel{} has the closest imitation to expert and ranks highly on progress too, striking a good balance between safety and progress towards the goal. 
When taking into consideration the priority safety has over progress, our method clearly outperforms the baselines.
Note that CIL and PlanT do not use lane based trajectory samples and this results in much worse TVR metric due to lane boundary violations.

\paragraph{Inference runtime}
Fig.~\ref{fig:runtime} shows how methods compare when considering the balance between driving quality and runtime, an important factor for real-world deployment.
We profile in a crowded scenario with 74 agents and 11 lanes to stress-test the different autonomy systems. 
We highlight that \ourmodel{} is faster than other occupancy-based autonomy models (\textsc{P3}, \textsc{OccFlow}), while attaining the best driving quality both in terms of mission achievement and safety.

\paragraph{Query point quantization ablation}
Our proposed query point quantization is necessary to reap the benefits from an implicit occupancy decoder without inquiring a prohibitive runtime.
The naive alternative of using the continuous query points directly after the generation step incurs very high inference times of $\sim$700ms.
This is due to the high number of points, caused by the high overlap between the candidate plans especially at the beginning of the temporal horizon.
Quantization becomes necessary in practice.
Utilizing a resolution of 0.5m (the one used in the rest of the experiments), autonomy runtime reduces to $\sim$250ms while the execution collision rate only increases from $\sim1\%$ to $\sim2\%$, thus providing significant improvements over the baselines.

\begin{figure}[t]
    \centering
    {\arrayrulecolor{gray}
    \setlength{\tabcolsep}{0pt}
    \renewcommand{\arraystretch}{0.5}
    \begin{tabular} {c c}
        \includegraphics[width=\columnwidth]{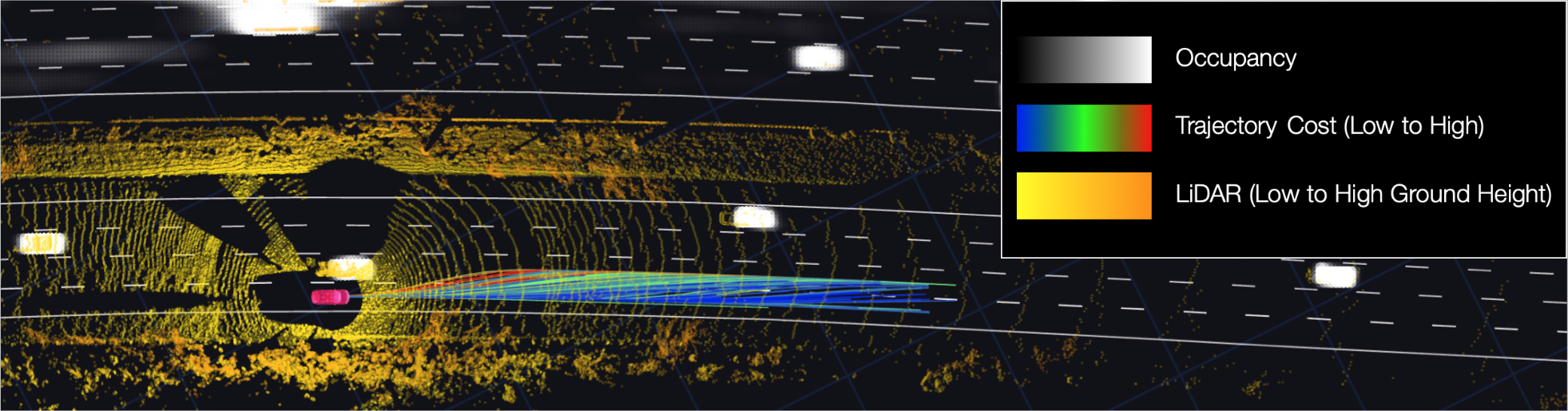}  \\
        \includegraphics[width=\columnwidth]{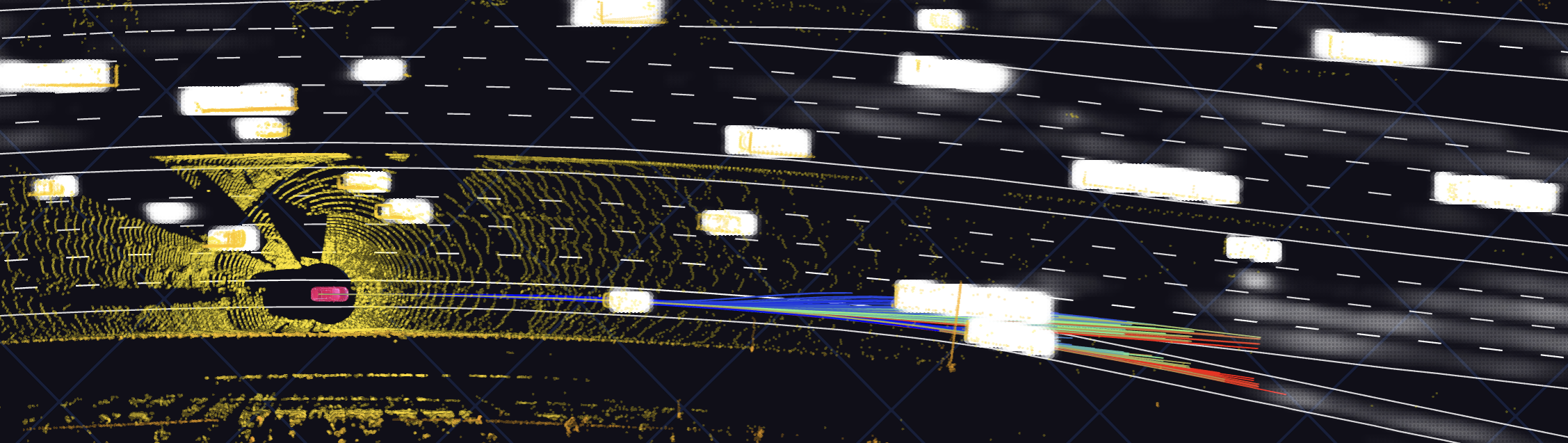} \\
        \includegraphics[width=\columnwidth]{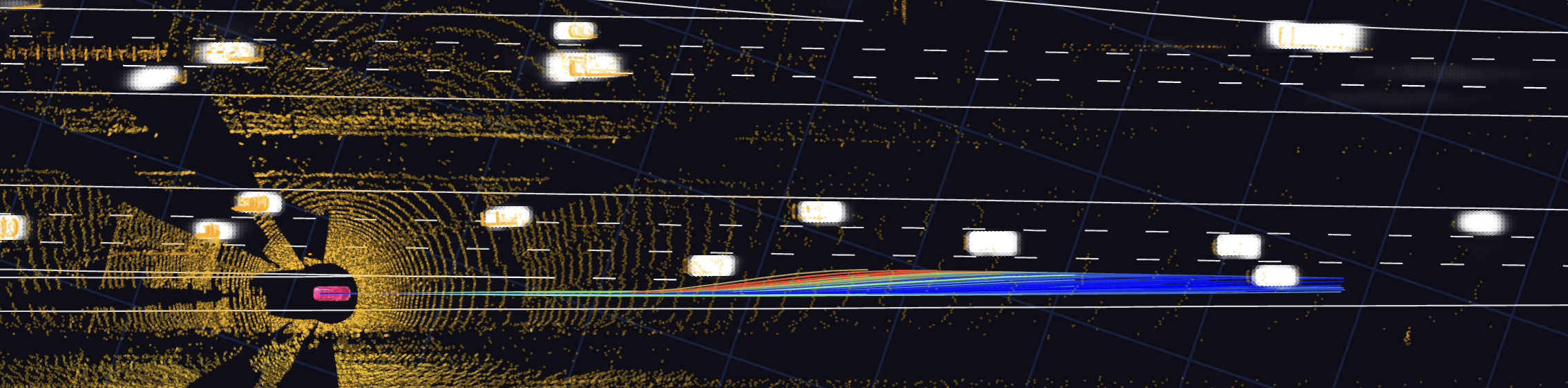} \\
    \end{tabular}
    }
    \caption{\textbf{Qualitative results}. 
    We visualize the LiDAR point cloud, map, predicted occupancy (by querying $\psi$ at a regular grid, solely for illustration purposes), and cost associated with the trajectory samples.
    From top to bottom: a lane change, a re-incorporation near an off-ramp, and a merge.}
    \vspace{-10pt}
    \label{fig:qualitative}
\end{figure}

\paragraph{Qualitative results}
Fig.~\ref{fig:qualitative} showcases example scenarios where the goals instructed to the ego consist of semantically diverse and interesting maneuvers.
From the cost distribution, we can observe that our planner can understand the mission route and progress towards the goal, it can speed up to pass slow moving vehicles when instructed to lane-change, and can plan smooth merge trajectories at on-ramps.

\section{Conclusion}
In this paper, we have proposed an interpretable motion planner leveraging spatio-temporal occupancy queries to effectively understand the current and future free-space from sensor data.
We showcased our proposed autonomy can drive more safely and progress further than contemporary object-based, sensor-to-plan and occupancy-based autonomy models.
Our framework also achieves faster runtime than its closest competitors.

{
\bibliographystyle{ieee_fullname}
\bibliography{egbib}
}

\pagebreak

\section{Appendix}

We divide the supplementary material into three primary sections. 
We start by providing additional details about our method and experimental setup in Section~\ref{sec:additional_details}.
We then include supporting quantitative results in Section~\ref{sec:supp_quant}, including tables with the complete set of metrics as well as an ablation study.
Finally, we show additional qualitative results of \ourmodel{} as well as the baselines in Section~\ref{sec:add_qualitative}.

\section{Additional Experiment Details}
\label{sec:additional_details}

\subsection{Costing Methodology}

In order to achieve satisfactory performance in both safety critical and nominal scenarios,
we require a sophisticated combination of different costs to be utilized. In this 
section we will go into more detail on how we perform this costing process. 

As mentioned in the main paper, this costing is carried out for every trajectory candidate in the set
$\mathcal{T} = \{ \tau_s \}_{s \in [1, S] }$, where $\tau_s = \{(x_t, y_t, \theta_t, v_t, a_t, \kappa_t)\}_{t \in [0,T]}$
and $T$ is the number of timesteps in the planning horizon and we are considering $S$ sample trajectories.

\subsubsection{Agent Agnostic Costs}

\emph{Comfort} cost penalizes high jerk, acceleration, lateral acceleration, and curvature.
\emph{Corridor} cost favors trajectories that stay close to lane centerlines and do not violate lane boundaries.
\emph{Solid lane boundary} cost helps avoid trajectories that drive over solid lane boundaries.
\emph{Speed limit} cost prevents the ego from reaching speeds above the limit over the planning horizon.
\emph{Progress} rewards distance traveled along the trajectory.
\emph{Route} prioritizes trajectories that bring us closer to the mission route, where the reward is modulated by the speed limit in the target lane of the trajectory plan.

\begin{figure}[h]
    \vspace{-5pt}
    \begin{center}
       \includegraphics[width=\columnwidth]{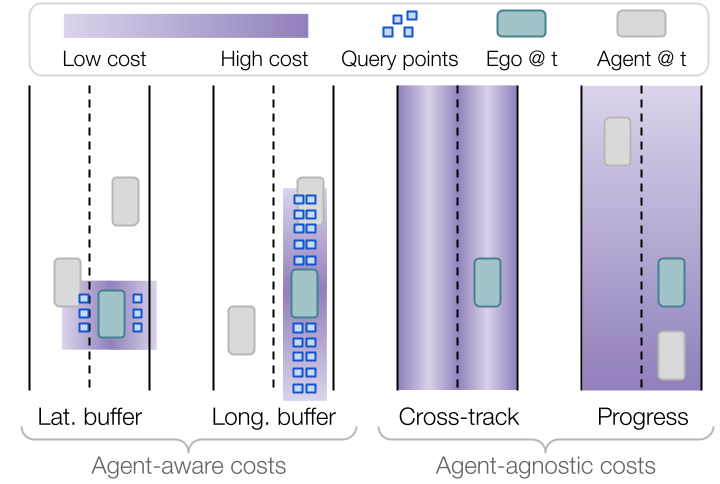}
    \end{center}
    \vspace{-10pt}
    \caption{\textbf{Illustration of example costs}. These costs are evaluated at multiple time steps $t$ along candidate plans, and aggregated to extract the trajectory cost. For agent-aware costs, the cost is modulated by the occupancy probability at the query points.}
    \label{fig:traj_costing}
    \vspace{-10pt}
 \end{figure}

Recall that $w_{i}$ is the learnable weight for the $i$-th cost, used for the linear combination of costs.
We define $\text{SD}_{l}(x,y)$ as the signed distance to a polyline $l$, and $v_{l}$ as the 
speed limit for lane $l$. Finally, we define $\text{lat}(x)$ as a function that isolates the component of vector $x$ that is perpendicular to the
nearest centerline, and $\text{long}(x)$ isolates the component along the nearest centerline.

\paragraph{Comfort cost}
\begin{equation}
\begin{split}
    \sum_t w_{\text{acc\_lat}} \text{lat}(a_t)^2 \\
    + w_{\text{acc\_long}} \text{long}(a_t)^2 
    + w_{\text{jerk}} \dot{a_t}^2 + w_{\text{curv}} \kappa_t^2 
\end{split}
\end{equation}

\paragraph{Corridor Costs}

\begin{equation}
    \sum_t w_{\text{corr}} |\text{SD}_{\text{centerline}}(x_t,y_t)|
\end{equation}

\paragraph{Solid Lane Boundary Cost}

\begin{equation}
\begin{split}
    \sum_t w_{\text{bound}}([\text{SD}_{\text{left\_boundary}}(x_t,y_t)]_+  \\
    + [\text{SD}_{\text{right\_boundary}}(x_t,y_t)]_+)
\end{split}
\end{equation}

\paragraph{Speed Limit}

\begin{equation}
    \sum_t w_{\text{speed}} ([v_t - v_l]_+)^2
\end{equation}

\paragraph{Progress}

\begin{equation}
    -\sum_t w_{\text{prog}} \sqrt{(x_t - x_{t-1}) ^ 2 + (y_t - y_{t-1}) ^ 2}
\end{equation}

\paragraph{Route}

\begin{equation}
    \sum_t w_{\text{route}} |\text{SD}_{\text{desired\_lane}}(x_t,y_t)|
\end{equation}

\subsubsection{Agent Aware Costing}

Here we elaborate on the mathematical definitions of the agent-aware costing. We denote 
$\mathcal{Q}_{B}$ to be the query points that we sample around a 
region $B$, and let $\mathcal{B}_t$ represent the motion bounding box at time t. 

At highway speeds (e.g., 30 m/s) a vehicle can travel a large distance in this time period. 
To avoid missing dangerous events due to limited temporal resolution, we are conservative and consider \textit{motion-blurred bounding boxes} $\mathcal{B}_t$ that are stretched to cover the space traveled during the time segment from $t$ to $t+1$.

We use $q = (x, y, t)$ to denote a spatio-temporal query point and $\psi(q, \mathbf{Z})$ to denote the occupancy probability for query point $q$.

\paragraph{Collision}

\begin{equation}
    w_{\text{col}} \sum_t (T - t) \max_{q \in \mathcal{Q}_{\mathcal{B}_t}}{[\psi(q, \mathbf{Z}) ]}
\end{equation}

where $t = 0, ..., T - 1$. %

\paragraph{Buffer Costs}

For both the longitudinal and lateral buffer costs, we use:

\begin{equation}
\begin{split}
    w_{\text{buf}} %
    \sum_t (T - t) \max_{q \in \mathcal{Q}_{\text{Buf}_t}}{[ \frac{dis(q)}{\max_{p \in \mathcal{Q}_{\text{Buf}_t}}{[dis(p)]}} \psi(q, \mathbf{Z}) ]}
\end{split}
\end{equation}

where $dis(q)$ is the euclidean distance between $q$ and the center of the motion bounding box $\mathcal{B}_t$.

We define the longitudinal buffer cost by $Buffer_t = forward(\mathcal{B}_t) \cup backward(\mathcal{B}_t)$,
where $forward$ translates the region $\mathcal{B}_t$ forward a distance equal to the length of $\mathcal{B}_t$. Similarly, $backward$ translates the region $\mathcal{B}_t$ backward a distance of the length of $\mathcal{B}_t$.

We define the lateral buffer cost by
$Buffer_t = right(\mathcal{B}_t) \cup left(\mathcal{B}_t)$ 
where $right$ translates the region $\mathcal{B}_t$ right a distance of the width of $\mathcal{B}_t$.
Similarly, $left$ translates the region $\mathcal{B}_t$ left a distance of the width of $\mathcal{B}_t$.

\subsection{Implementation Details} 
We train our autonomy model in two stages:
First, our scene encoder and implicit query function $\psi$ are trained in canonical scenarios with background traffic to ensure a diverse set with plenty of agents in the scene are seen during training. 
For this first stage, we use the Adam optimizer over 20 epochs with a learning rate of $10^{-3}$ which decays to 0.25 of its value every 6 epochs, and a batch size of 2. We also use a weight decay factor of $10^{-4}$. 
In the second stage, we train the cost aggregation weights on safety-focused scenarios since safety is the main priority of a planner, and this set is richer in ego-to-agent close interactions. 
For this second stage, we use the Adam optimizer over 200 epochs with a learning rate of $10^{-1}$ and no weight decay along with a batch size of 64. 

\subsection{Dataset Evaluation}

We employ the use of two different kinds of scenario sets. The first is the 
canonical driving scenario set that consists of generic highway driving with realistic actors. 
The second consists of the safety set that is made up of very challenging 
safety critical scenarios. We will now break down the scenarios in each of these 
subsets. 

\subsubsection{Canonical Set}

The canonical training set consists of 663 simulations each 20s long. The closed loop canonical evaluation set consists 
of 321 scenarios simulated using a different set of initial condition and random seeds, 
but that ultimately follow the same distribution as the training set. We will discuss 
this distribution in the following section. 

The maps are taken from real world highways, and the background is similarly reconstructed using real scenes. 
We now provide a conceptual decomposition of 
the different high-level scenes that are in each scenario. 

\paragraph{Map Categorization}

\begin{figure*}[t]
    \centering
    {\arrayrulecolor{gray}
    \setlength{\tabcolsep}{0pt}
    \renewcommand{\arraystretch}{0.5}
    \begin{tabularx} {\textwidth} {X X X} %
        \includegraphics[width=0.7\columnwidth]{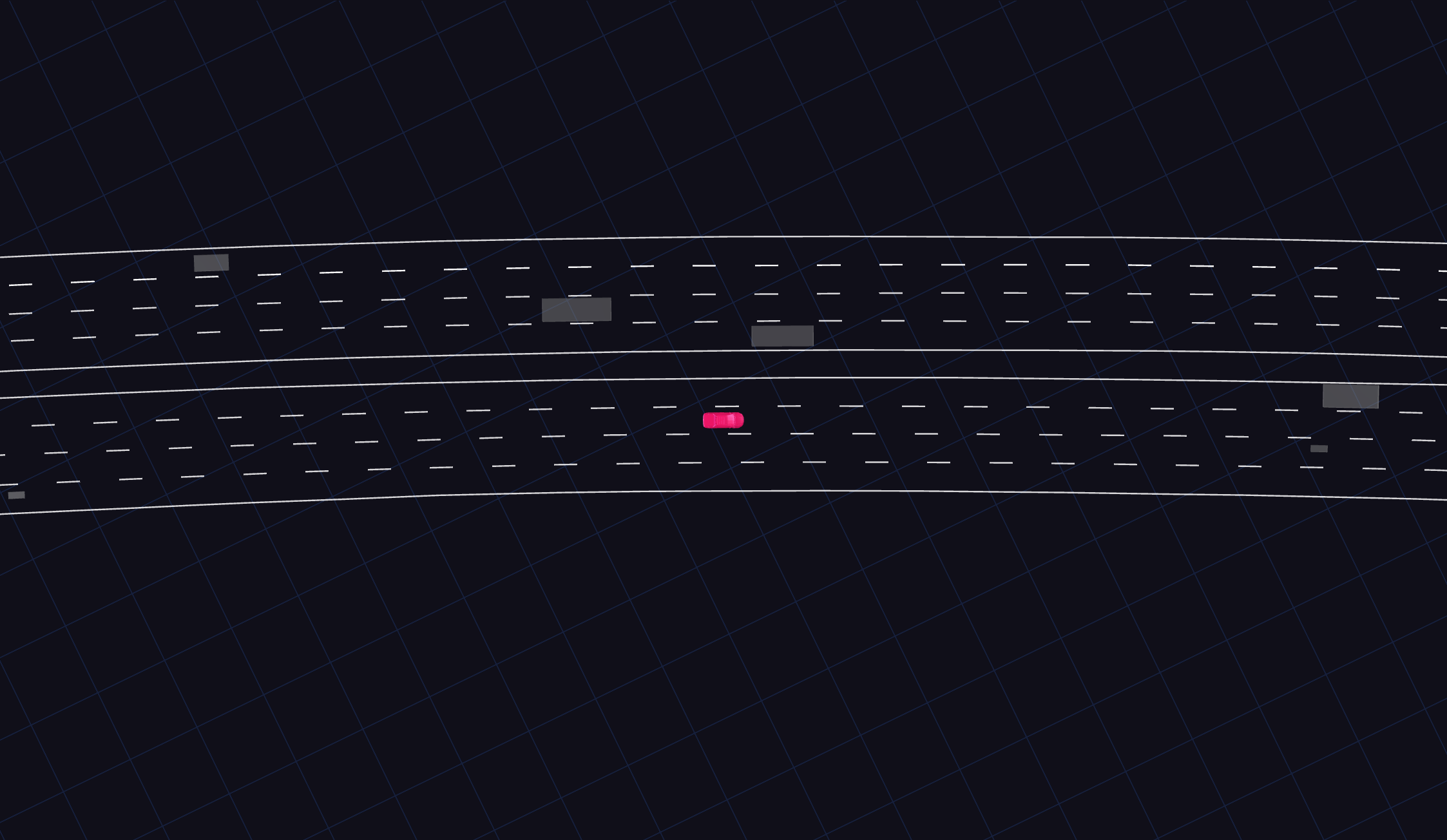} &
        \includegraphics[width=0.7\columnwidth]{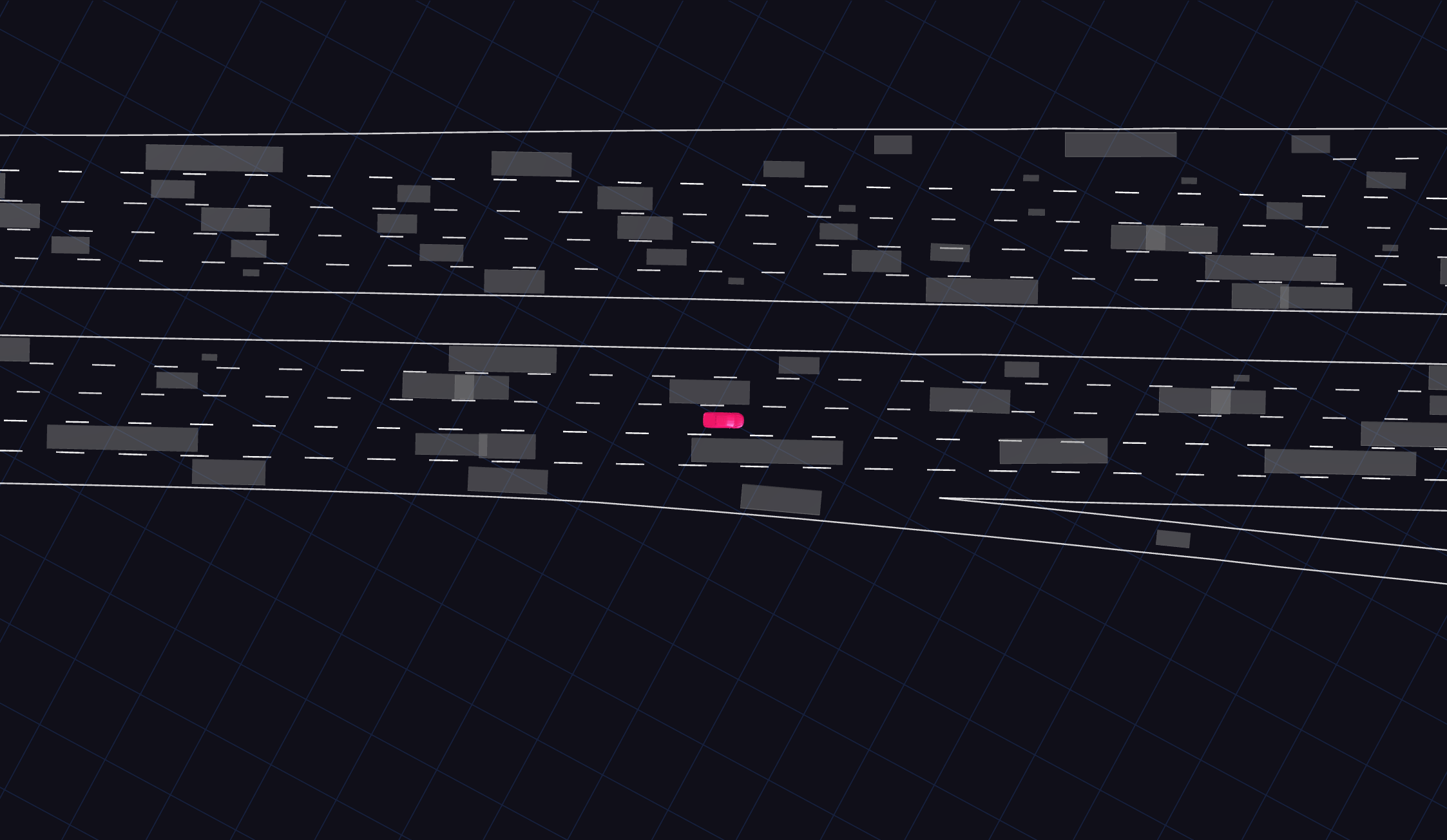} &
        \includegraphics[width=0.7\columnwidth]{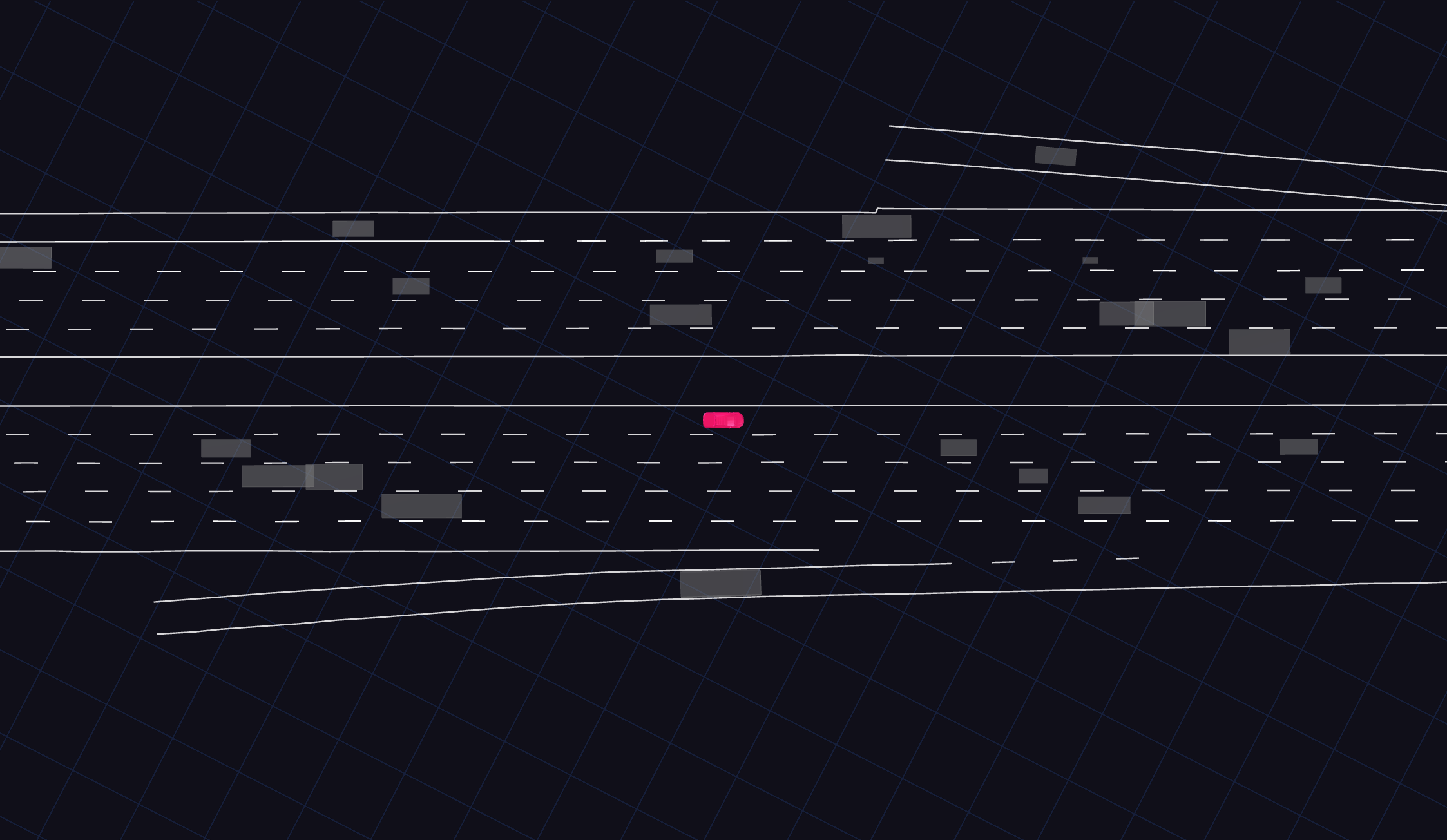}
    \end{tabularx}
    }
    \caption{\textbf{Map Topology Examples}. We illustrate a scenario in our dataset with
    (left) a standard highway road, (center) a fork in the road and (right)
    a merge into the road.}
    \vspace{-10pt}
    \label{fig:map_topology}
\end{figure*}

To choose an initial road map for simulation, we select a stretch of map
that contains the desired topology. In a third of the cases the desired topology 
is a merge lane, in another third it is an off-ramp, and the final third it is 
a stretch of road with neither of the above. Examples of these can be seen in
figure \ref{fig:map_topology}

\paragraph{Unpredictable Vehicles}

In order to simulate a wide diversity of different road actors we 
randomly sample how conservative vs. aggressive they are. 
There are additional static actors placed in the environment that the ego 
and the background actors have to navigate around. 
Finally, we also vary the density of the traffic in the scenes, with about 10\% 
of the scenes having extremely sparse traffic and about 10\% of the scenes being 
extremely dense traffic. 

\subsubsection{Safety Set}

Here we will describe the types of scenarios used to categorize our safety set. 
We note that we duplicate many of the high level scenarios with slightly altered 
initial conditions to ensure better coverage. Each one of these scenarios is simulated 
largely in isolation without background actors or a complex background mesh. 
The intent is to replicate a specific 
challenging scenario to evaluate the planner's interaction. For training, we have 690 
scenarios that are simulated for up to 20s (depending on when the goal is reached this 
may be shorter).

\begin{figure*}[t]
    \centering
    {\arrayrulecolor{gray}
    \setlength{\tabcolsep}{0pt}
    \renewcommand{\arraystretch}{0.5}
    \begin{tabularx} {\textwidth} {X X X} %
        \includegraphics[width=0.7\columnwidth]{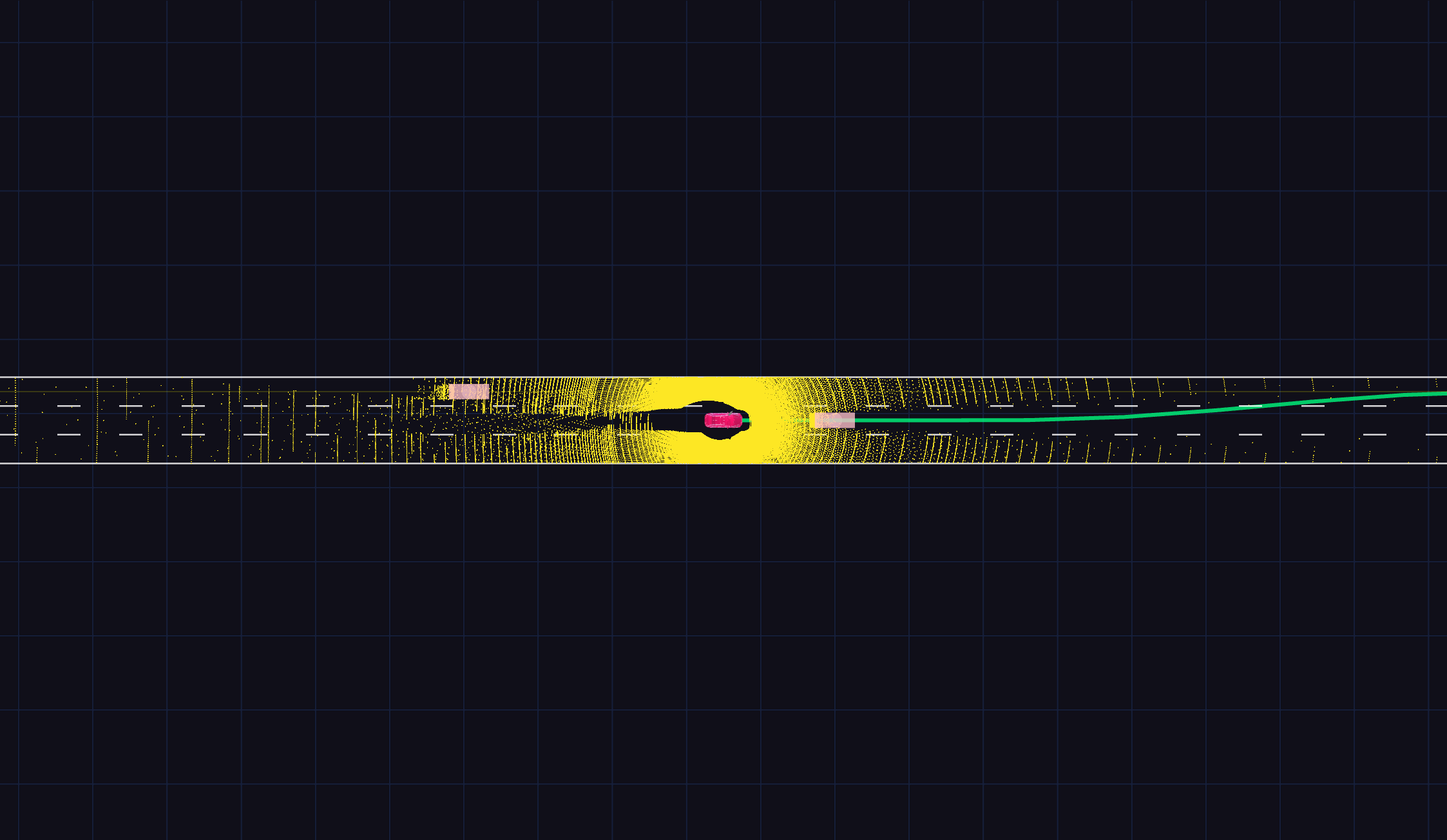} &
        \includegraphics[width=0.7\columnwidth]{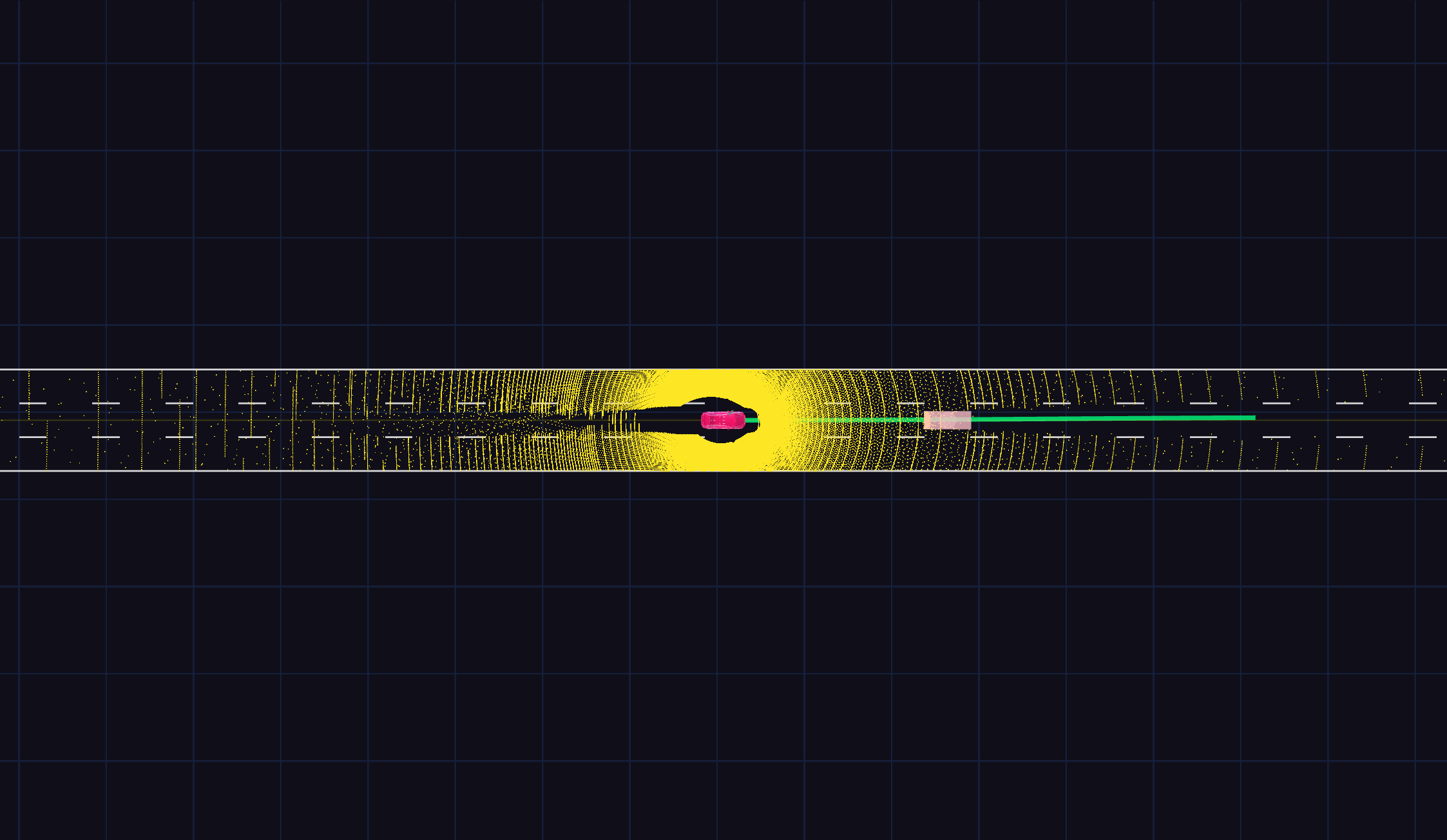} &
        \includegraphics[width=0.7\columnwidth]{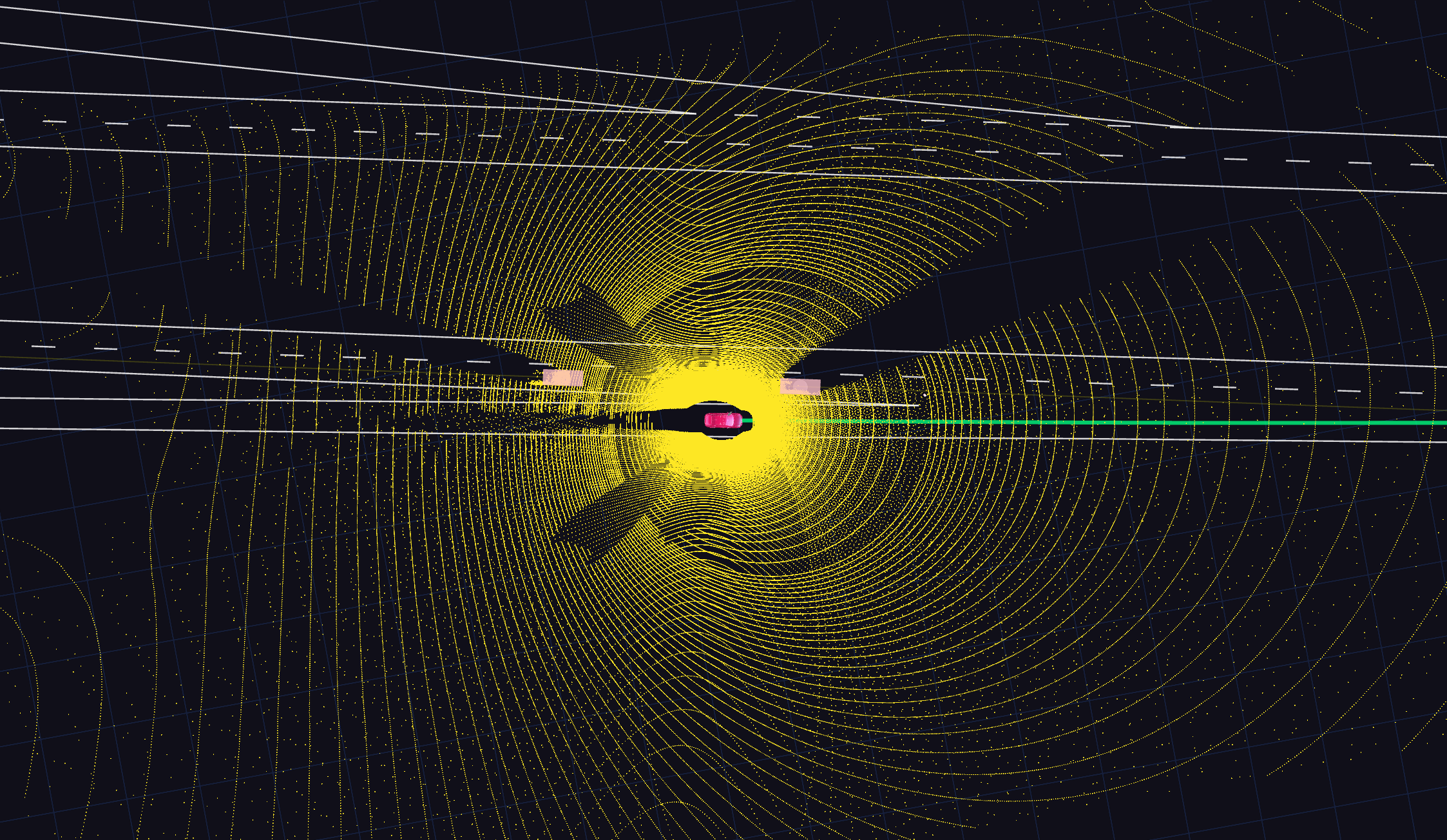}
    \end{tabularx}
    }
    \caption{\textbf{Safety Scenario Examples}. We illustrate scenarios in our 
    safety set where 
    (left) the self-driving vehicle must merge into a lane while contending 
    with other actors, (center) the vehicle must slow down to avoid crashing 
    into a slow vehicle in its lane and (right) the vehicle must merge into 
    the highway lane on an on-ramp.}
    \vspace{-10pt}
    \label{fig:safety_set_examples}
\end{figure*}

The safety scenarios consist of the following: first we have
\textbf{lane change} scenarios, where the goal lane for the
self-driving vehicle (SDV) is a different lane and there are actors around the vehicle 
in different configurations. This also includes
scenarios where scripted actors will try to merge into the target
lane, or will cut off the SDV. Next, we have the \textbf{lane follow}
scenarios where the goal of the SDV is to remain in the current lane
while contending with a series of challenging scenarios, such as cutting
in actors, suddenly braking actors, forking lanes, blocked lanes, and
multiple on-ramps converging to the current SDV position. Finally,
we have the \textbf{lane merge} scenarios. Here the self-driving vehicle
must merge into the target lane under a number of different scenarios.
Some of which include actors in the target lane moving at various
speeds, and some of which there are actors in front and behind the SDV
that make merging a complex balancing act. This also includes scenarios where
other actors taking an on-ramp must merge into the SDV's lane. Examples of each
of these can be seen in figure \ref{fig:safety_set_examples}.

\subsection{Expert Motion Planner}

In our work, we assume access to a privileged motion planner that
the different methods will learn to imitate during training. 
This expert motion planner has access to the true state and future plan of 
every actor on the road, and therefore is able to plan 
more optimal trajectories without any perception uncertainty. 

In practice this expert is a sample-based motion planner \cite{sadat2019joint,fan2018baidu} with hand tuned costs. 
As mentioned before, it 
has access to the current intention of each of the agents. Therefore, 
while these intentions may change as time goes on, avoiding collisions with these plans is 
an effective heuristic to plan good trajectories. 
The actors generally re-plan based on the surroundings (i.e., other actors and ego actions).
The specific costs
that are used by this ground truth planner are as follows:

\begin{itemize}
    \item \textbf{Collision cost}: cost for colliding into any other actors.
    \item \textbf{Headway cost}: cost for safety buffer violations between the SDV and the lead actor, or any future lead actors.
    \item \textbf{Comfort cost}: cost for acceleration/jerk.
    \item \textbf{Crosstrack cost}: cost for deviating from the centerline.
    \item \textbf{Progress cost}: penalizes travelling shorter distances.
    \item \textbf{Speed limit}: cost for exceeding the prescribed speed limit.
    \item \textbf{Corridor cost}: cost for moving outside the boundaries of the lane. 
    \item \textbf{Contingency}: cost for penalizing scenarios where a collision is guaranteed given maximum deceleration and no lane changing.
\end{itemize}

The hand tuned weighted combination of these costs, while not flawless, still gives the 
best overall results for our planning metrics, and therefore serves as a good guide 
for supervising the other methods.

\section{Additional Quantitative Results}
\label{sec:supp_quant}

\subsection{Full Results}

For completeness, we include here the main closed loop and open loop results on both 
the canonical and safety sets. Some less relevant metrics were omitted in the 
main body of the paper due to space constraints. We begin with the complete closed loop metrics 
on the canonical set. The results for this can be seen in Table~\ref{tab:closed-loop-canonical}.

\begin{table*}[t]
    \footnotesize
    \centering                     
    \begin{tabularx}{\textwidth}{l ssssssss ssss}\toprule
        \textbf{} &\multicolumn{7}{c}{\textbf{Safety and Compliance}} &\multicolumn{4}{c}{\textbf{Progress, Consistency and Comfort}} 
        \\ \cmidrule(l{5pt}r{5pt}){2-8} \cmidrule(l{5pt}r{5pt}){9-12}
        &ECR\da &PCR\da &\multicolumn{4}{c}{MinTTC\ua} &TVR\da &Progr.\ua &L2E\da &P2P\da &Jerk\da 
        \\\cmidrule(l{5pt}r{5pt}){4-7}
        & & &$p10$ &$< 1s$ &$< 2s$ &$< 5s$ & & & & & \\\midrule
        \textsc{Expert} &0.0\% &0.0\% &10.00 &1.0\% &1.0\% &8.6\% &2.2\% &317.6 &0.0 &19.7 &0.40 \\
        \midrule
        \textsc{PlanT} \cite{renz2022plant} &29.6\% &18.2\% &0.40 &22.9\% &27.4\% &51.9\% &86.6\% &214.6 &66.3 &33.7 &1.08 \\
        \textsc{CIL} \cite{codevilla2018end} &49.0\% &28.8\% &0.40 &28.3\% &44.6\% &66.2\% &92.0\% &50.7 &255.1 &25.9 &1.35 \\
        \textsc{NMP} \cite{zeng2019end} &7.0\% &22.8\% &1.67 &7.3\% &13.1\% &77.7\% &8.3\% &214.9 &80.6 &43.9 &1.20 \\ 
        \textsc{P3} \cite{sadat2020perceive} &2.9\% &6.0\% &3.88 &2.9\% &2.9\% &25.5\% &4.8\% &\textbf{317.3} &36.2 &22.7 &0.28 \\ 
        \textsc{OccFlow} \cite{mahjourian2022occupancy} &20.4\% &20.9\% &0.40 &19.7\% &19.7\% &36.6\% &22.9\% &306.0 &39.6 &\textbf{13.8} &\textbf{0.23} \\
        \midrule
        \ourmodel{} &\textbf{0.0\%} &\textbf{0.5\%} &\textbf{4.64} &\textbf{0.3\%} &\textbf{0.3\%} &\textbf{12.1\%} &\textbf{2.2\%} &299.1 &\textbf{33.7} &30.3 &0.43 \\
        \bottomrule
    \end{tabularx}
    \caption{\textbf{[Canonical set] Closed-loop simulation results}}
    \vspace{-10pt}
    \label{tab:closed-loop-canonical}
\end{table*}

The results tell a very similar story to the safety scenarios. Namely, that our proposed 
method outperforms the other when it comes to key safety metrics, while 
still achieving competitive progress, consistency and comfort. When driving in the real world 
it is of paramount importance that safety takes precedence over secondary progress based metrics.
However, it is still important to be able to operate reasonably well in the nominal case, which 
is exactly what we observe in the canonical closed loop metrics. In particular, while we are 
the best on all safety metrics, we also have very good progress, the best $L2E$, and comparable 
jerk to the expert.

We will now analyze the 
open loop results on the safety set, as seen in Table~\ref{tab:open-loop-safety}, and the 
open loop results on the canonical set as seen in Table~\ref{tab:open-loop-canonical}.

\begin{table*}[t]
    \footnotesize
    \centering                   
    \begin{tabularx}{\textwidth}{l ssssssssss}\toprule
    &\multicolumn{5}{c}{\textbf{Safety and Compliance}} &\multicolumn{4}{c}{\textbf{Progress, Consistency and Comfort}} 
    \\ \cmidrule(l{5pt}r{5pt}){2-6} \cmidrule(l{5pt}r{5pt}){7-10}
    &PCR\da &\multicolumn{4}{c}{MinTTC\ua} &Progr.\ua &L2E\da &P2P\da &Jerk\da 
    \\\cmidrule(l{5pt}r{5pt}){3-6}
    & &$p10$ &$< 1s$ &$< 2s$ &$< 5s$ & & & & \\ \midrule
    \textsc{Expert} &0.0\% &10.00 &0.0\% &0.0\% &0.0\% &117.9 &30.9 &15.7 &0.44 \\ 
    \midrule
    \textsc{PlanT} &35.2\% &1.45 &4.7\% &14.2\% &40.5\% &88.9 &74.4 &28.8 &1.04 \\ 
    \textsc{CIL} &18.6\% &1.55 &1.6\% &18.4\% &43.7\% &102.5 &40.8 &17.5 &1.94 \\ 
    \textsc{NMP} &11.8\% &2.54 &\textbf{0.0\%} &3.2\% &54.2\% &103.3 &43.8 &46.1 &1.39 \\ 
    \textsc{P3} &1.4\% &4.30 &\textbf{0.0\%} &0.5\% &20.0\% &118.9 &\textbf{29.0} &15.3 &\textbf{0.26} \\ 
    \textsc{OccFlow} &2.9\% &3.50 &\textbf{0.0\%} &2.1\% &23.7\% &\textbf{119.1} &29.2 &\textbf{15.0} &\textbf{0.26} \\
    \midrule 
    \ourmodel{} &\textbf{0.0\%} &\textbf{10.00} &\textbf{0.0\%} &\textbf{1.1\%} &\textbf{7.9\%} &118.6 &37.3 &31.2 &0.41 \\ 
    \bottomrule
    \end{tabularx}
    \caption{\textbf{[Safety set] Open-loop simulation results}}
    \label{tab:open-loop-safety}
    \end{table*}

Table~\ref{tab:open-loop-safety} summarizes the quality of the plans in open-loop simulation on the same safety-focused scenario set. In this evaluation, autonomy is free of any distributional shift, making this a much easier task.
The goal of this analysis is more so understanding which approaches are most robust to distribution shifts rather than finding out which approach did best in open-loop, as open-loop metrics are not very indicative of driving performance in the real-world.
The first observation when comparing these results to closed-loop safety-focused results in the main paper is that the plan collision rate ($PCR$) when planning from expert states is much lower than when the state visitation is determined by the policy under test. 
While this is expected to an extent, we observe that our method goes from $0\%$ to $1.0\%$, while for the strongest baselines the increase is very substantial (e.g., from $1.4\%$ to $8.3\%$ for \textsc{P3} and from $2.9\%$ to $28.0\%$ for \textsc{OccFlow}). Similar trends are observed in MinTTC. 
We note that this increased difficulty from open-loop to closed-loop is not obvious from the table for \textsc{PlanT}, \textsc{CIL}. After further inspection, the reason is that these methods quickly veer off-road during closed-loop simulation due to being extremely brittle to distribution shift, where there are no agents to collide with. In contrast, in open-loop the ego vehicle is driven by the expert and therefore forces these methods to plan when being surrounded by other agents.
Another metric that shows our method's robustness to distribution shift is the L2 to expert ($L2E$). In open-loop, our model plans do not show the best imitation to the expert executions.
However, when re-planning in states determined by our own previous actions in closed-loop, our method attains a driving trajectory that is the closest to the expert.

\begin{table*}[t]
    \footnotesize
    \centering                     
    \begin{tabularx}{\textwidth}{l ssssssssss}\toprule
    &\multicolumn{5}{c}{\textbf{Safety and Compliance}} &\multicolumn{4}{c}{\textbf{Progress, Consistency and Comfort}} 
    \\ \cmidrule(l{5pt}r{5pt}){2-6} \cmidrule(l{5pt}r{5pt}){7-10}
    &PCR\da &\multicolumn{4}{c}{MinTTC\ua} &Progr.\ua &L2E\da &P2P\da &Jerk\da 
    \\\cmidrule(l{5pt}r{5pt}){3-6}
    & &$p10$ &$< 1s$ &$< 2s$ &$< 5s$ & & & & \\ \midrule
    \textsc{Expert} &0.0\% &10.00 &0.7\% &0.7\% &1.0\% &93.5 &20.4 &12.3 &0.56 \\ 
    \midrule 
    \textsc{PlanT} &38.3\% &2.15 &1.0\% &8.5\% &60.7\% &87.6 &68.3 &30.4 &1.07 \\ 
    \textsc{CIL} &39.1\% &1.35 &1.4\% &37.3\% &75.6\% &69.6 &54.4 &24.8 &1.54 \\ 
    \textsc{NMP} &39.1\% &1.35 &1.4\% &37.3\% &75.6\% &69.6 &54.4 &24.8 &1.54 \\ 
    \textsc{P3} &1.3\% &4.50 &\textbf{0.3\%} &\textbf{0.3\%} &16.6\% &\textbf{95.7} &30.8 &34.2 &0.35 \\ 
    \textsc{OccFlow} &1.8\% &4.27 &\textbf{0.3\%} &\textbf{0.3\%} &18.6\% &\textbf{95.7} &\textbf{27.3} &\textbf{16.0} &\textbf{0.29} \\ 
    \midrule 
    \ourmodel{} &\textbf{1.0\%} &\textbf{4.74} &\textbf{0.3\%} &\textbf{0.3\%} &\textbf{12.5\%} &94.2 &37.2 &45.2 &0.58 \\ 
    \bottomrule
    \end{tabularx}
    \caption{\textbf{[Canonical set] Open-loop simulation results}}
    \label{tab:open-loop-canonical}
    \end{table*}

In general, we still observe the same overall trend seen in closed loop. Namely, that our method
is able to achieve better safety and compliance performance when compared to other
methods. We also see further evidence that occupancy-based methods are able to
produce safer driving patterns while simultaneously being more interpretable than other methods.

\subsection{Dataset Aggregation Iterations}

Our model attempts to mitigate distributional shift from open loop learning by
performing a Dagger-like \cite{ross2011reduction} aggregation of the dataset.
We start with a dataset generated by the expert planner, meaning the states visited are caused by actions from the expert.
Then, over multiple iterations we consider unrolling the simulation with our learned planner to learn to compensate our own mistakes by training to imitate the expert trajectory plan at every state. 
Intuitively, this process exposes the network to
out of distribution states that otherwise are only discovered during closed loop execution.
Here we conduct a brief investigation into the effect of this aggregation on the performance
of the model during closed loop evaluation, the results of which can be
seen in Table~\ref{tab:dagger_ablation}.

\begin{table*}[t]
    \footnotesize
    \centering                    
    \begin{tabularx}{\textwidth}{l sssssssssssss}\toprule
    &\textbf{Mission} &\multicolumn{7}{c}{\textbf{Safety and Compliance}} &\multicolumn{4}{c}{\textbf{Progress, Consistency and Comfort}}
    \\ \cmidrule(l{5pt}){2-2} \cmidrule(l{5pt}r{5pt}){3-9} \cmidrule(l{5pt}r{5pt}){10-13}
    &GSR\ua &ECR\da &PCR\da &\multicolumn{4}{c}{MinTTC\ua} &TVR\da &Progr.\ua &L2E\da &P2P\da &Jerk\da 
    \\\cmidrule(l{5pt}r{5pt}){5-8}
    & & & &$p10$ &$< 1s$ &$< 2s$ &$< 5s$ & & & & & \\\midrule
    \ourmodel{} w/o Dagger \cite{ross2011reduction} &81.1\% &\textbf{1.6\%} &2.5\% &4.35 &\textbf{1.1\%} &\textbf{1.1\%} &20.0\% &\textbf{7.9\%} &\textbf{433.5} &37.3 &\textbf{15.0} &\textbf{0.18} \\ 
    \ourmodel{} \textsc{1 Iteration} &\textbf{84.7\%} &2.1\% &\textbf{1.0\%} &\textbf{4.67} &1.6\% &1.6\% &14.2\% &8.4\% &430.6 &36.6 &15.6 &0.29 \\ 
    \ourmodel{} \textsc{2 Iterations} &84.2\% &\textbf{1.6\%} &2.3\% &4.45 &\textbf{1.1\%} &\textbf{1.1\%} &13.2\% &8.4\% &418.3 &35.7 &15.6 &0.27 \\ 
    \ourmodel{} \textsc{3 Iterations} &83.7\% &\textbf{1.6\%} &\textbf{1.0\%} &4.60 &\textbf{1.1\%} &\textbf{1.1\%} &\textbf{12.1\%} &\textbf{7.9\%} &417.3 &\textbf{34.8} &19.5 &0.27 \\
    \bottomrule
    \end{tabularx}
    \caption{\textbf{[Safety set] Dagger ablation}}
    \vspace{-10pt}
    \label{tab:dagger_ablation}
    \end{table*}

We observe that dataset aggregation does not necessarily increase the overall performance of the model after a certain point. 
There are some noticeable gains in the collision metrics after 1 iteration, but 
after this the improvements to performance gradually begin to wane.
Importantly, we highlight that the results of our model without any dataset aggregation still outperform the baselines.

\subsection{Training Dataset Composition}

One of the notable aspects of our method is our two stage training paradigm.
Namely, we train the backbone of $\ourmodel{}$ on the canonical set to generate a
reasonable set of occupancy predictions, and optimize the cost aggregation weights in the training safety set
so that we can have better safety guarantees. 
In this portion of the ablation we explore the effect of optimizing the cost aggregation weights
with different combinations of the safety and canonical set.
The results of these experiments can be seen
in Tables \ref{tab:safety-autotuning-ablation} and \ref{tab:canonical-autotuning-ablation}.
Similar to TRAVL \cite{zhang2022rethinking}, we observe that training on safety critical scenarios yields better evaluation performance on both safety critical scenarios as well as nominal cases.

\begin{table*}[t]
    \footnotesize
    \centering                    
    \begin{tabularx}{\textwidth}{l sssssssssssss}\toprule
    &\textbf{Mission} &\multicolumn{7}{c}{\textbf{Safety and Compliance}} &\multicolumn{4}{c}{\textbf{Progress, Consistency and Comfort}} 
    \\ \cmidrule(l{5pt}){2-2} \cmidrule(l{5pt}r{5pt}){3-9} \cmidrule(l{5pt}r{5pt}){10-13}
    &GSR\ua &ECR\da &PCR\da &\multicolumn{4}{c}{MinTTC\ua} &TVR\da &Progr.\ua &L2E\da &P2P\da &Jerk\da 
    \\\cmidrule(l{5pt}r{5pt}){5-8}
    & & & &$p10$ &$< 1s$ &$< 2s$ &$< 5s$ & & & & & \\\midrule
    \ourmodel{} \textsc{100\% safety} &\textbf{81.1\%} &\textbf{1.6\%} &\textbf{2.5\%} &\textbf{4.35} &\textbf{1.1\%} &\textbf{1.1\%} &\textbf{20.0\%} &\textbf{7.9\%} &\textbf{433.5} &\textbf{37.3} &15.0 &0.18 \\ 
    \ourmodel{} \textsc{50\% / 50\% } &75.3\% &8.9\% &13.0\% &3.00 &6.8\% &6.8\% &28.9\% &14.7\% &381.1 &40.9 &13.6 &0.12 \\ 
    \ourmodel{} \textsc{100\% canonical} &55.3\% &36.3\% &36.8\% &0.40 &29.5\% &33.2\% &42.6\% &40.0\% &310.9 &91.2 &\textbf{13.2} &\textbf{0.09} \\ 
    \bottomrule
    \end{tabularx}
    \caption{\textbf{[Safety set] Ablating the dataset for cost aggregation learning}}
    \label{tab:safety-autotuning-ablation}
\end{table*}

\begin{table*}[t]
    \footnotesize
    \centering                    
    \begin{tabularx}{\textwidth}{l ssssssssssss}\toprule
    &\multicolumn{7}{c}{\textbf{Safety and Compliance}} &\multicolumn{4}{c}{\textbf{Progress, Consistency and Comfort}} 
    \\ \cmidrule(l{5pt}r{5pt}){2-8} \cmidrule(l{5pt}r{5pt}){9-12}
    &ECR\da &PCR\da &\multicolumn{4}{c}{MinTTC\ua} &TVR\da &Progr.\ua &L2E\da &P2P\da &Jerk\da 
    \\\cmidrule(l{5pt}r{5pt}){4-7}
    & & &$p10$ &$< 1s$ &$< 2s$ &$< 5s$ & & & & & \\\midrule
    \ourmodel{} \textsc{100\% safety} &\textbf{0.0\%} &\textbf{0.5\%} &\textbf{4.90} &\textbf{0.3\%} &\textbf{0.6\%} &\textbf{10.9\%} &\textbf{1.9\%} &\textbf{303.3} &36.6 &23.6 &0.28 \\ 
    \ourmodel{} \textsc{50\% / 50\%} &2.2\% &1.3\% &4.66 &2.2\% &2.6\% &13.1\% &2.9\% &302.9 &\textbf{35.9} &9.6 &0.14 \\ 
    \ourmodel{} \textsc{100\% canonical} &21.4\% &23.5\% &0.40 &18.5\% &19.5\% &27.5\% &22.0\% &284.6 &44.1 &\textbf{7.9} &\textbf{0.09} \\ 
    \bottomrule
    \end{tabularx}
    \caption{\textbf{[Canonical set] Ablating the dataset for cost aggregation learning}}
    \vspace{-10pt}
    \label{tab:canonical-autotuning-ablation}
\end{table*}

\subsection{Sub Cost Ablation} 
Table~\ref{tab:sub-cost-ablation}, analyzes the effect of a cost being removed. 
Each row is named after the dropped cost.

\begin{table}[h]
    \scriptsize
    \vspace{-8pt}
    \centering
    \begin{tabular}{lrrrrrrr}\toprule
    &GSR\ua &ECR\da &TTC$< 5s$\da &TVR\da &Progr.\ua &Jerk\da \\
    \midrule 
    \textsc{collision}      &\color{red}{-6.4}\%        &\color{red}{5.3}\%          &\color{red}{10.8}\%         &\color{red}{7.5}\% &\color{ForestGreen}{9.3} &\color{ForestGreen}{-0.05} \\ 
    \textsc{buffer}         &\color{ForestGreen}{1.1}\% &\color{ForestGreen}{-0.6}\% &\color{red}{1.1}\%          &{0.0}\% &\color{ForestGreen}{3.8} &\color{ForestGreen}{-0.02} \\ 
    \textsc{comfort}        &\color{ForestGreen}{1.1}\% &\color{ForestGreen}{-1.1}\% &\color{ForestGreen}{-0.5}\% &\color{red}{1.6}\% &\color{ForestGreen}{12.9} &\color{red}{0.42} \\ 
    \textsc{corridor}       &\color{red}{-1.6}\%        &\color{ForestGreen}{-0.6}\% &\color{ForestGreen}{-1.0}\% &{0.0}\% &\color{ForestGreen}{5.9} &\color{ForestGreen}{-0.03} \\ 
    \textsc{boundary}       &\color{red}{-2.7}\%        &{0.0}\%                     &\color{red}{1.6}\%          &\color{red}{3.8}\% &\color{red}{-6.0} &{0.00} \\ 
    \textsc{speed limit}    &\color{ForestGreen}{1.1}\% &{0.0}\%                     &\color{ForestGreen}{-0.5}\% &{0.0}\% &\color{ForestGreen}{4.3} &\color{ForestGreen}{-0.01} \\ 
    \textsc{progress}       &\color{red}{-15.0}\%       &\color{ForestGreen}{-0.6}\% &{0.0}\%                     &\color{red}{2.7}\% &\color{red}{-113.9} &\color{ForestGreen}{-0.06} \\ 
    \textsc{route}          &\color{red}{-25.8}\%       &\color{ForestGreen}{-0.6}\% &\color{ForestGreen}{-5.3}\% &\color{red}{11.3}\% &\color{ForestGreen}{84.8} &\color{ForestGreen}{-0.03} \\
    \bottomrule
    \end{tabular}
    \caption{\textbf{[Safety set] Sub Cost Ablation}}
    \vspace{-12pt}
    \label{tab:sub-cost-ablation}
\end{table}

The results are intuitive: no collision cost yields the worst ECR, no comfort cost yields the worst jerk, no progress cost yields the worst progress, no route cost yields the worst GSR.

\subsection{Query Sample Quantization}

$\ourmodel{}$ queries uniformly over a motion blurred bounding box to estimate the occupancy of
the targeted regions of interest. Specifically, these regions correspond to the areas
the self-driving vehicle could occupy at some future point in time. This point
sampling effectively discretizes the continuous occupancy problem. We note that the density of these 
points in the motion bounding box can be tuned to achieve a different runtime to precision 
ratio. In Table \ref{tab:resolution-ablation} we explore the effect point sampling density has on the 
final close loop performance.

\begin{table*}[t]
    \footnotesize
    \centering                    
    \begin{tabularx}{\textwidth}{l ssssssssssssss}\toprule
    &\textbf{Mission} 
    &\multicolumn{7}{c}{\textbf{Safety and Compliance}} 
    &\multicolumn{4}{c}{\textbf{Progress, Consistency and Comfort}} 
    &\textbf{Latency} 
    \\ \cmidrule(l{5pt}){2-2} \cmidrule(l{5pt}r{5pt}){3-9} \cmidrule(l{5pt}r{5pt}){10-13} \cmidrule(l{5pt}r{5pt}){14-14}
    &GSR\ua &ECR\da &PCR\da &\multicolumn{4}{c}{MinTTC\ua} &TVR\da &Progr.\ua &L2E\da &P2P\da &Jerk\da &Runtime\da 
    \\\cmidrule(l{5pt}r{5pt}){5-8}
    & & & &$p10$ &$< 1s$ &$< 2s$ &$< 5s$ & & & & & & (ms)\\\midrule
    \ourmodel{} \textsc{continuous points} &82.9\% &\textbf{1.0\%} &\textbf{0.4\%} &4.89 &\textbf{1.0\%} &\textbf{1.0\%} &11.6\% &11.1\% &\textbf{444.9} &\textbf{35.8} &15.4 &\textbf{0.28} &705 \\
    \ourmodel{} \textsc{0.1m} &83.9\% &\textbf{1.0\%} &\textbf{0.4\%} &\textbf{4.90} &\textbf{1.0\%} &\textbf{1.0\%} &\textbf{11.1\%} &10.1\% &443.4 &37.8 &15.7 &\textbf{0.28} &293 \\
    \ourmodel{} \textsc{0.25m} &\textbf{85.9\%} &\textbf{1.0\%} &0.5\% &4.89 &\textbf{1.0\%} &\textbf{1.0\%} &12.1\% &\textbf{8.0\%} &441.8 &37.8 &\textbf{15.3} &\textbf{0.28} &259 \\
    \ourmodel{} \textsc{0.5m (default)} &84.9\% &2.0\% &1.0\% &4.83 &1.5\% &1.5\% &13.6\% &9.0\% &436.7 &37.8 &15.6 &0.29 &256 \\
    \ourmodel{} \textsc{1m} &83.9\% &1.5\% &1.0\% &4.72 &\textbf{1.0\%} &1.5\% &14.1\% &10.6\% &442.5 &36.0 &19.8 &\textbf{0.28} &\textbf{254} \\
    \ourmodel{} \textsc{2m} &84.4\% &3.0\% &2.0\% &4.45 &2.5\% &2.5\% &14.6\% &9.0\% &436.5 &38.1 &22.9 &0.29 &257 \\
    \bottomrule
    \end{tabularx}
    \caption{\textbf{[Safety set] Query point quantization ablation}}
    \label{tab:resolution-ablation}
    \end{table*}

We find that having a finer
discretization does improve results, but after a certain point the improvements seem to be
marginal while coming at the expense of additional runtime. 
Thus, we selected 0.5 m resolution for our final model.
This is to be expected since the actors in the scene are fairly large, and
further refinement wouldn't be necessary to identify their location during the costing
process.

\section{Qualitative Results}
\label{sec:add_qualitative}

\begin{figure*}[t] %
    \centering
    \begin{tabularx} {\textwidth} {l | X X X} %
        & t = start & t = middle & t = end\\
        \midrule
        \raisebox{3.2\height}{\textsc{$\ourmodel{}$}} & \includegraphics[trim=1100 300 0 300,clip,width=0.6\columnwidth]{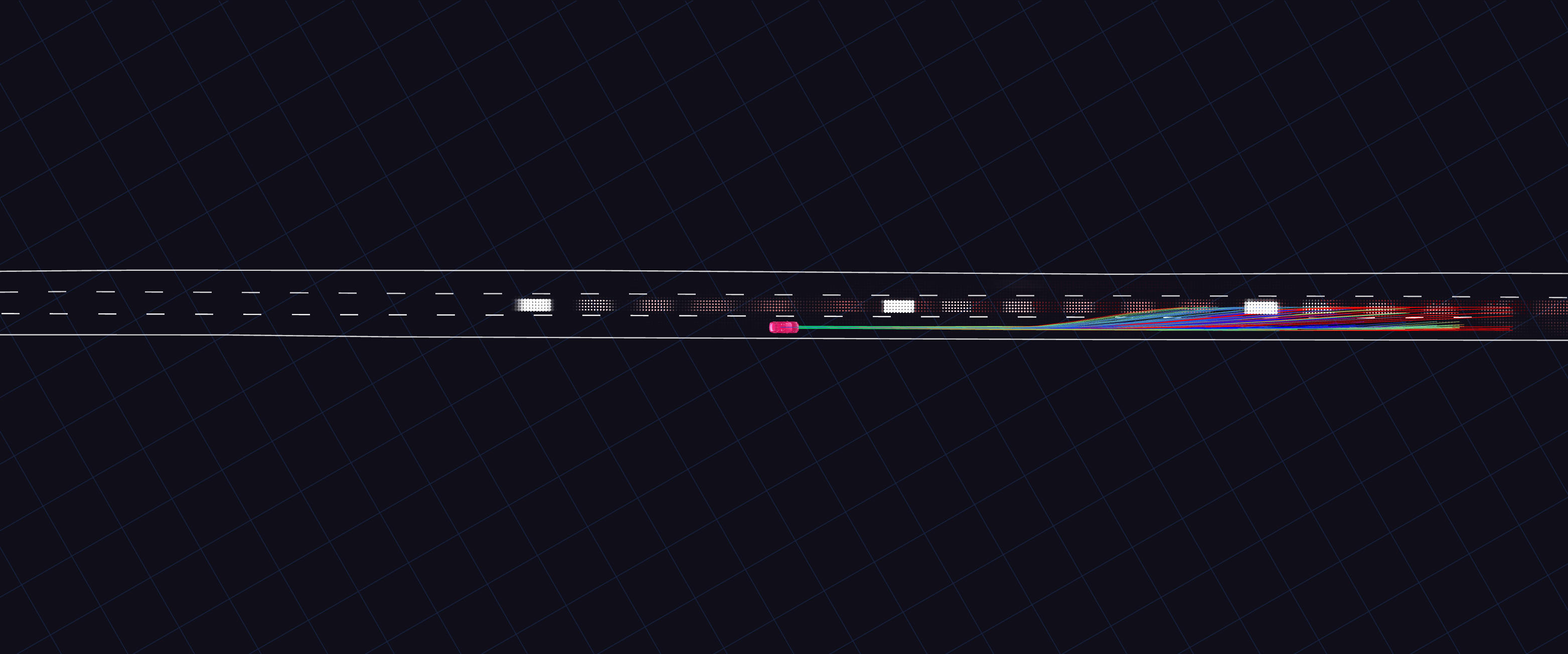} &
        \includegraphics[trim=1100 300 0 300,clip,width=0.6\columnwidth]{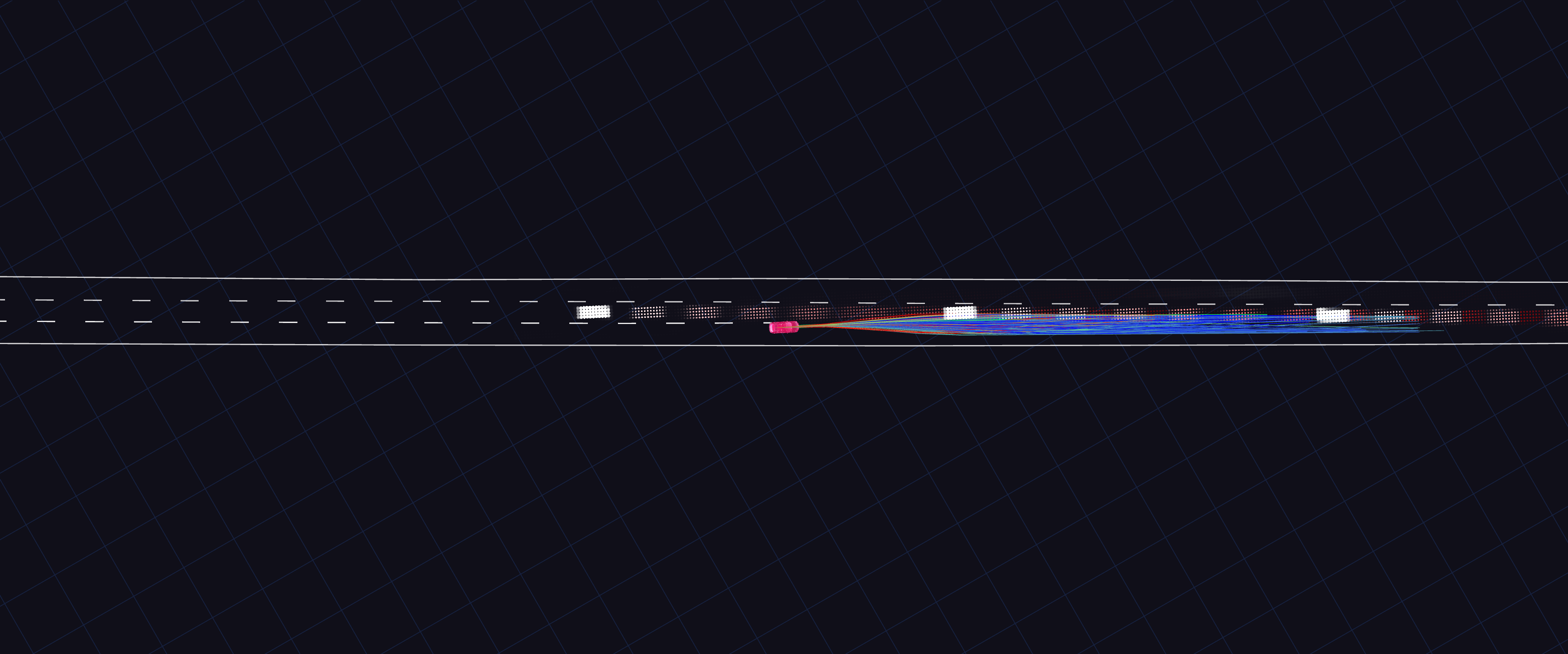} &
        \includegraphics[trim=1100 300 0 300,clip,width=0.6\columnwidth]{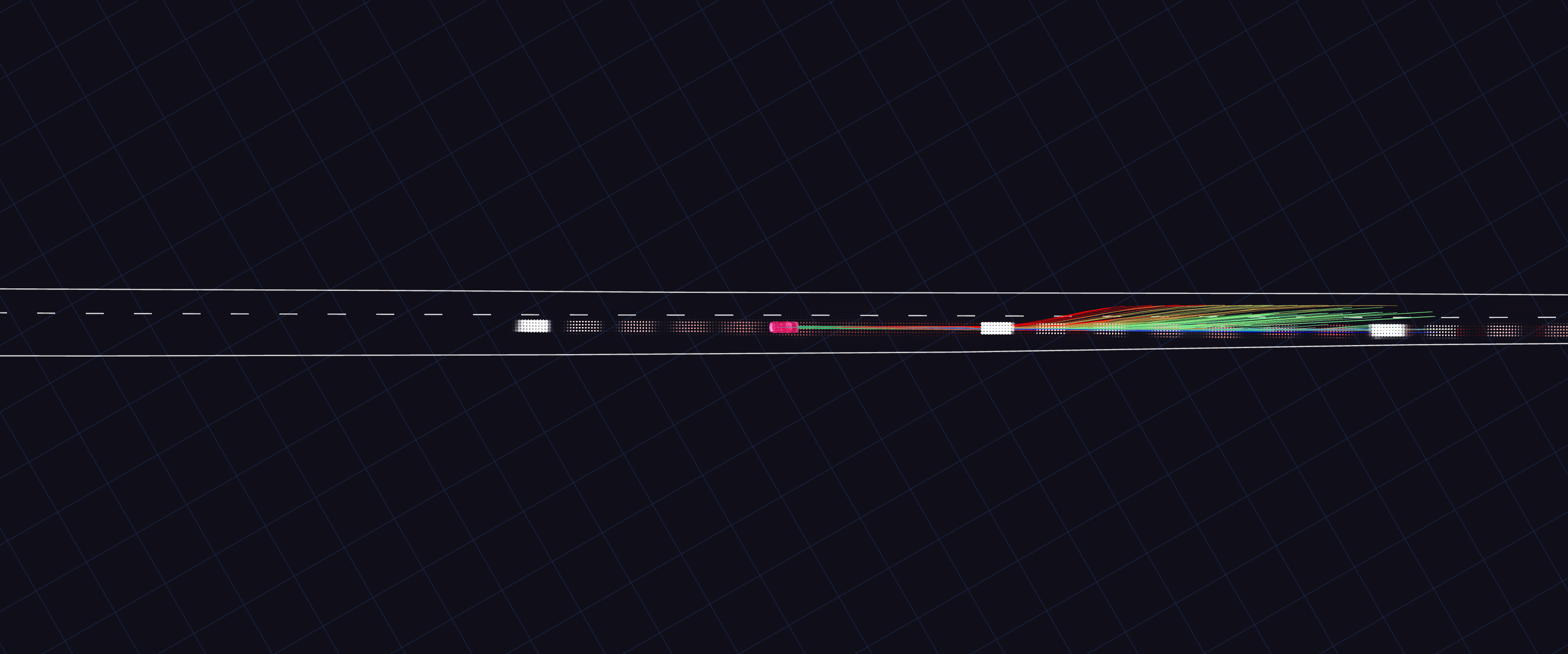} \\

        \raisebox{3.2\height}{\textsc{P3}} & \includegraphics[trim=1100 300 0 300,clip,width=0.6\columnwidth]{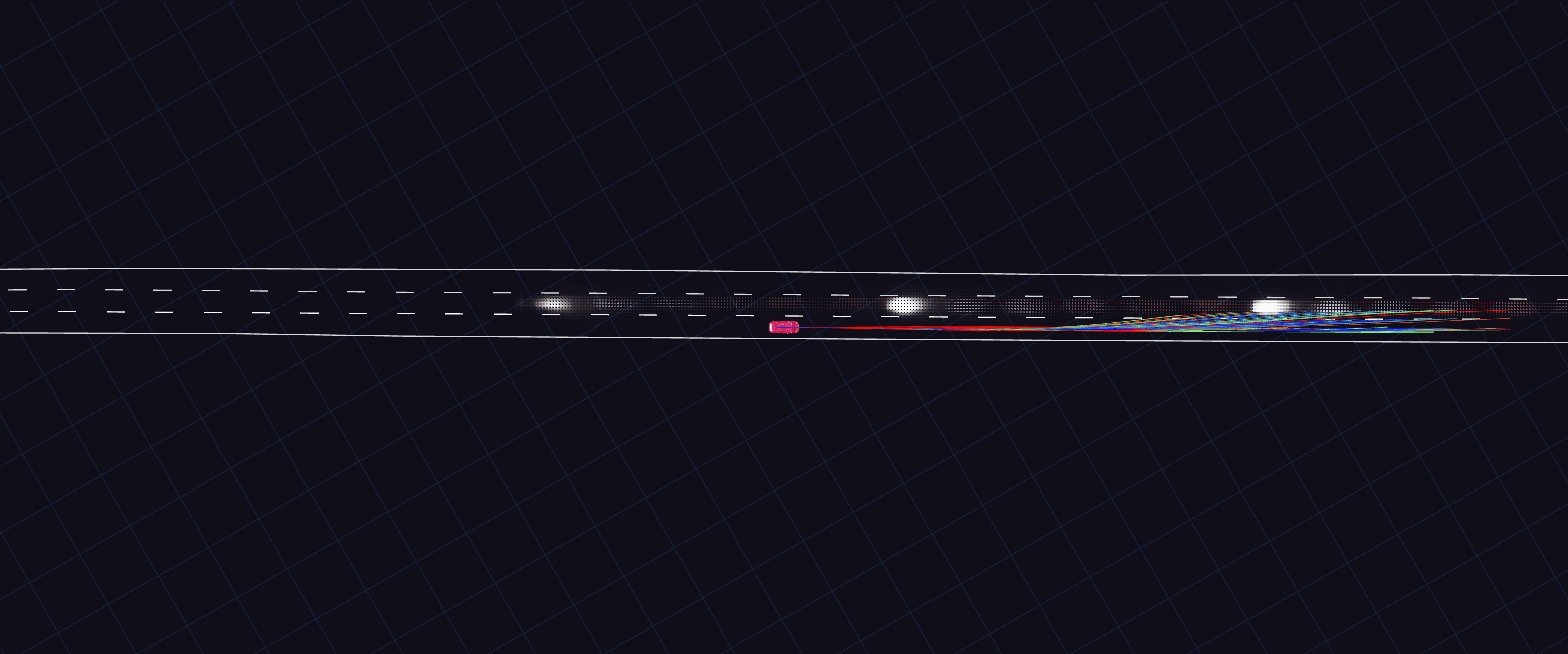} &
        \includegraphics[trim=1100 300 0 300,clip,width=0.6\columnwidth]{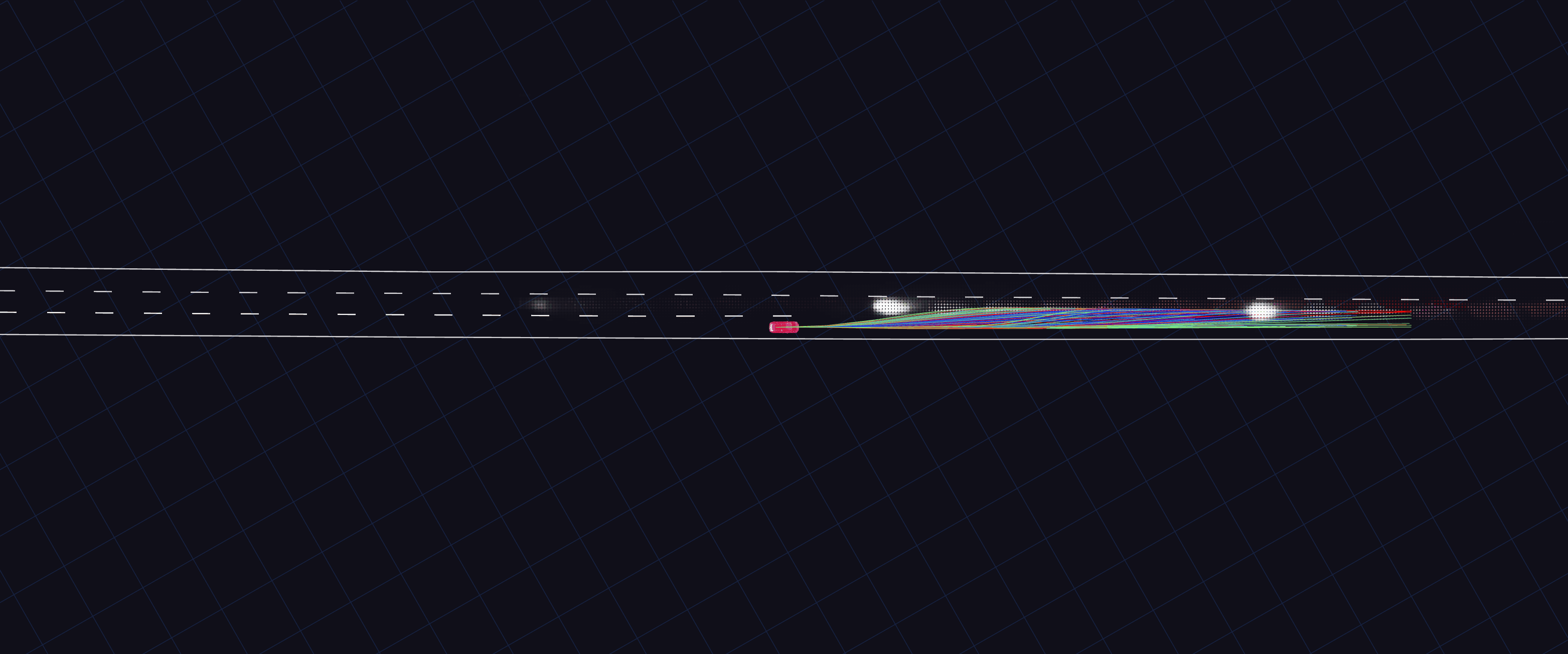} &
        \includegraphics[trim=1100 300 0 300,clip,width=0.6\columnwidth]{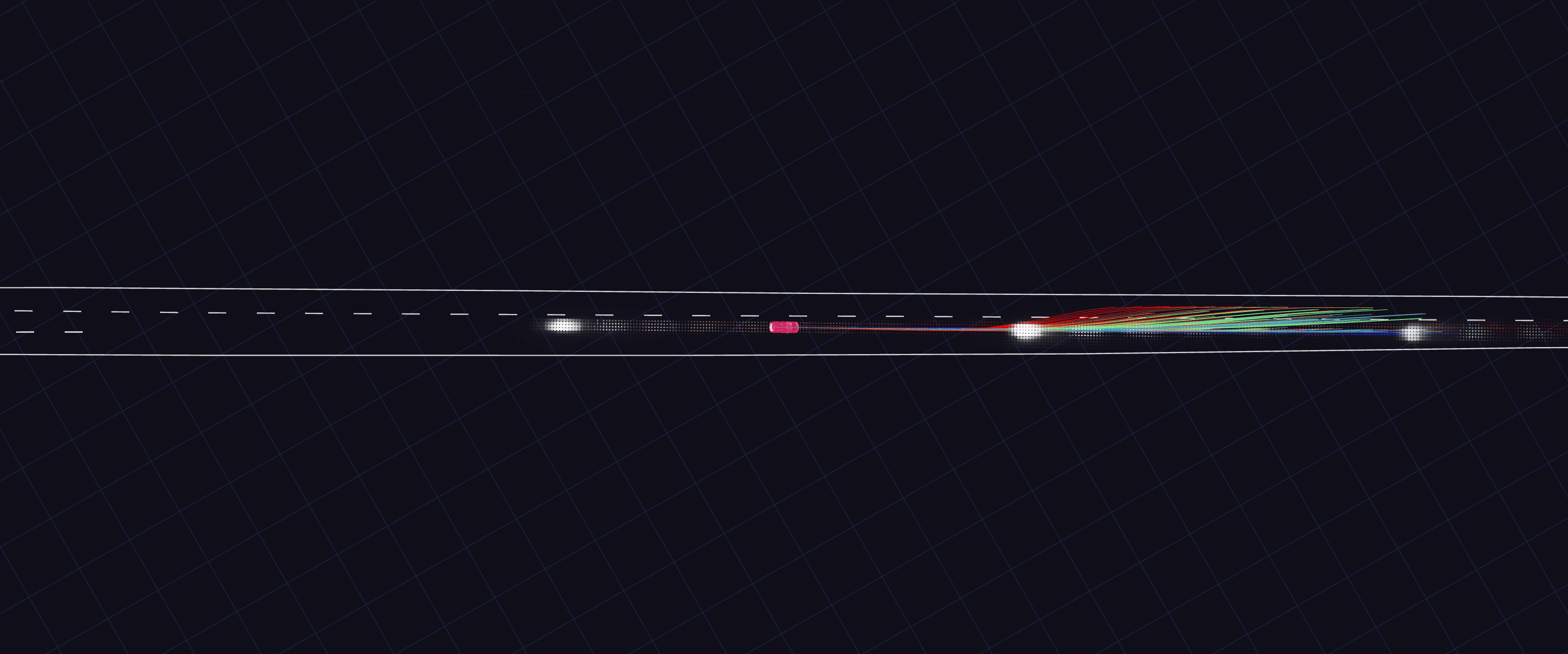} \\

        \raisebox{3.2\height}{\textsc{OccFlow}} & \includegraphics[trim=1100 300 0 300,clip,width=0.6\columnwidth]{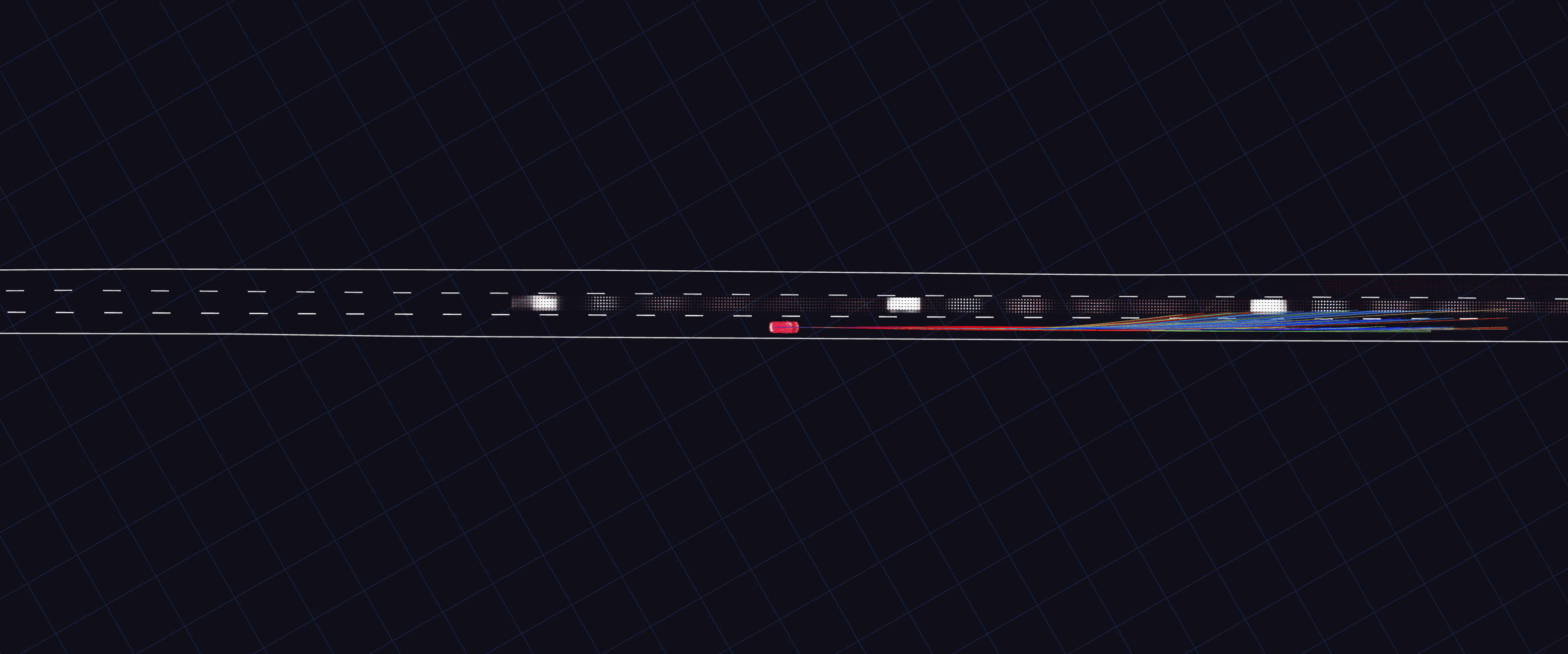} &
        \includegraphics[trim=1100 300 0 300,clip,width=0.6\columnwidth]{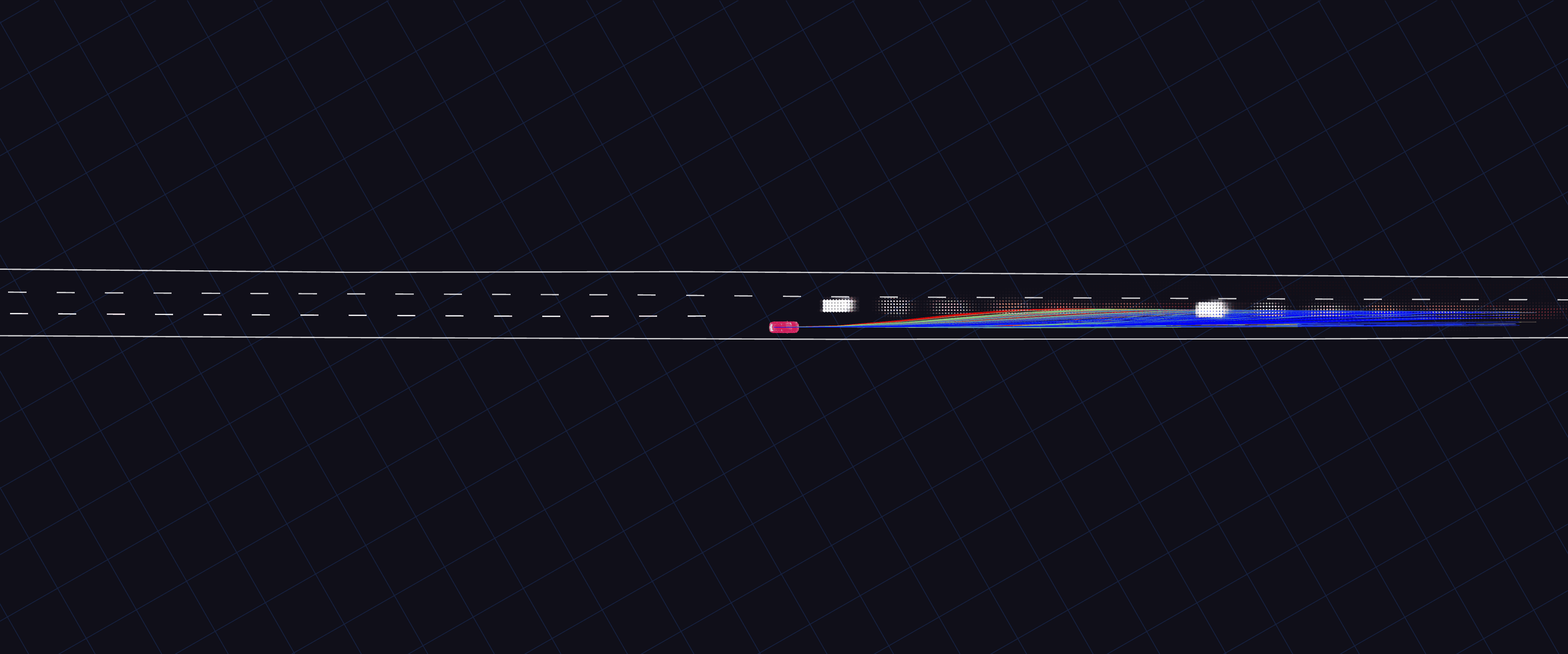} &
        \includegraphics[trim=1100 300 0 300,clip,width=0.6\columnwidth]{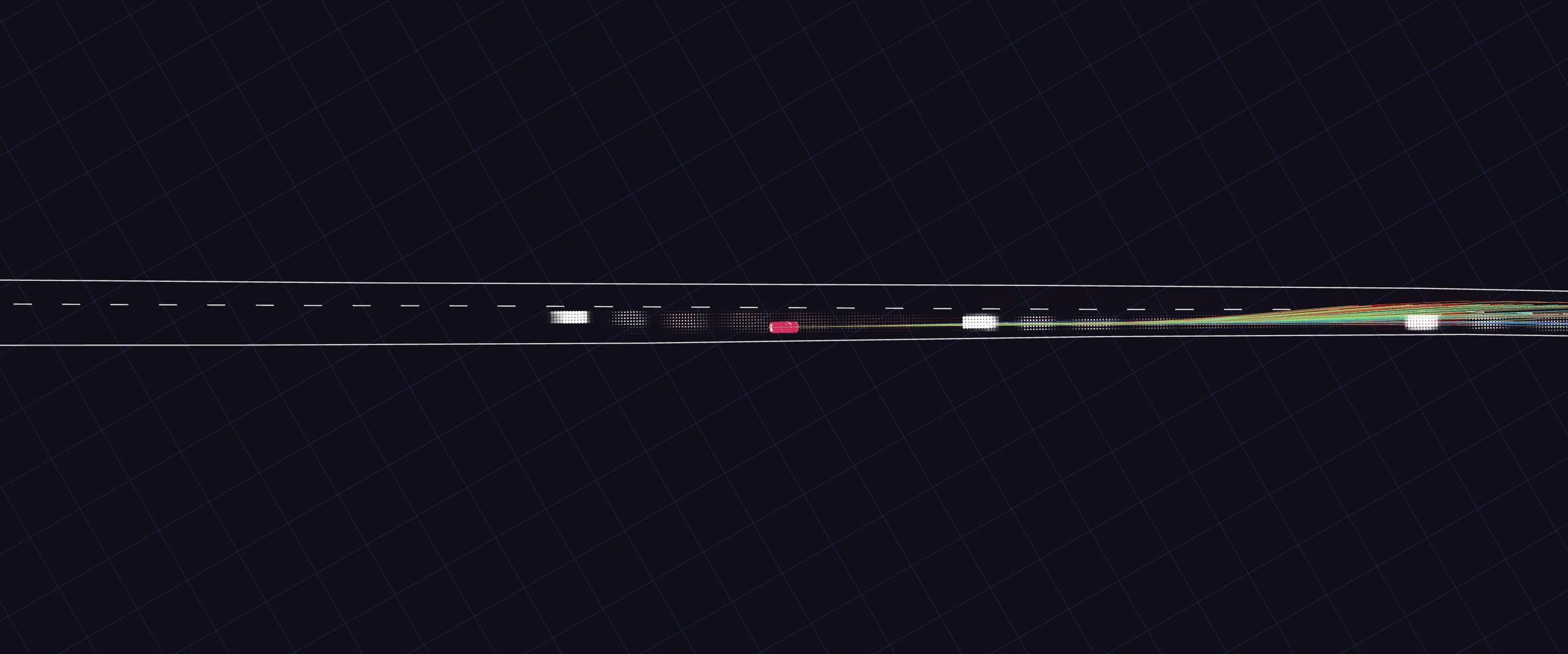} \\

        \raisebox{3.2\height}{\textsc{NMP}} & \includegraphics[trim=1100 300 0 300,clip,width=0.6\columnwidth]{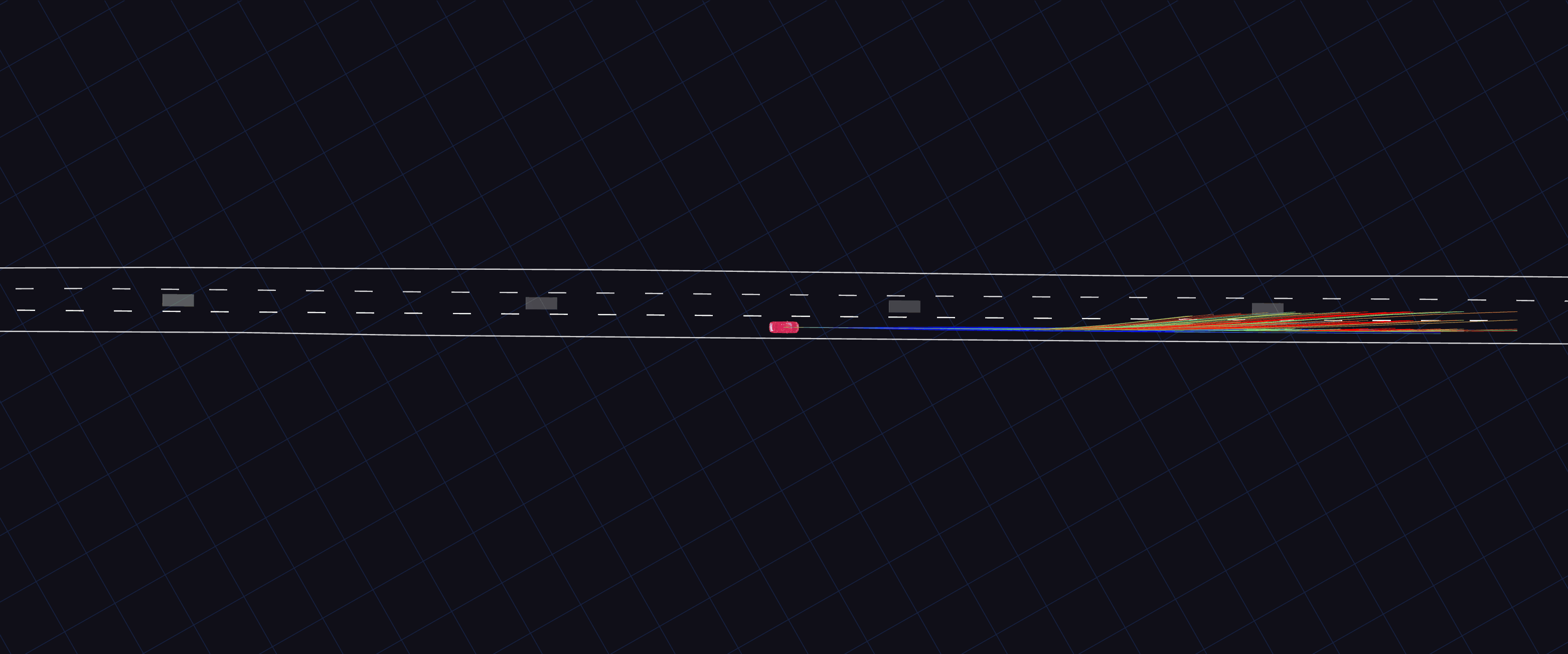} &
        \includegraphics[trim=1100 300 0 300,clip,width=0.6\columnwidth]{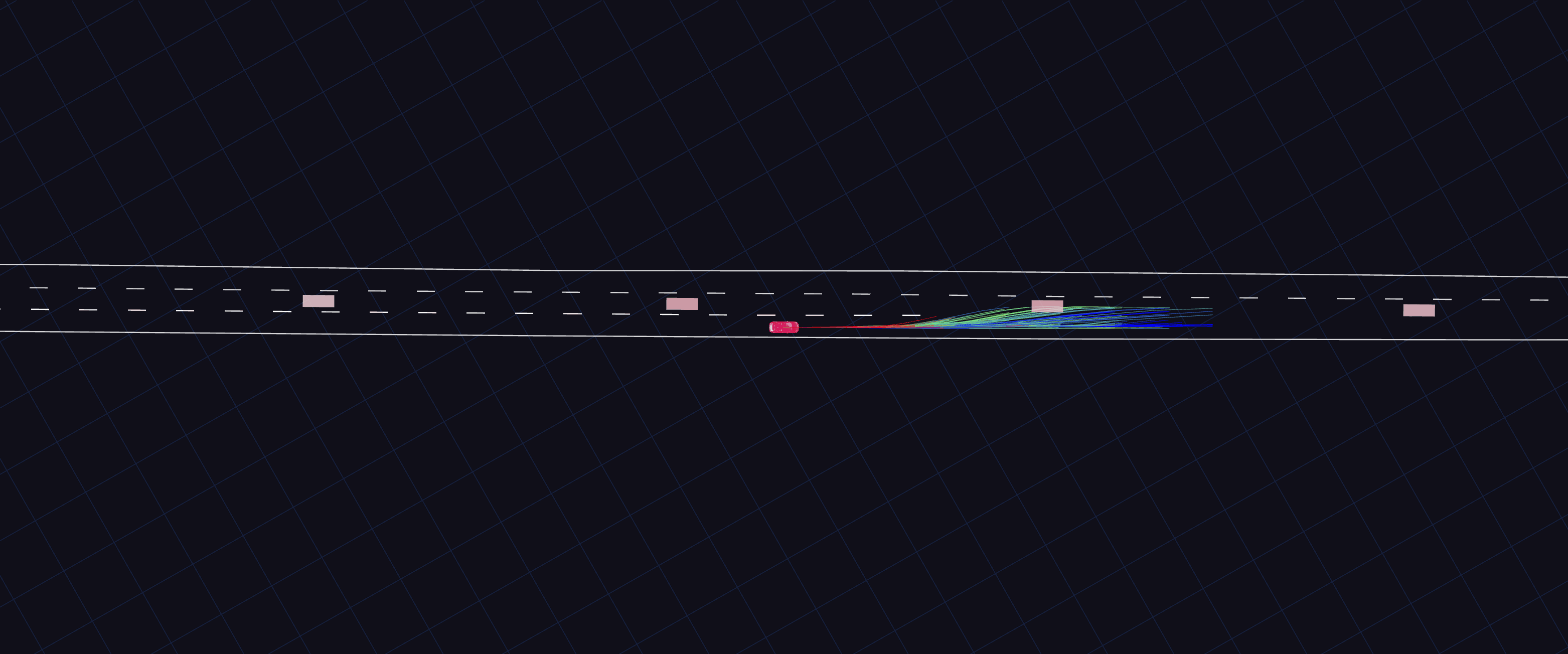} &
        \includegraphics[trim=1100 300 0 300,clip,width=0.6\columnwidth]{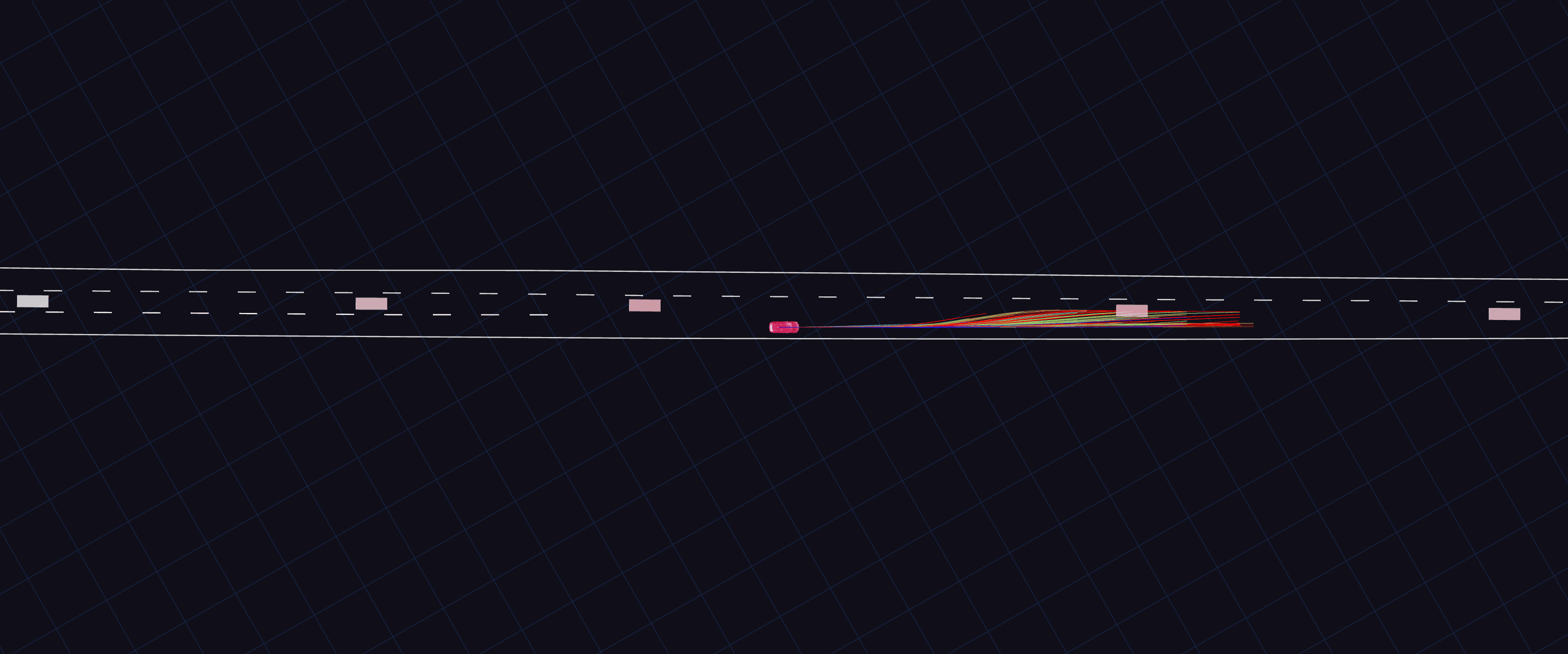} \\

        \raisebox{3.2\height}{\textsc{PlanT}} & \includegraphics[trim=1100 300 0 300,clip,width=0.6\columnwidth]{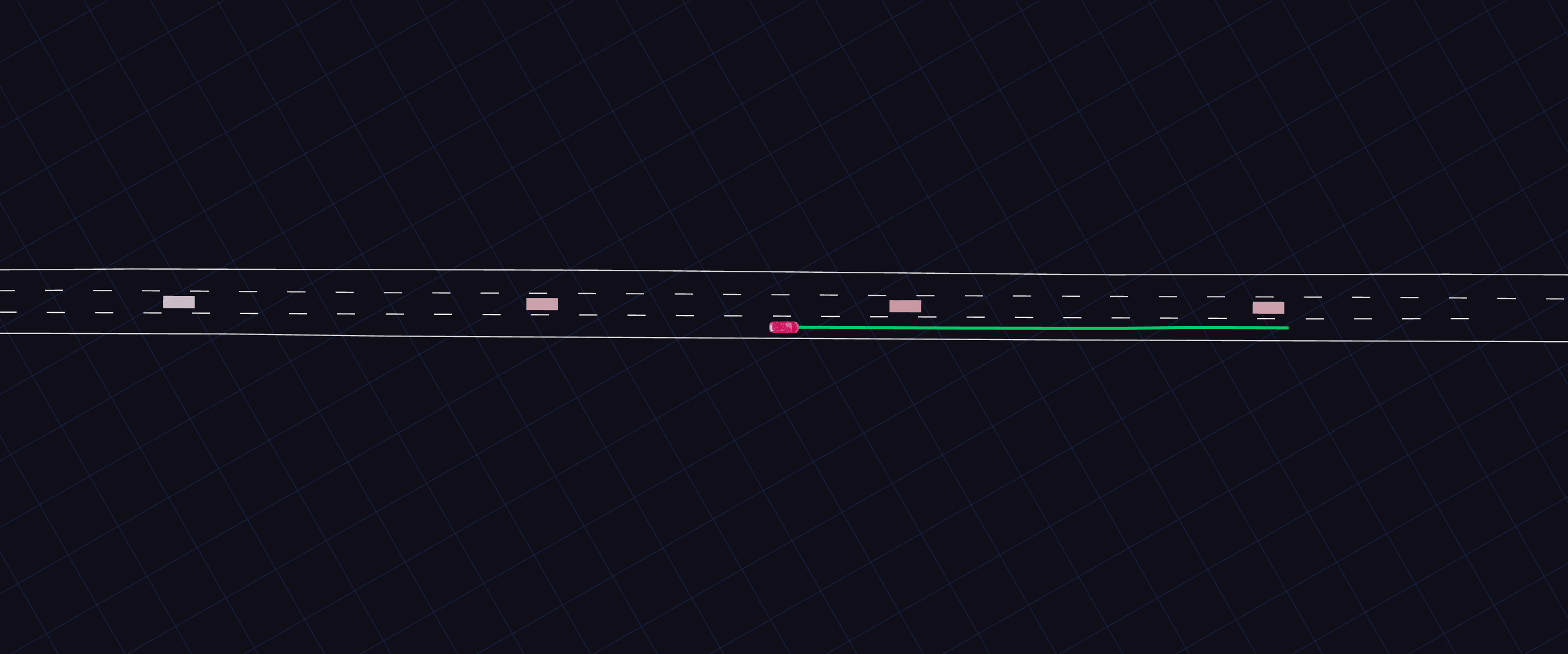} &
        \includegraphics[trim=1100 300 0 300,clip,width=0.6\columnwidth]{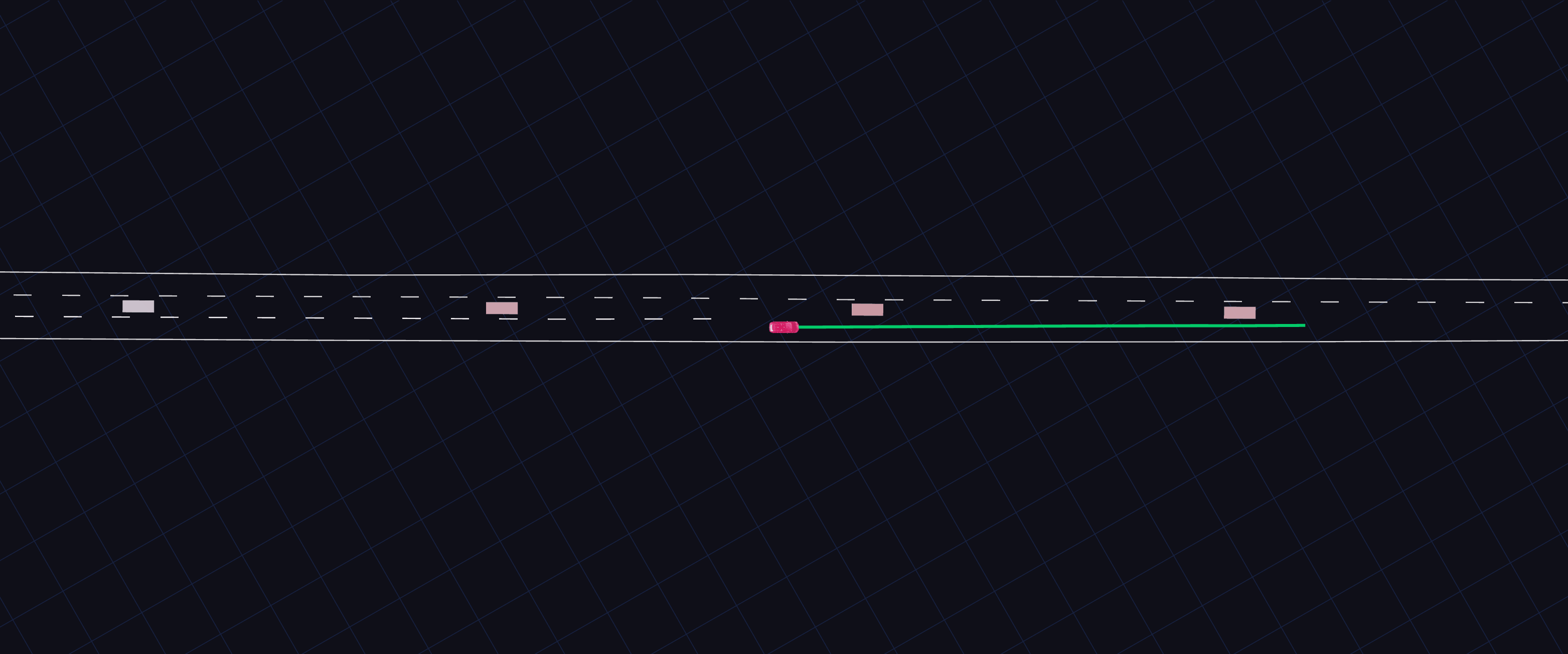} &
        \includegraphics[trim=1100 300 0 300,clip,width=0.6\columnwidth]{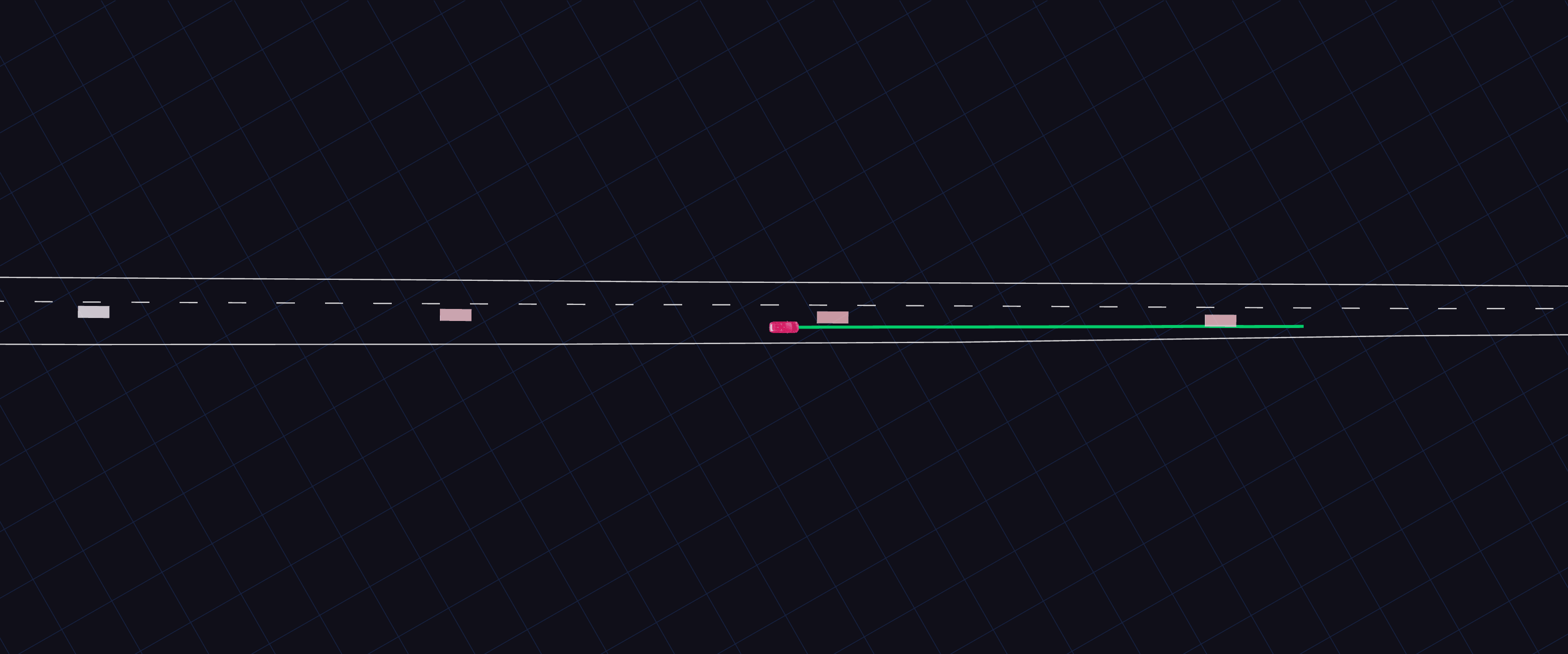} \\
    \end{tabularx}
    \caption{\textbf{Safety Scenario Comparison 3} Here we have a scenario where the SDV must merge into 
    traffic filled lane before the current lane ends. While $\ourmodel{}$ and P3 are able to efficiently 
    perform the lane change, the other models struggle to do it in before their current lane comes to an end.}
    \vspace{-10pt}
    \label{fig:scene_3_qual}
\end{figure*}

In the following figures \ref{fig:scene_1_qual}, \ref{fig:scene_2_qual}, \ref{fig:scene_3_qual} we have the additional qualitative results comparing the performance 
of our model in safety critical scenarios to other baselines. Note that in each of the visuals, 
we omit the lidar point clouds for clarity.

\begin{figure*}[t] %
    \centering
    \begin{tabularx} {\textwidth} {l | X X X} %
        & t = start & t = middle & t = end\\
        \midrule
        \raisebox{3.2\height}{\textsc{$\ourmodel{}$}} & \includegraphics[trim=1100 300 0 300,clip,width=0.6\columnwidth]{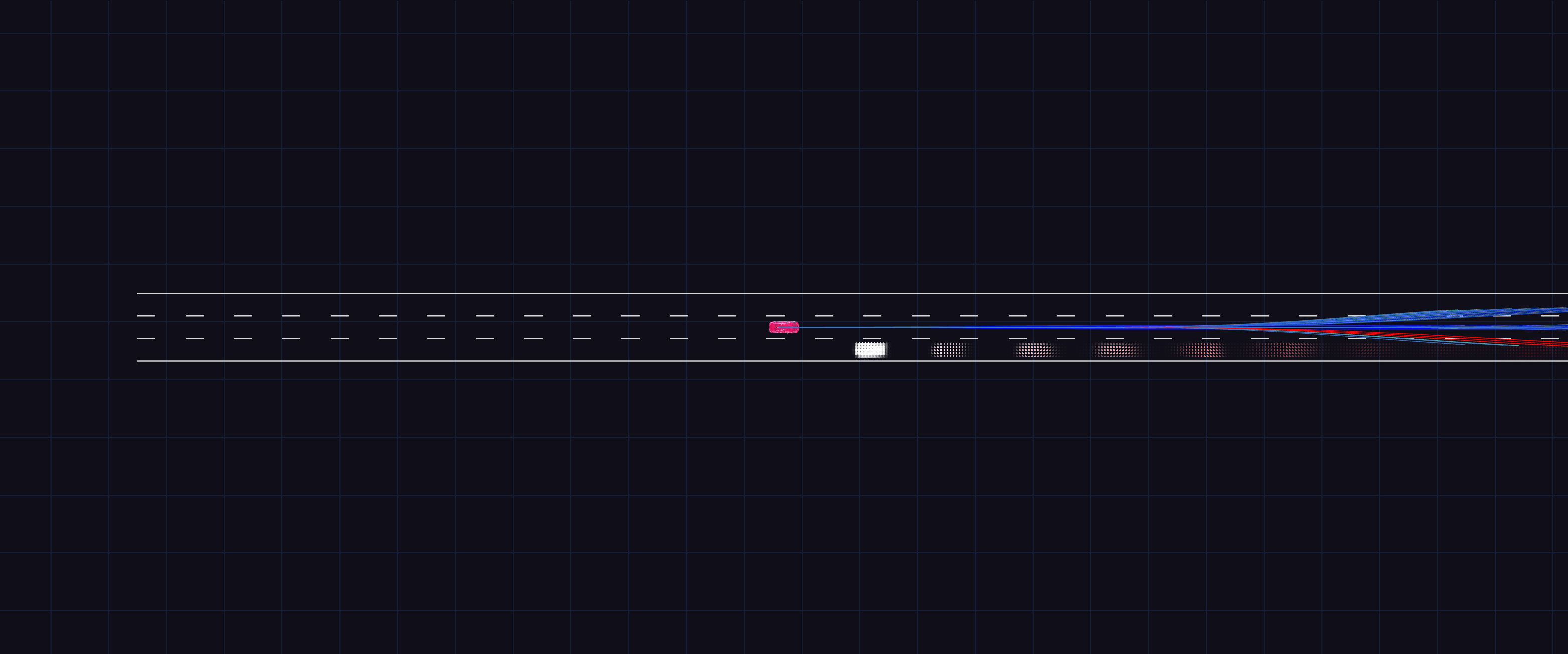} &
        \includegraphics[trim=1100 300 0 300,clip,width=0.6\columnwidth]{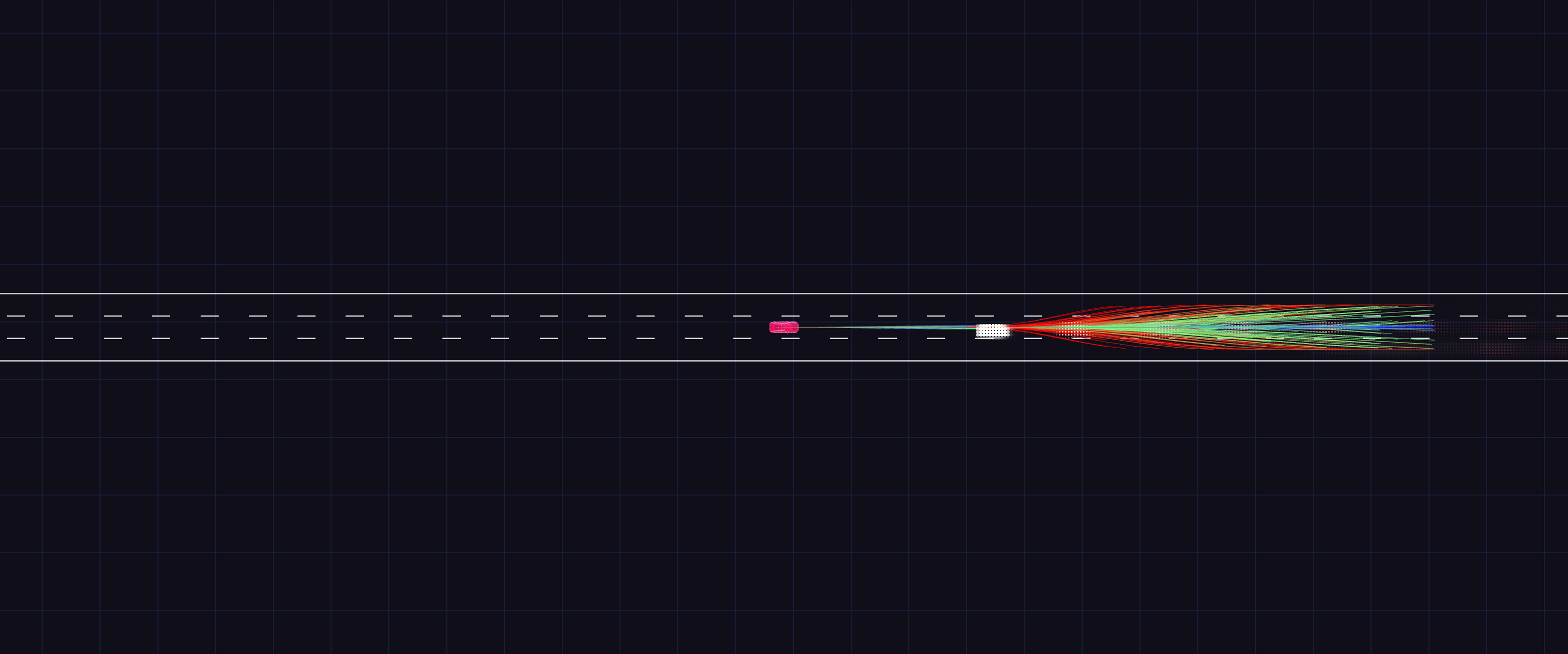} &
        \includegraphics[trim=1100 300 0 300,clip,width=0.6\columnwidth]{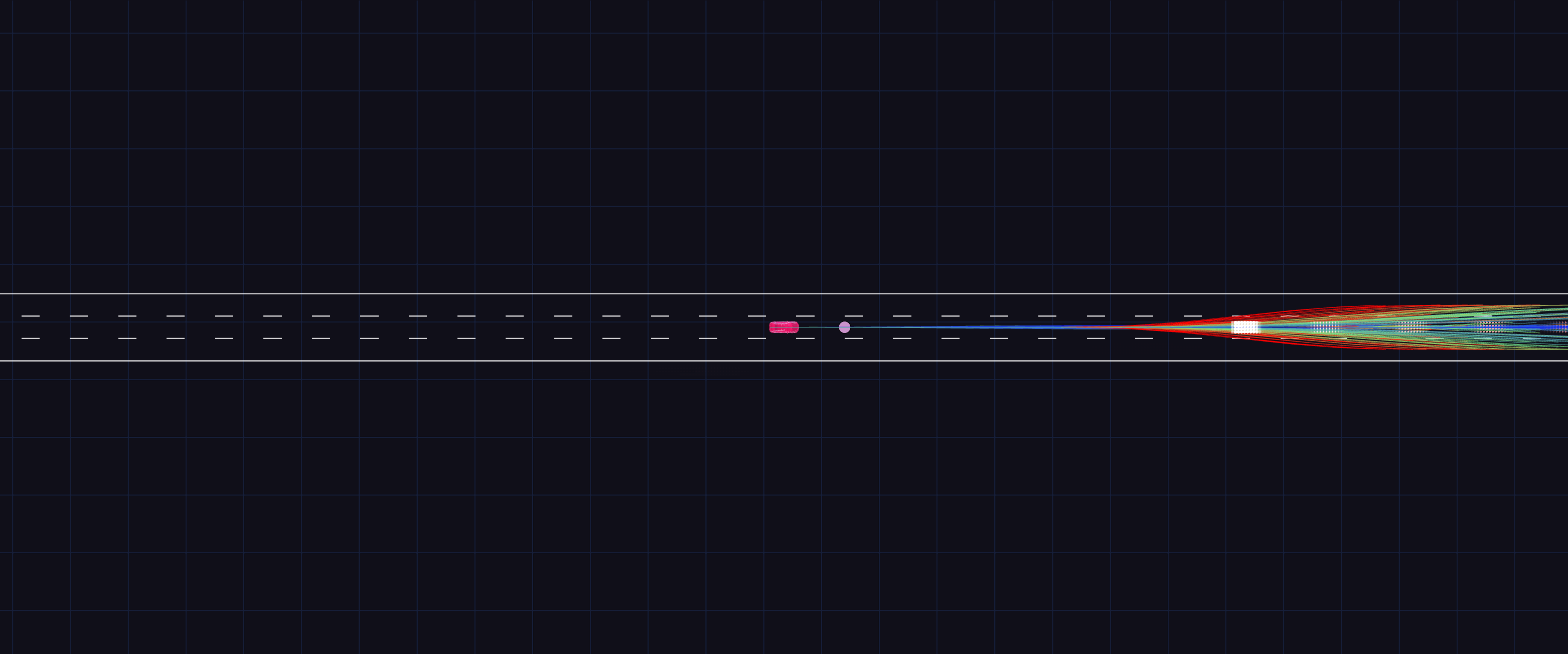} \\

        \raisebox{3.2\height}{\textsc{P3}} & \includegraphics[trim=1100 300 0 300,clip,width=0.6\columnwidth]{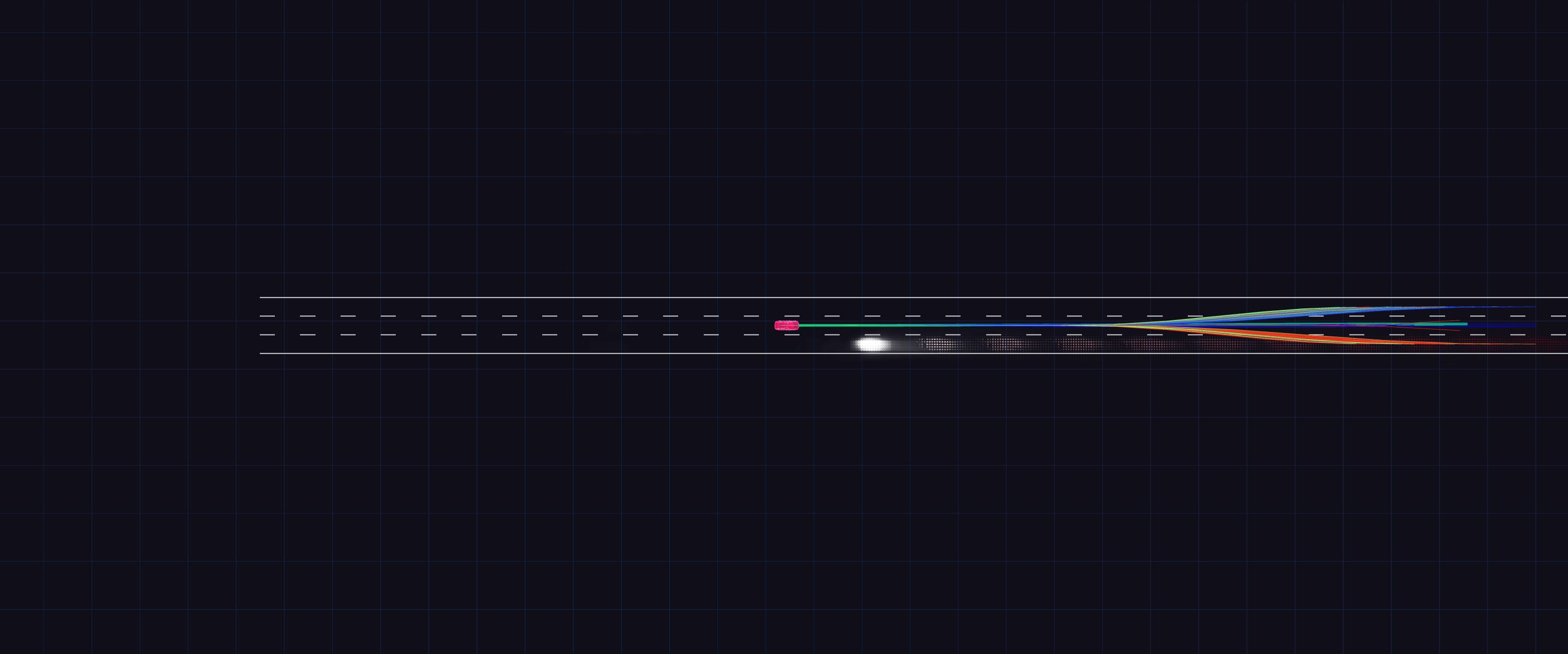} &
        \includegraphics[trim=1100 300 0 300,clip,width=0.6\columnwidth]{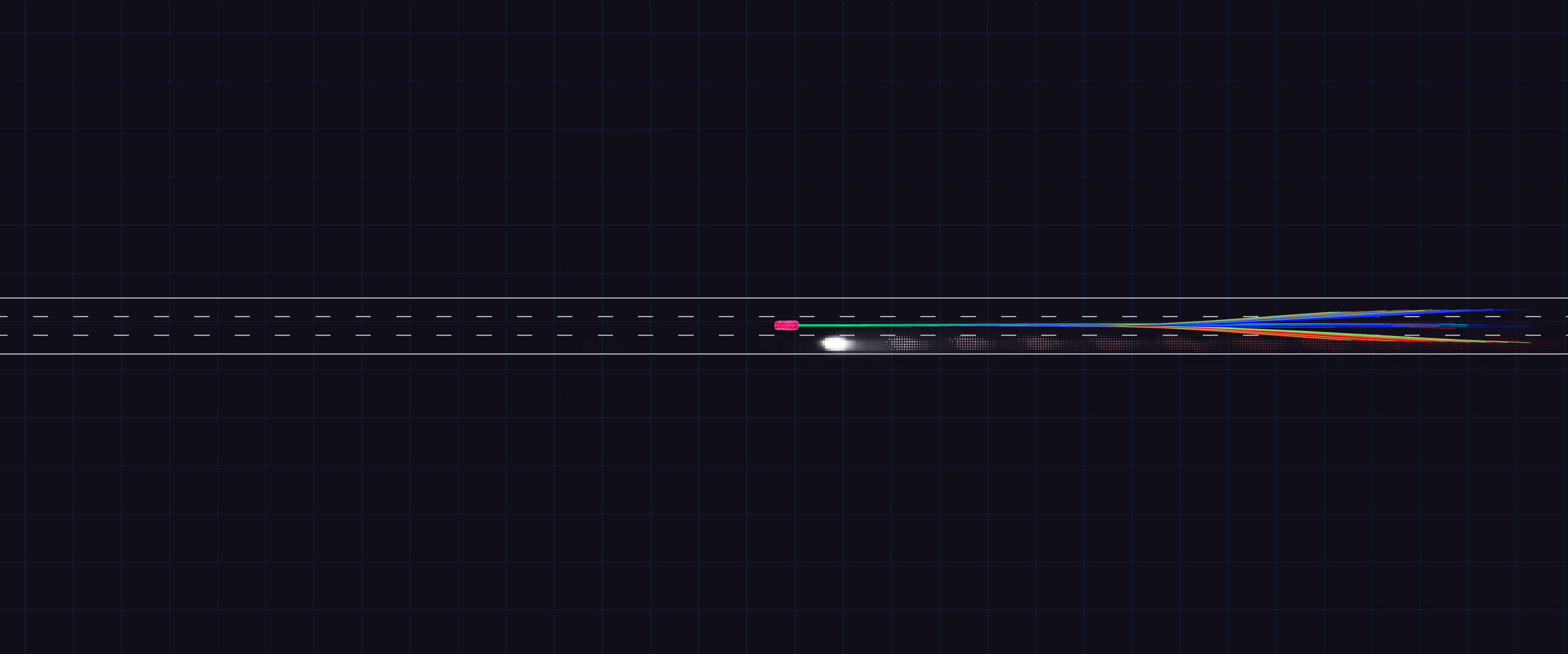} &
        \includegraphics[trim=1100 300 0 300,clip,width=0.6\columnwidth]{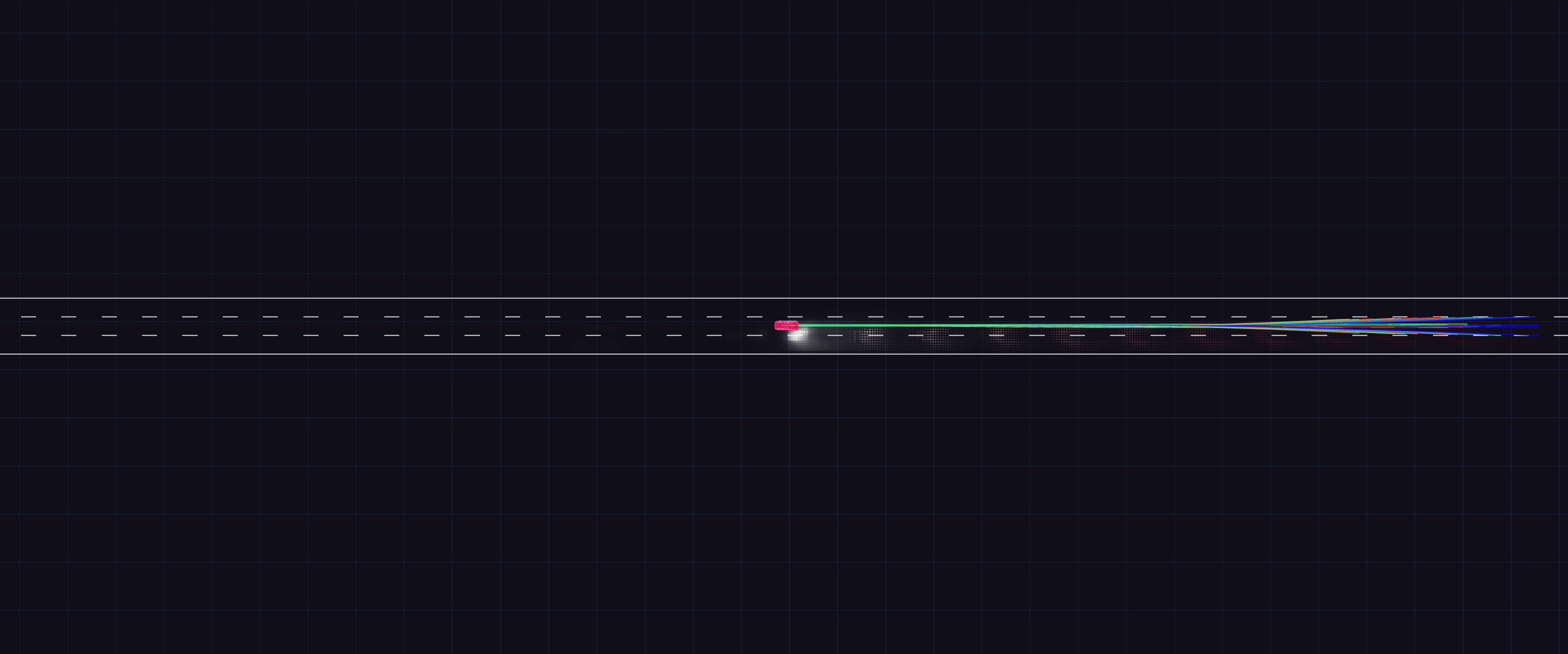} \\

        \raisebox{3.2\height}{\textsc{OccFlow}} & \includegraphics[trim=1100 300 0 300,clip,width=0.6\columnwidth]{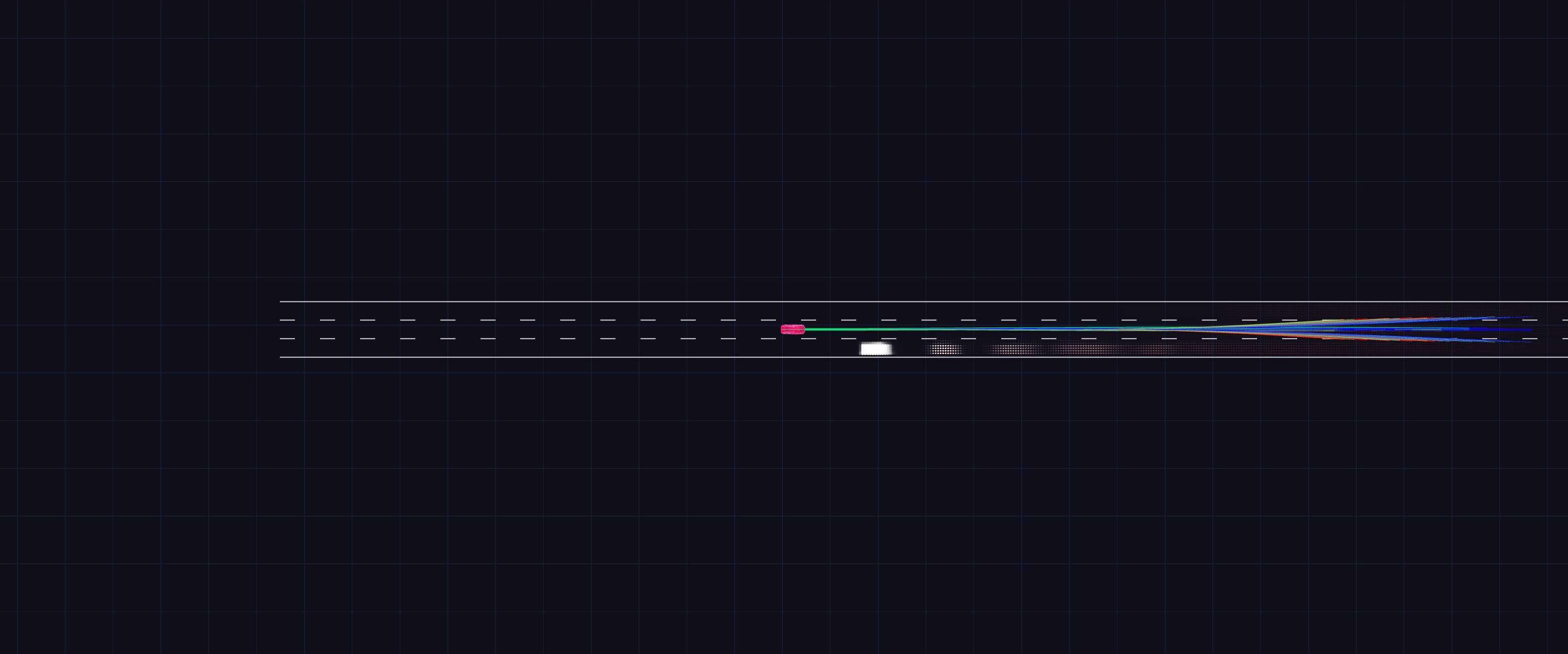} &
        \includegraphics[trim=1100 300 0 300,clip,width=0.6\columnwidth]{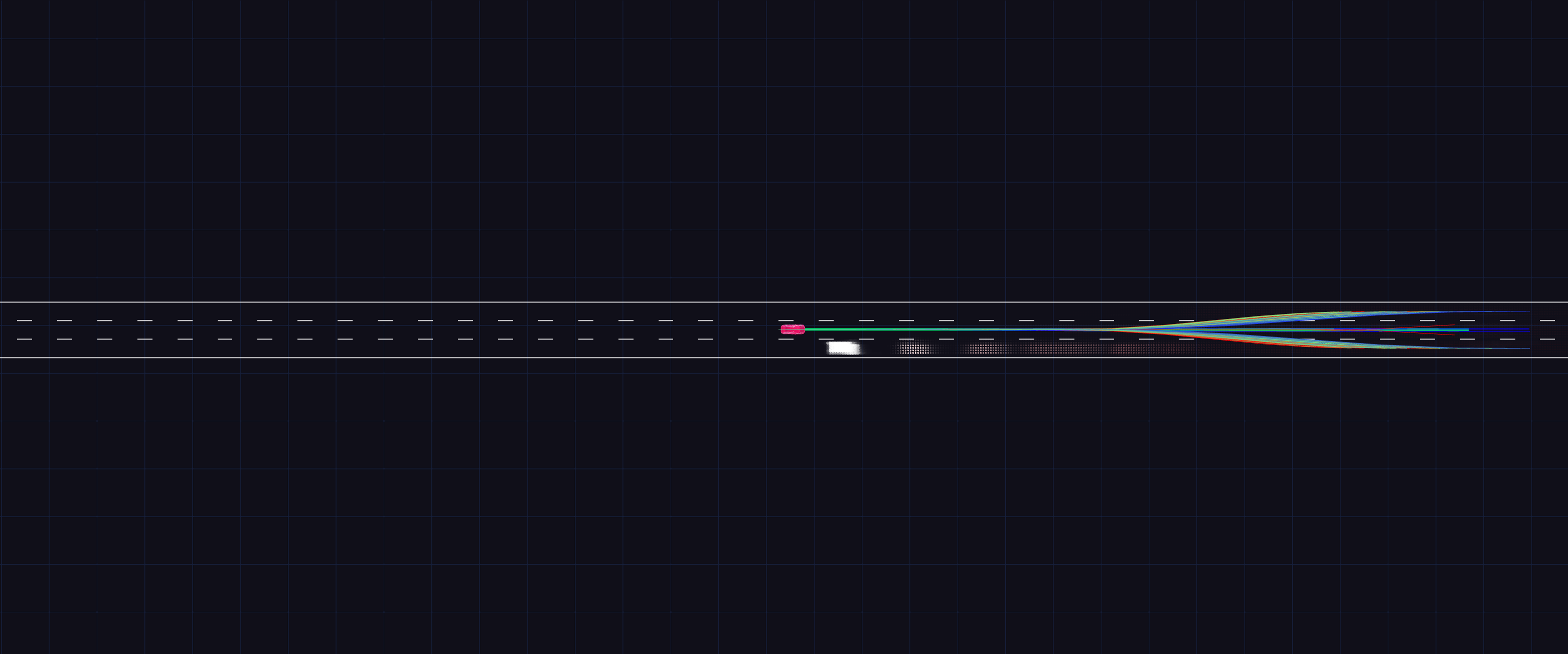} &
        \includegraphics[trim=1100 300 0 300,clip,width=0.6\columnwidth]{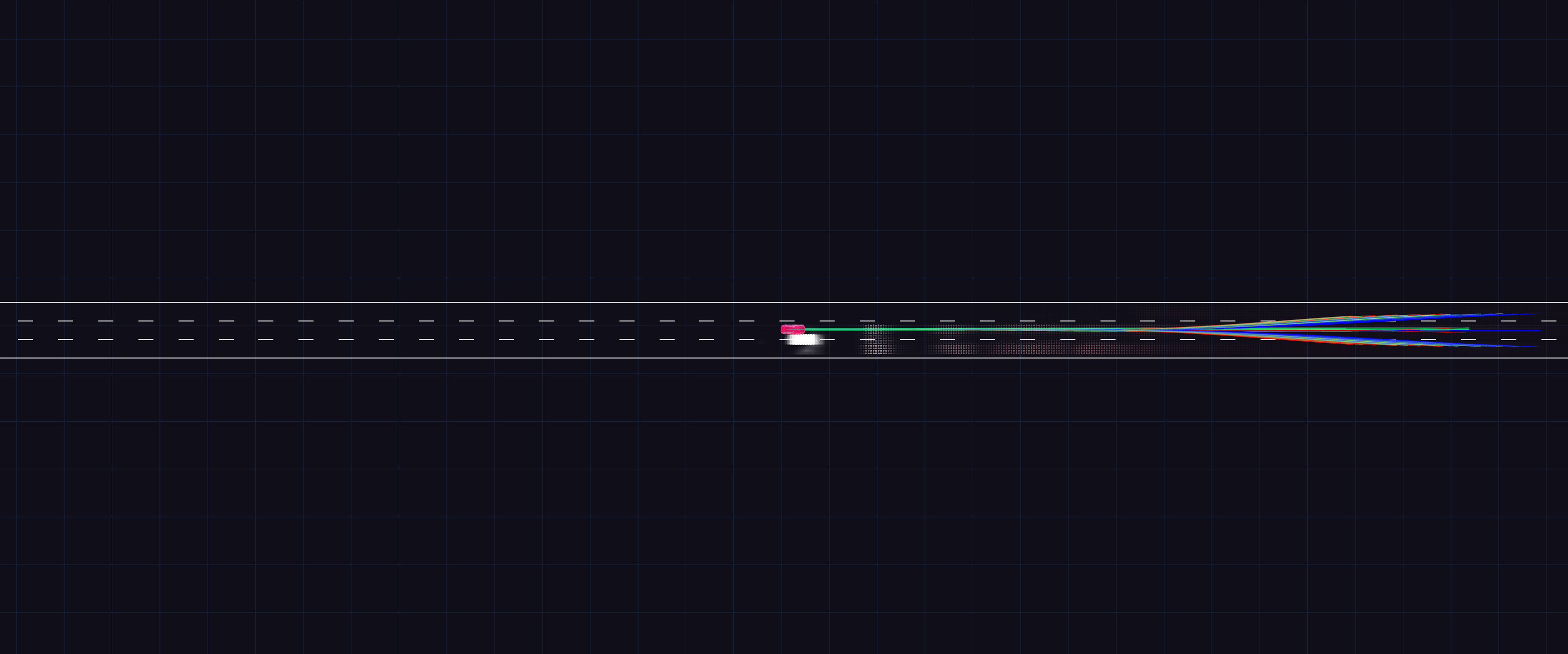} \\

        \raisebox{3.2\height}{\textsc{NMP}} & \includegraphics[trim=1100 300 0 300,clip,width=0.6\columnwidth]{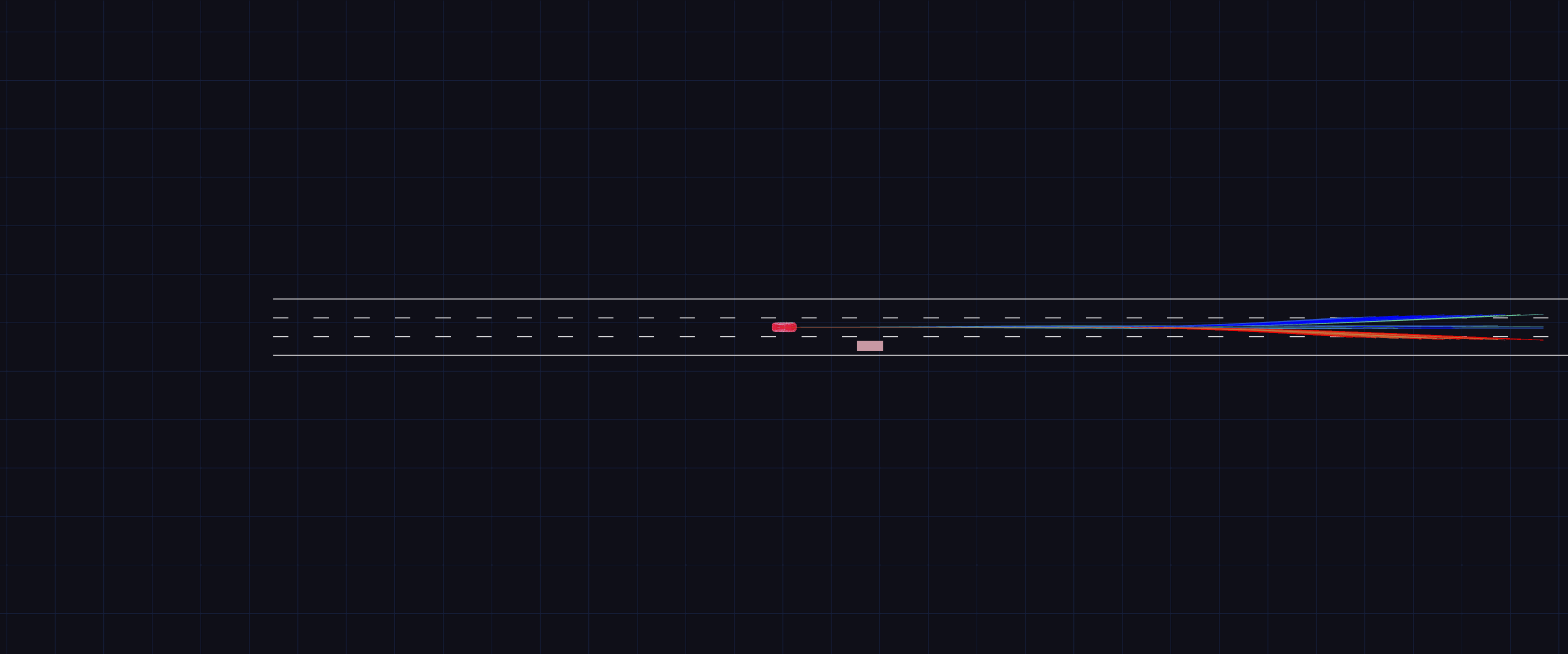} &
        \includegraphics[trim=1100 300 0 300,clip,width=0.6\columnwidth]{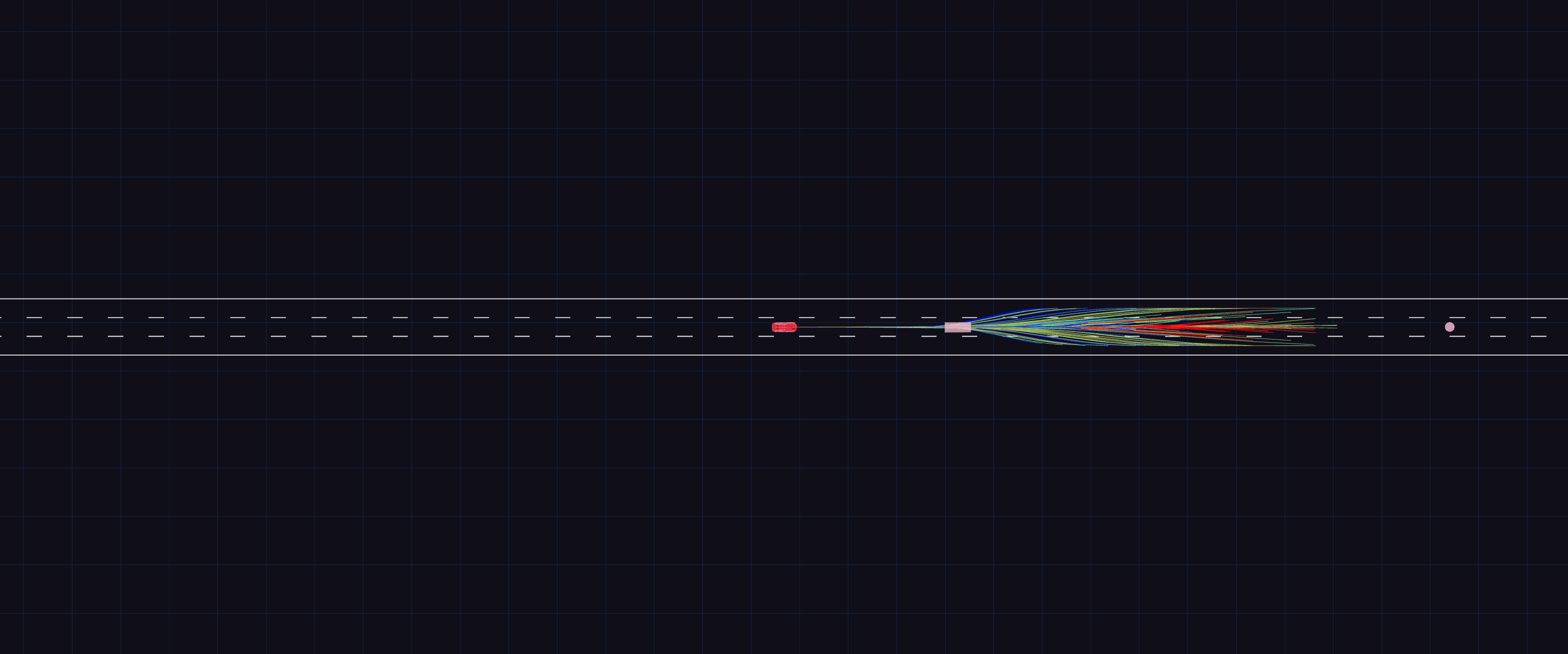} &
        \includegraphics[trim=1100 300 0 300,clip,width=0.6\columnwidth]{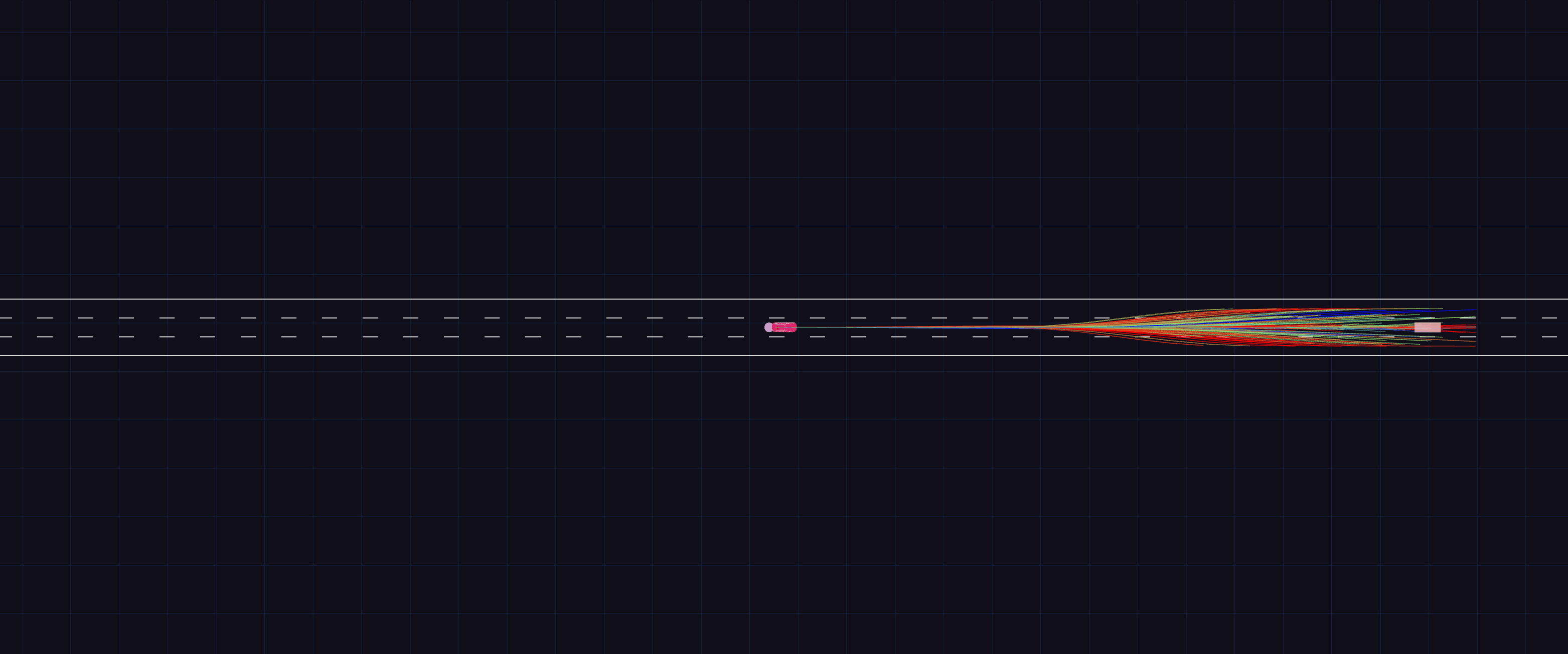} \\

        \raisebox{3.2\height}{\textsc{PlanT}} & \includegraphics[trim=1100 300 0 300,clip,width=0.6\columnwidth]{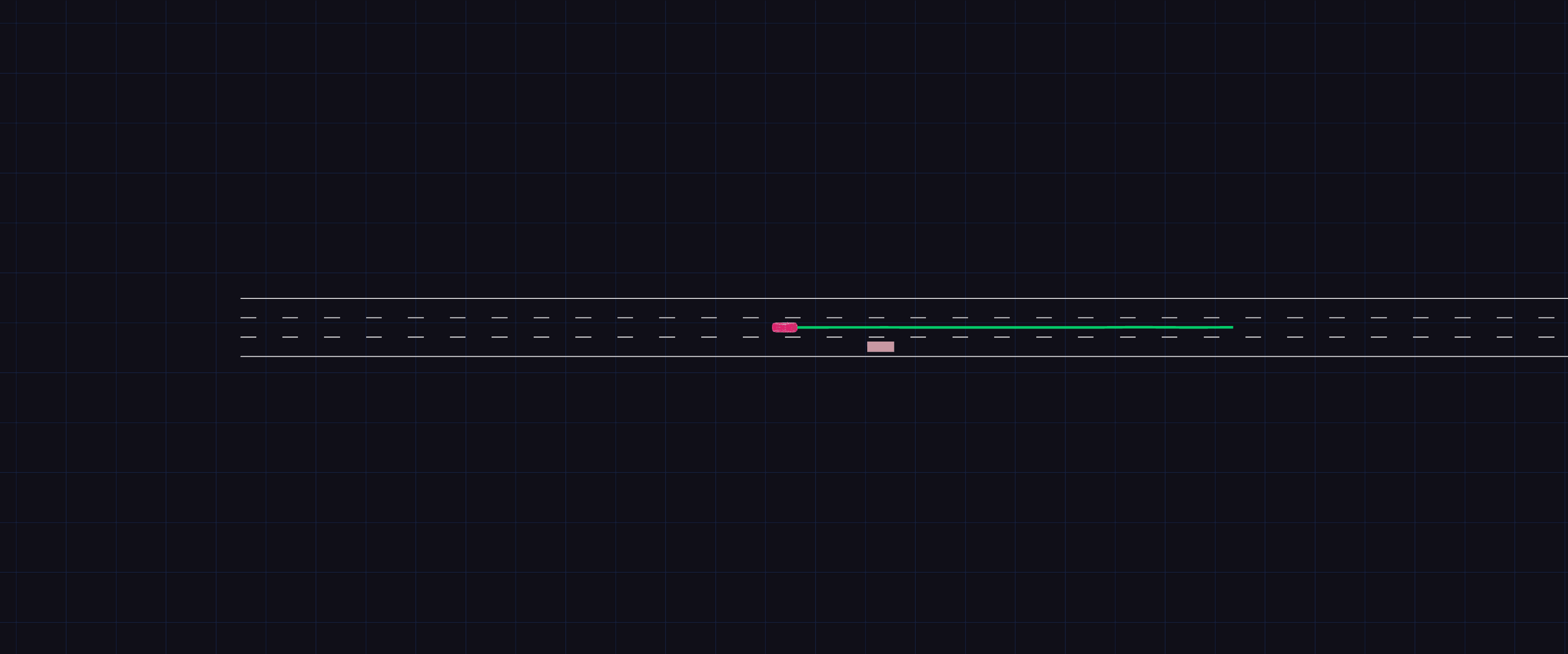} &
        \includegraphics[trim=1100 300 0 300,clip,width=0.6\columnwidth]{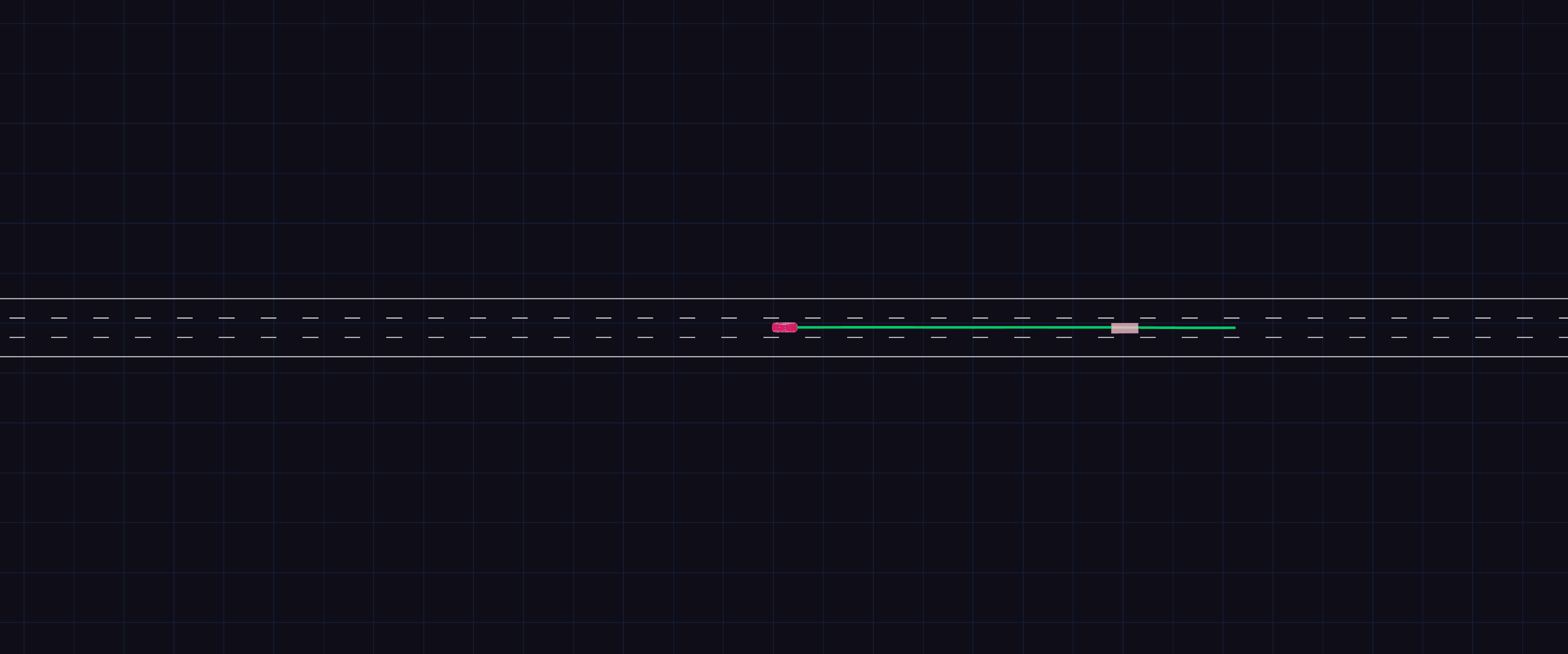} &
        \includegraphics[trim=1100 300 0 300,clip,width=0.6\columnwidth]{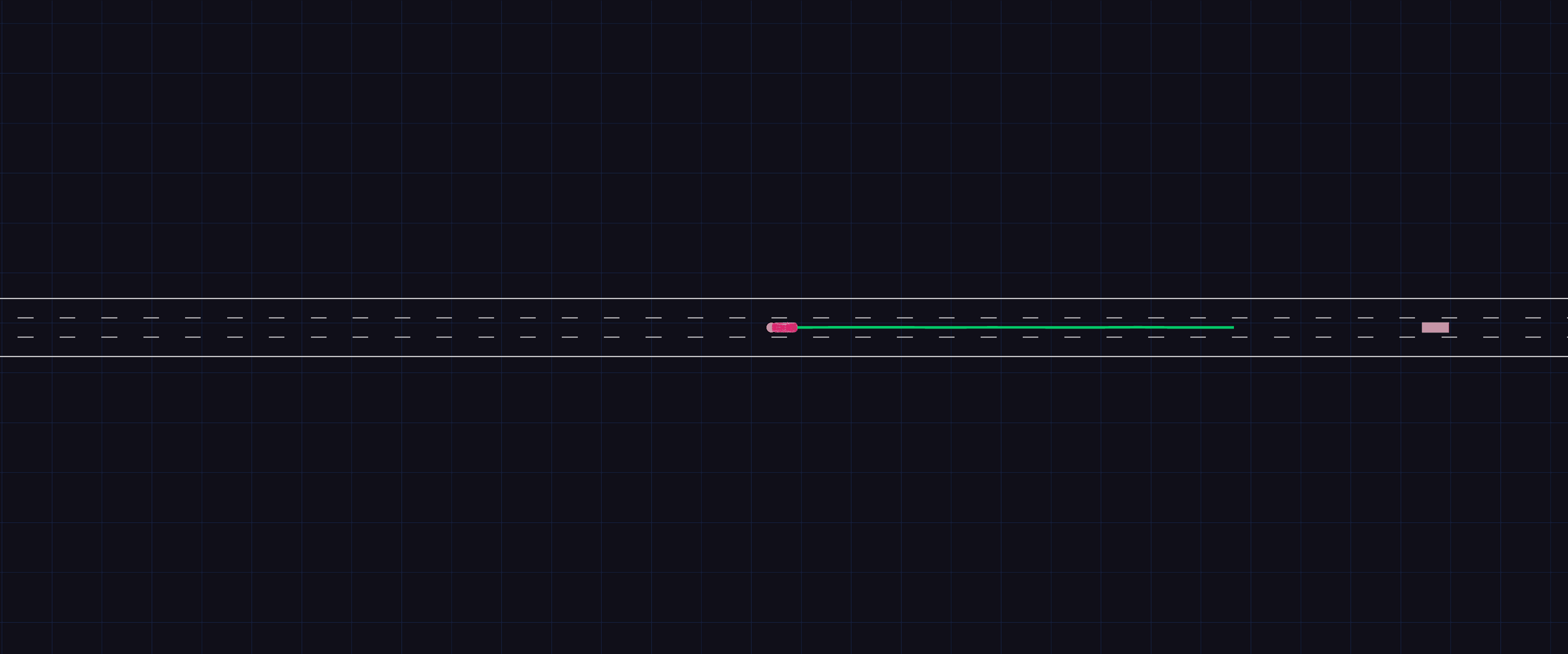} \\
    \end{tabularx}
    \caption{\textbf{Safety Scenario Comparison 1} Here we compare the various models 
    in a scenario where a car cuts in front of the SDV. While our method is able to adapt to the 
    cut-in, the other occupancy models fail to consider that modality and crash into the vehicle. NMP and 
    PlanT are however able to adapt. Note that the red coloring on the trajectories corresponds to 
    a high cost while the blue cost on the trajectories corresponds to a low cost.}
    \vspace{-10pt}
    \label{fig:scene_1_qual}
\end{figure*}

\begin{figure*}[t] %
    \centering
    \begin{tabularx} {\textwidth} {l | X X X} %
        & t = start & t = middle & t = end\\
        \midrule
        \raisebox{3.2\height}{\textsc{$\ourmodel{}$}} & \includegraphics[trim=1100 300 0 300,clip,width=0.6\columnwidth]{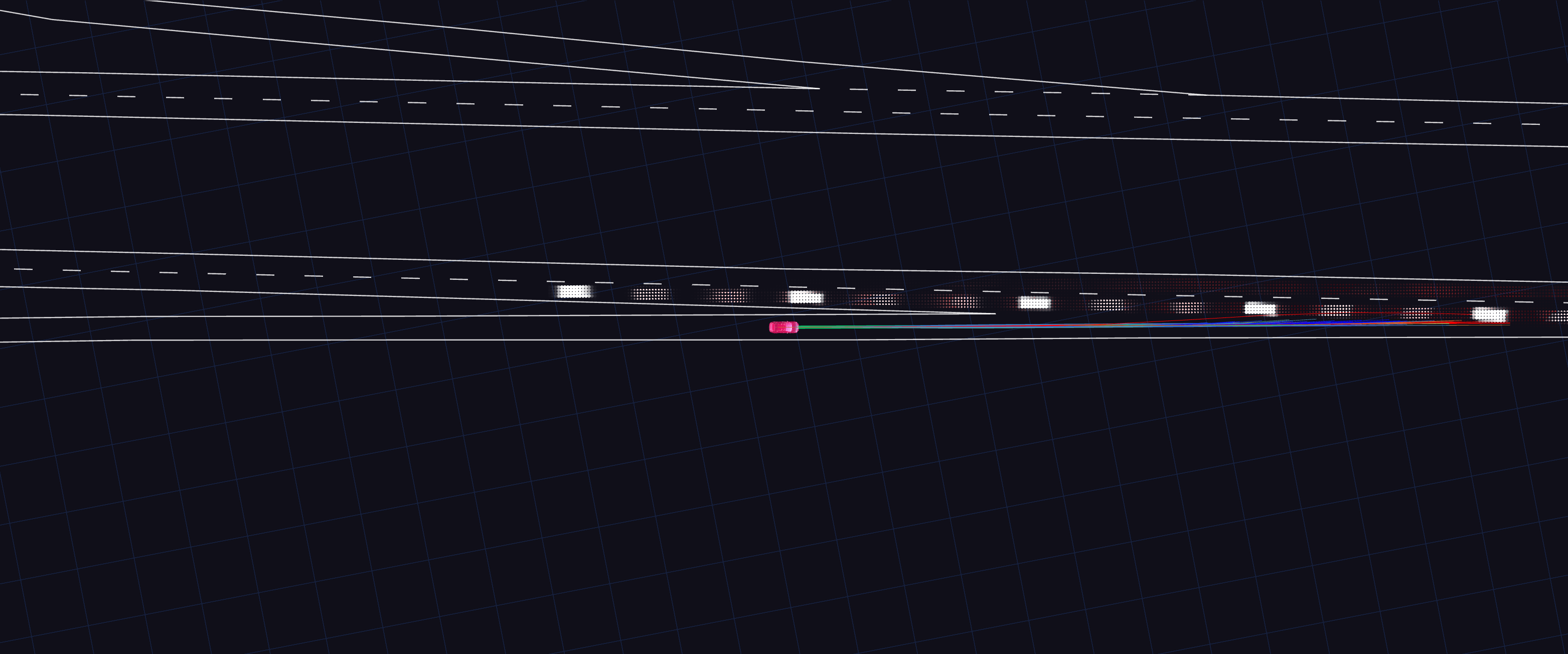} &
        \includegraphics[trim=1100 300 0 300,clip,width=0.6\columnwidth]{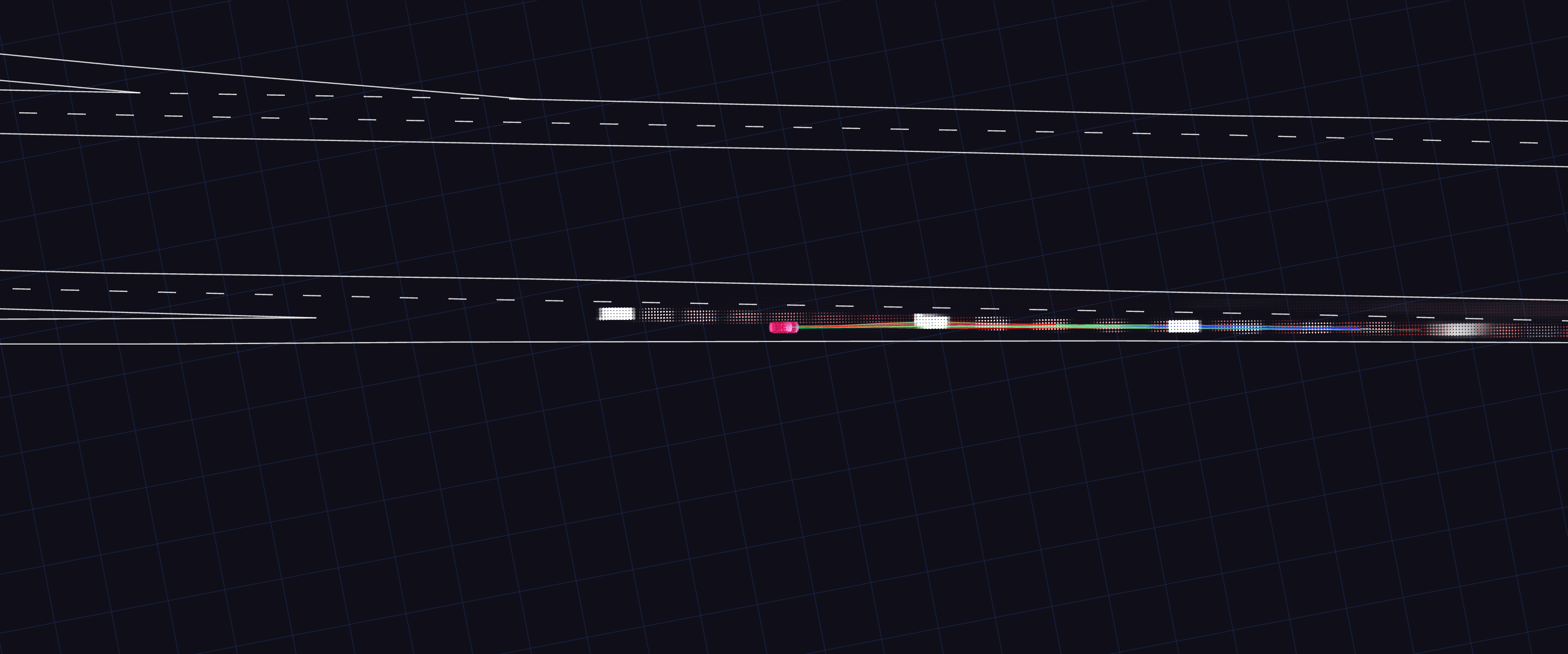} &
        \includegraphics[trim=1100 300 0 300,clip,width=0.6\columnwidth]{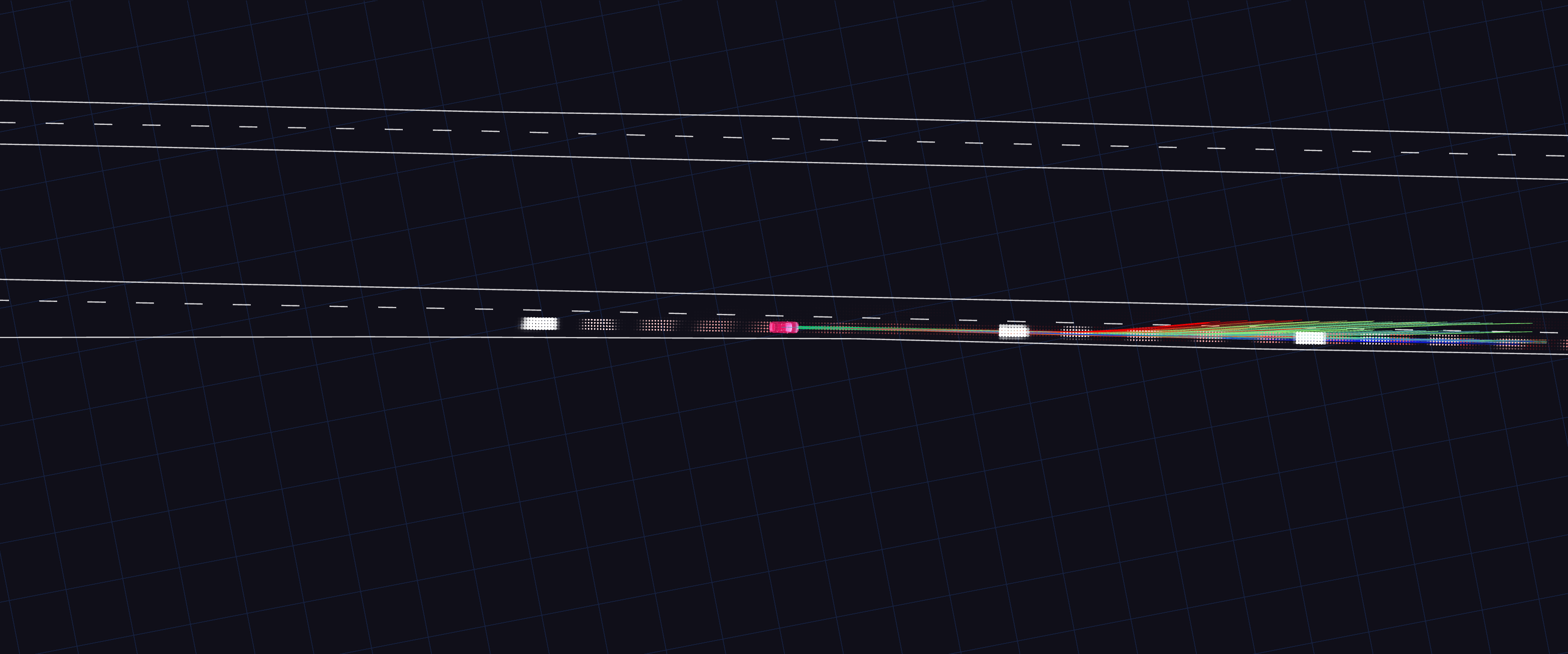} \\

        \raisebox{3.2\height}{\textsc{P3}} & \includegraphics[trim=1100 300 0 300,clip,width=0.6\columnwidth]{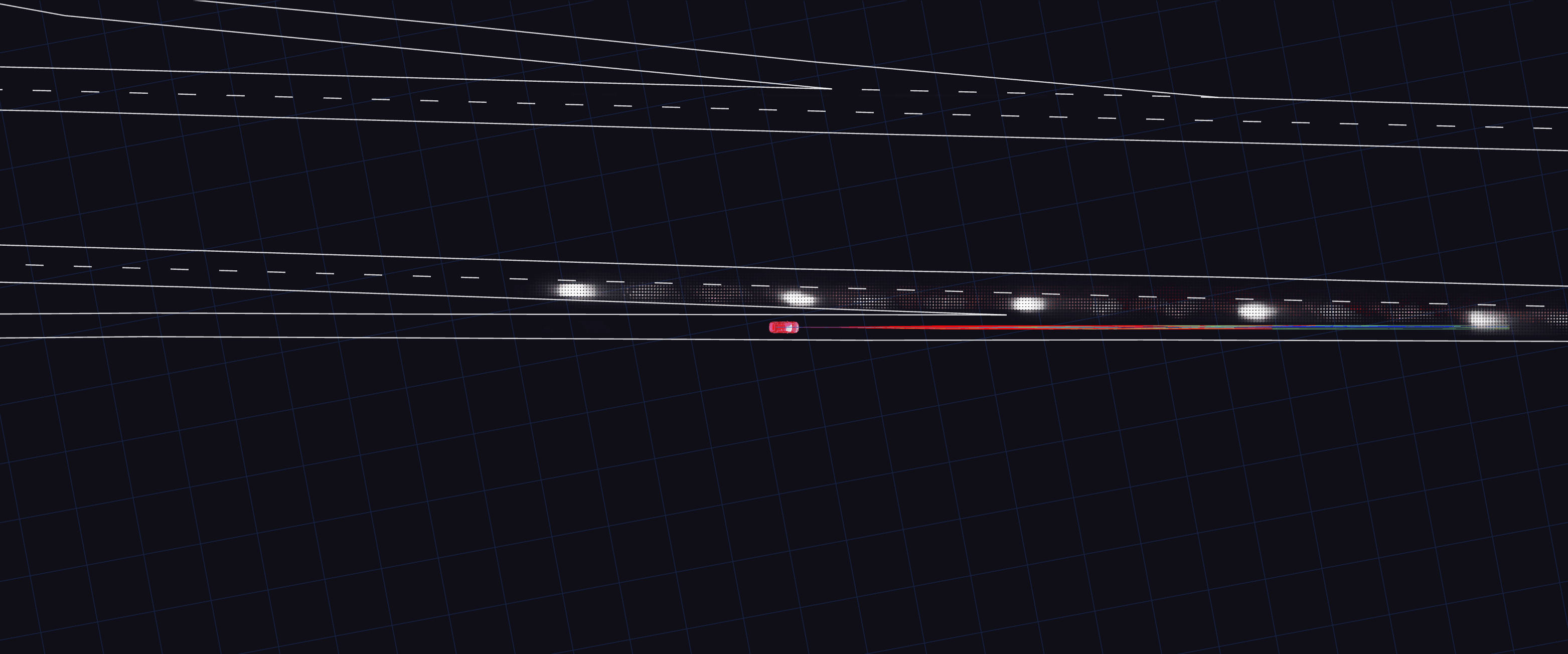} &
        \includegraphics[trim=1100 300 0 300,clip,width=0.6\columnwidth]{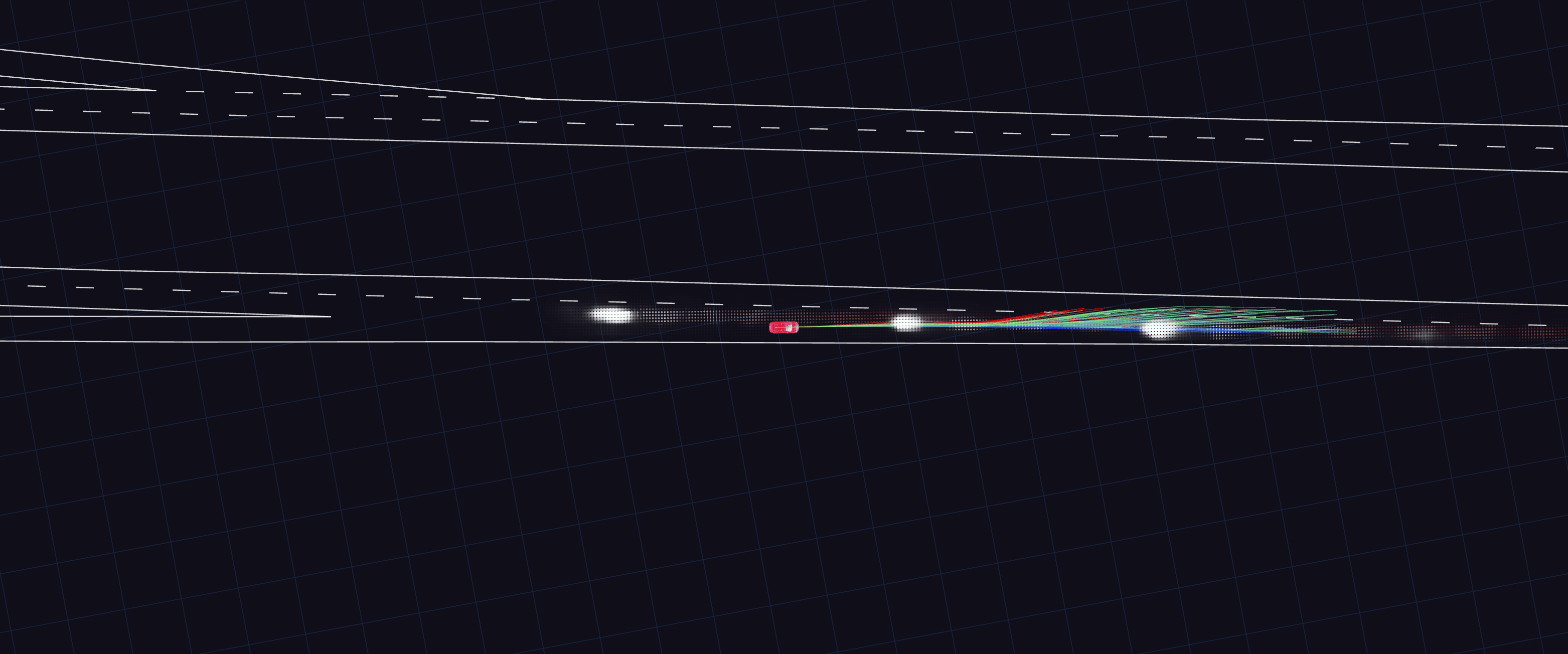} &
        \includegraphics[trim=1100 300 0 300,clip,width=0.6\columnwidth]{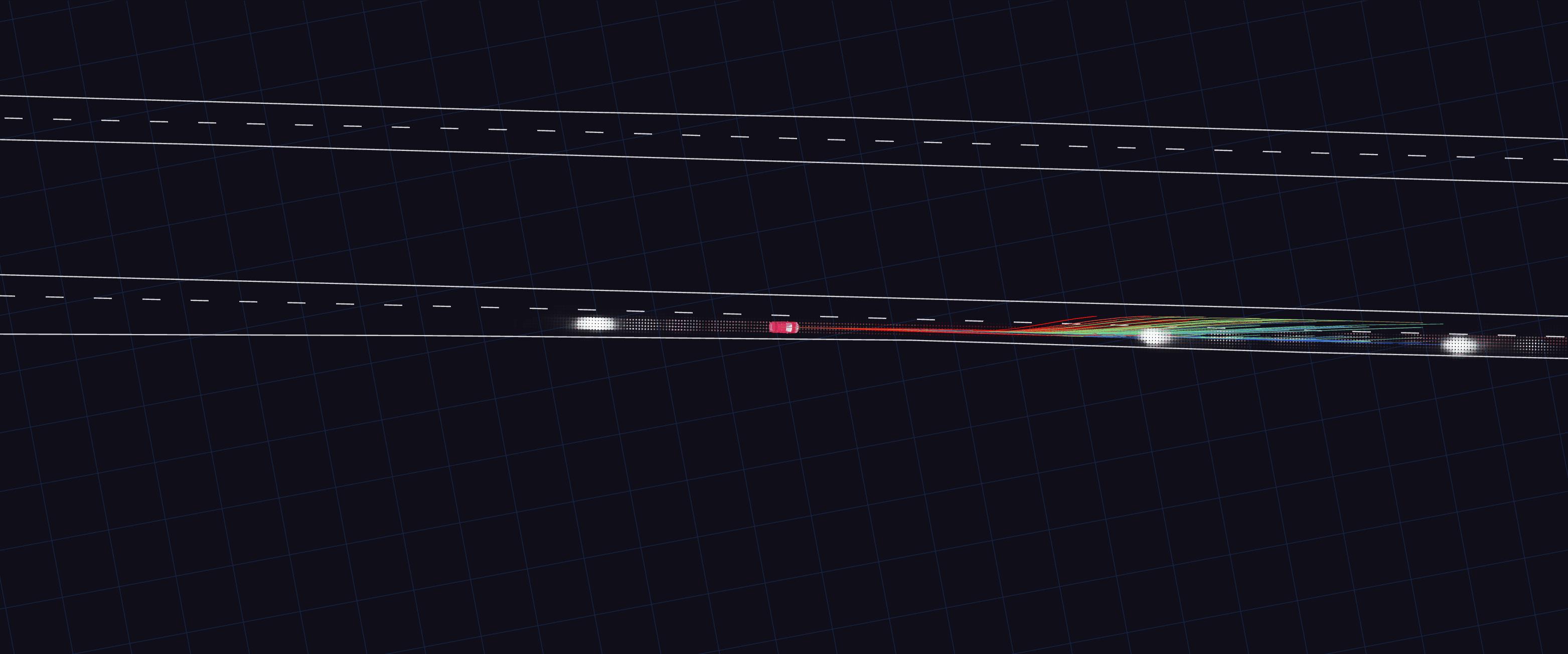} \\

        \raisebox{3.2\height}{\textsc{OccFlow}} & \includegraphics[trim=1100 300 0 300,clip,width=0.6\columnwidth]{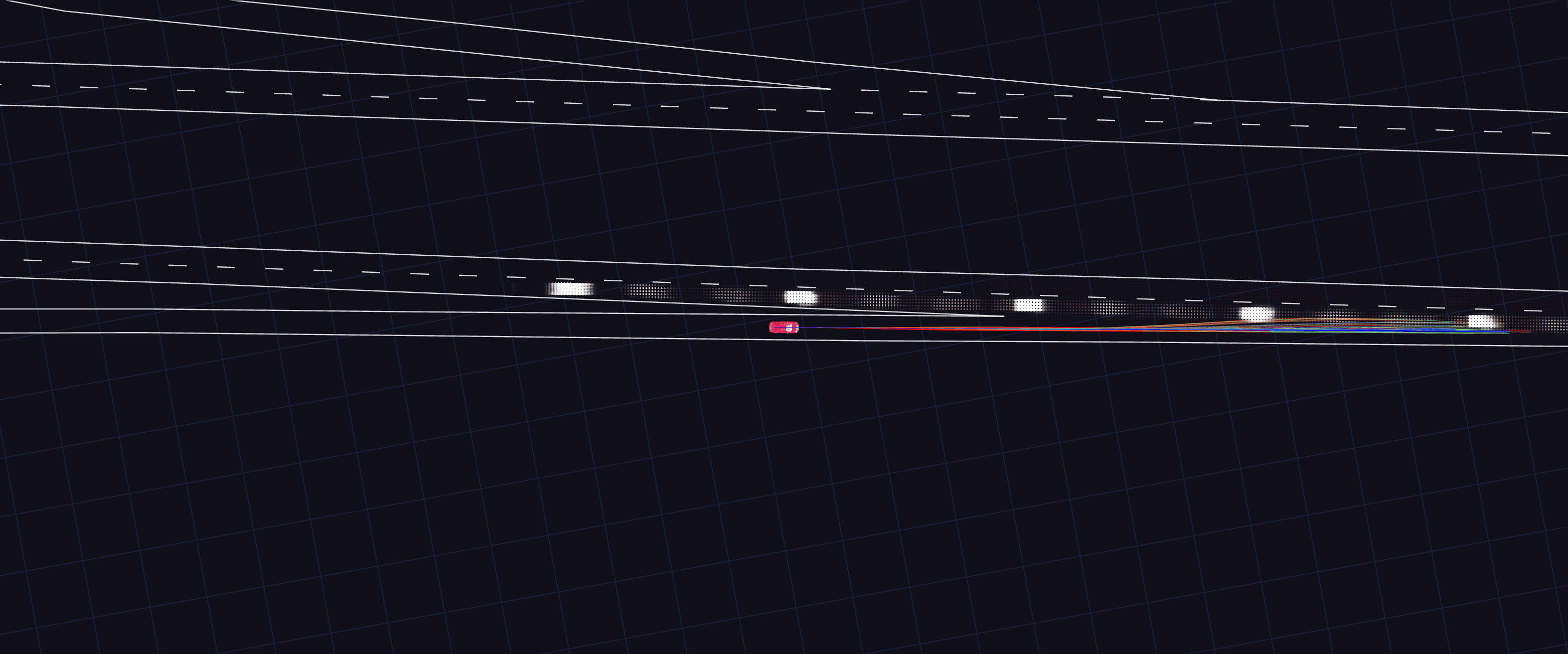} &
        \includegraphics[trim=1100 300 0 300,clip,width=0.6\columnwidth]{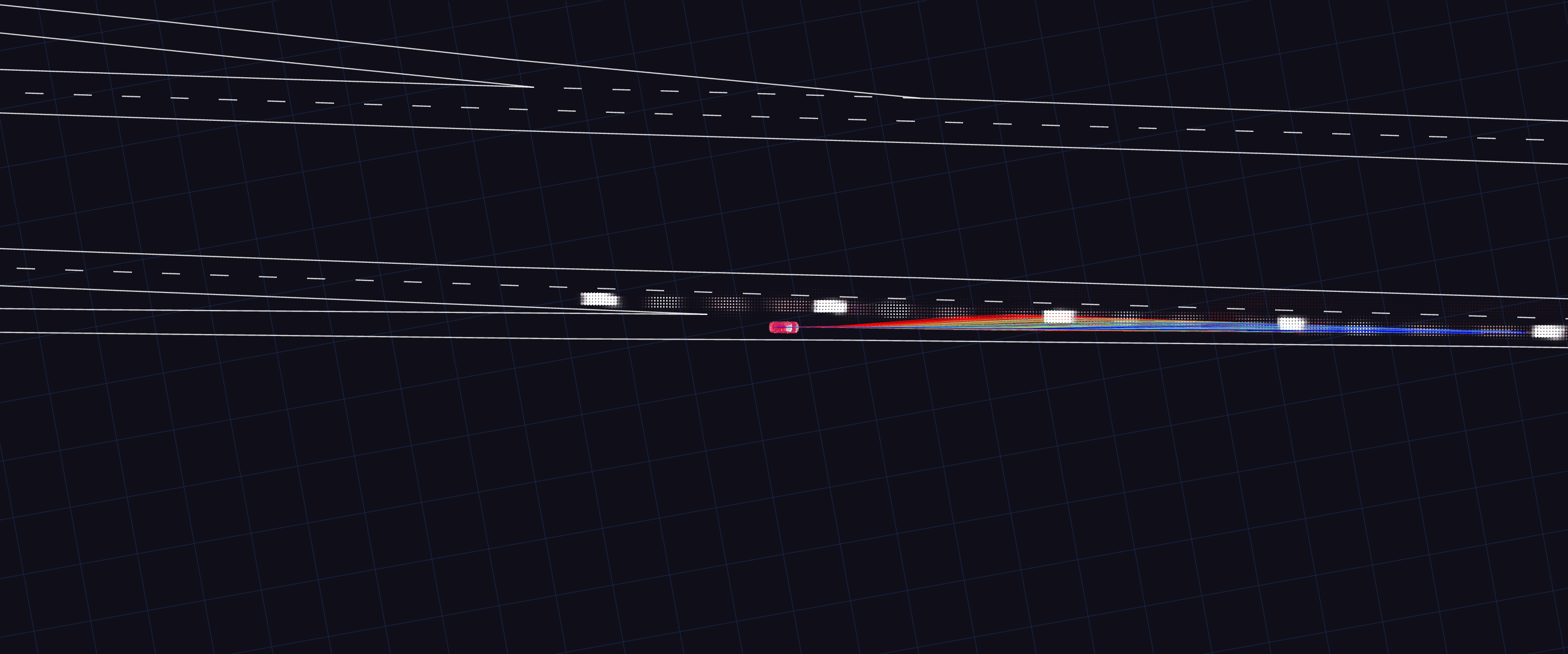} &
        \includegraphics[trim=1100 300 0 300,clip,width=0.6\columnwidth]{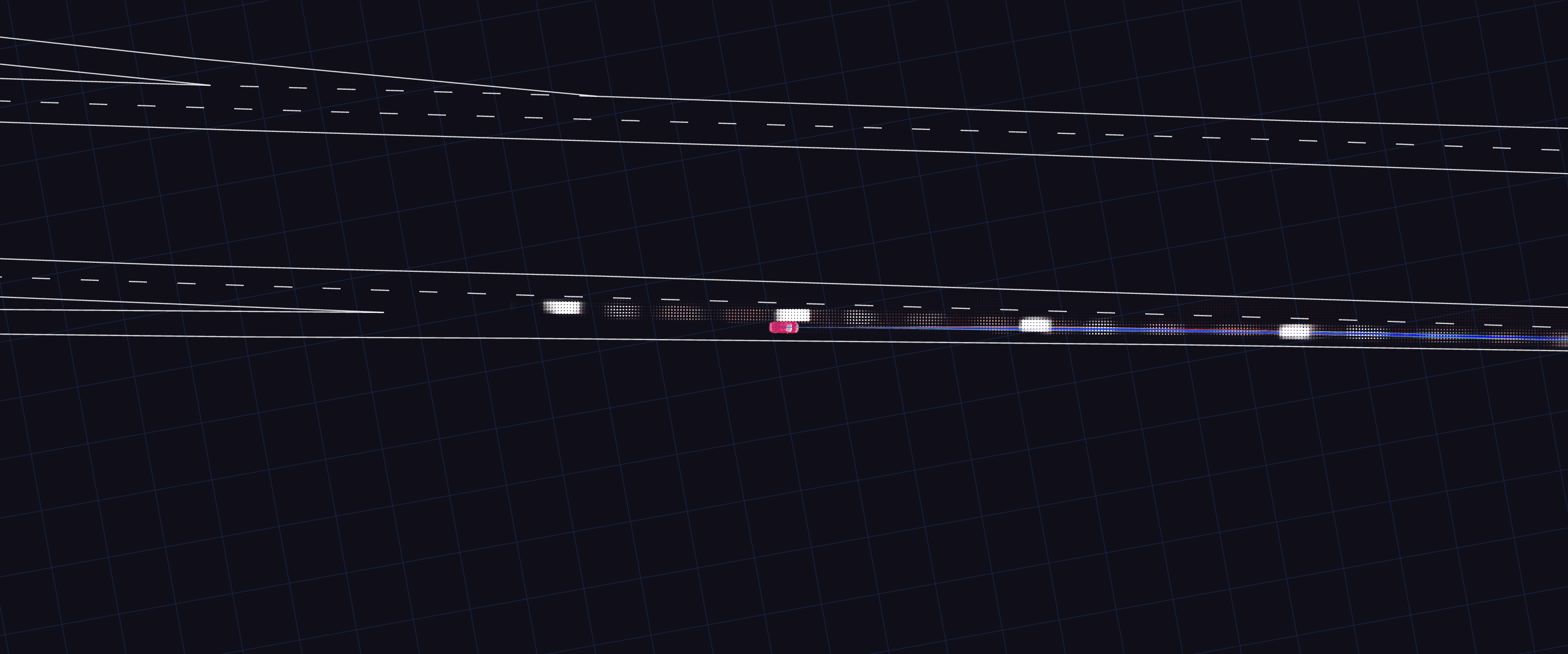} \\

        \raisebox{3.2\height}{\textsc{NMP}} & \includegraphics[trim=1100 300 0 300,clip,width=0.6\columnwidth]{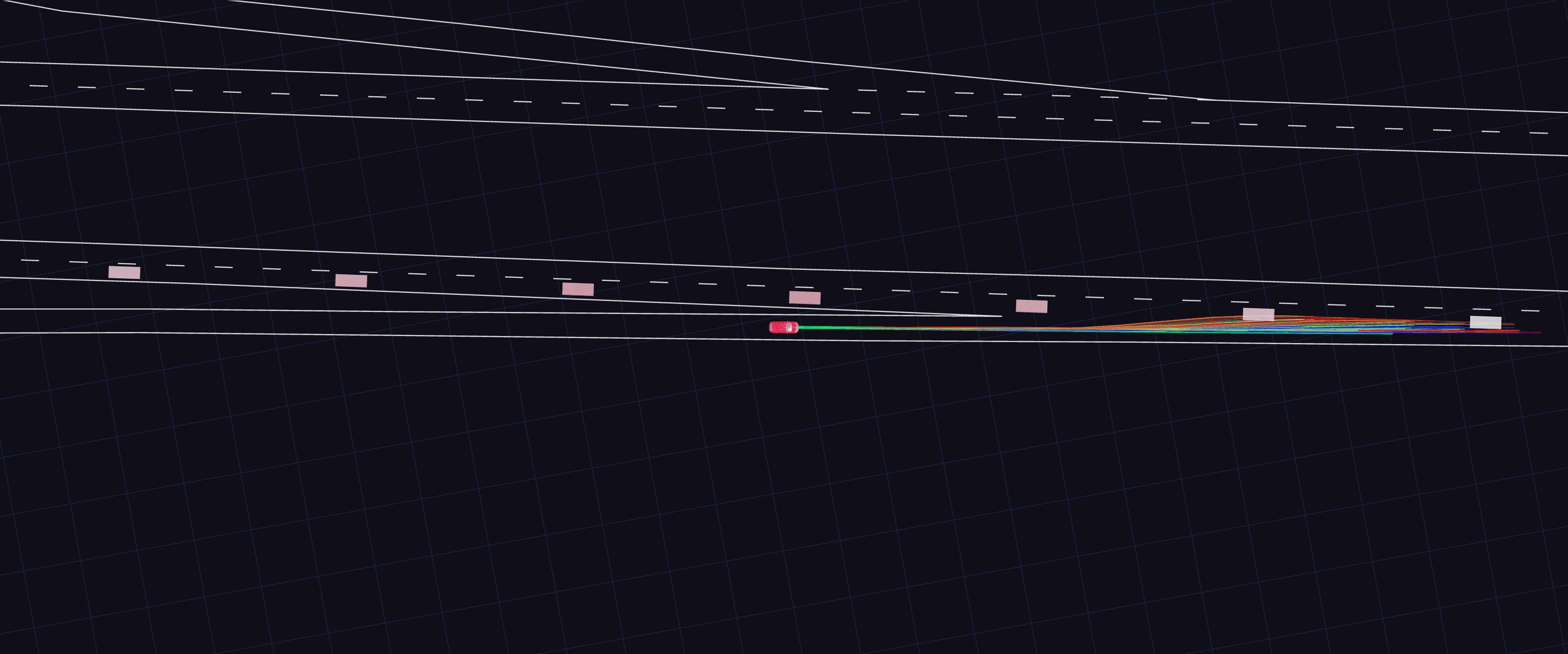} &
        \includegraphics[trim=1100 300 0 300,clip,width=0.6\columnwidth]{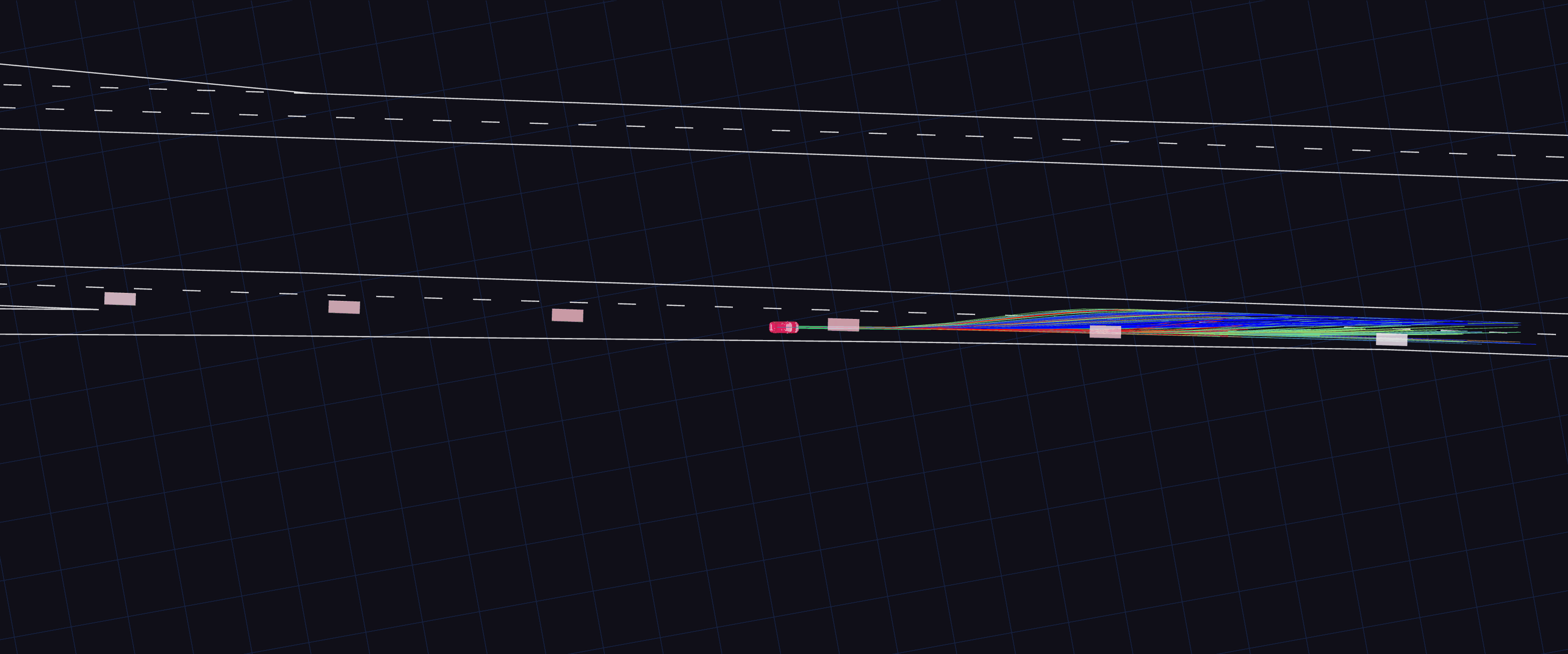} &
        \includegraphics[trim=1100 300 0 300,clip,width=0.6\columnwidth]{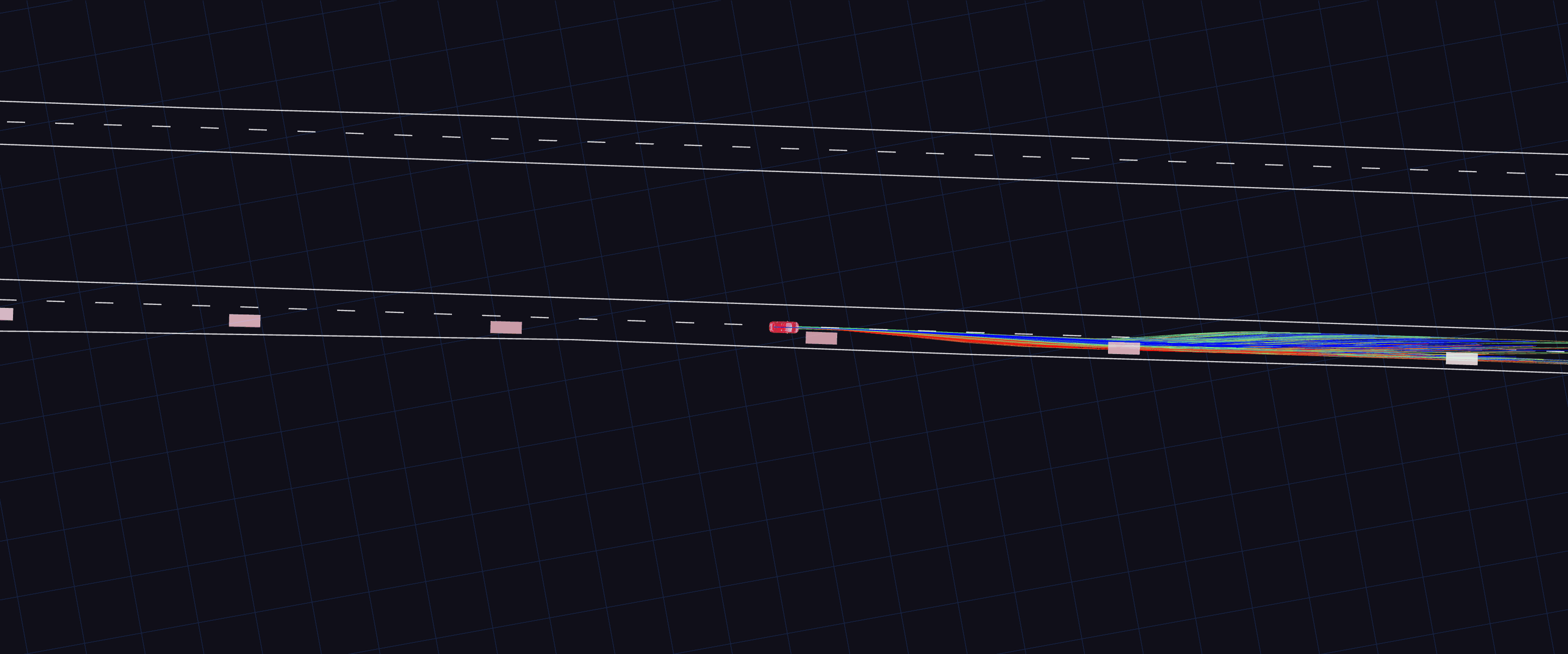} \\

        \raisebox{3.2\height}{\textsc{PlanT}} & \includegraphics[trim=1100 500 0 100,clip,width=0.6\columnwidth]{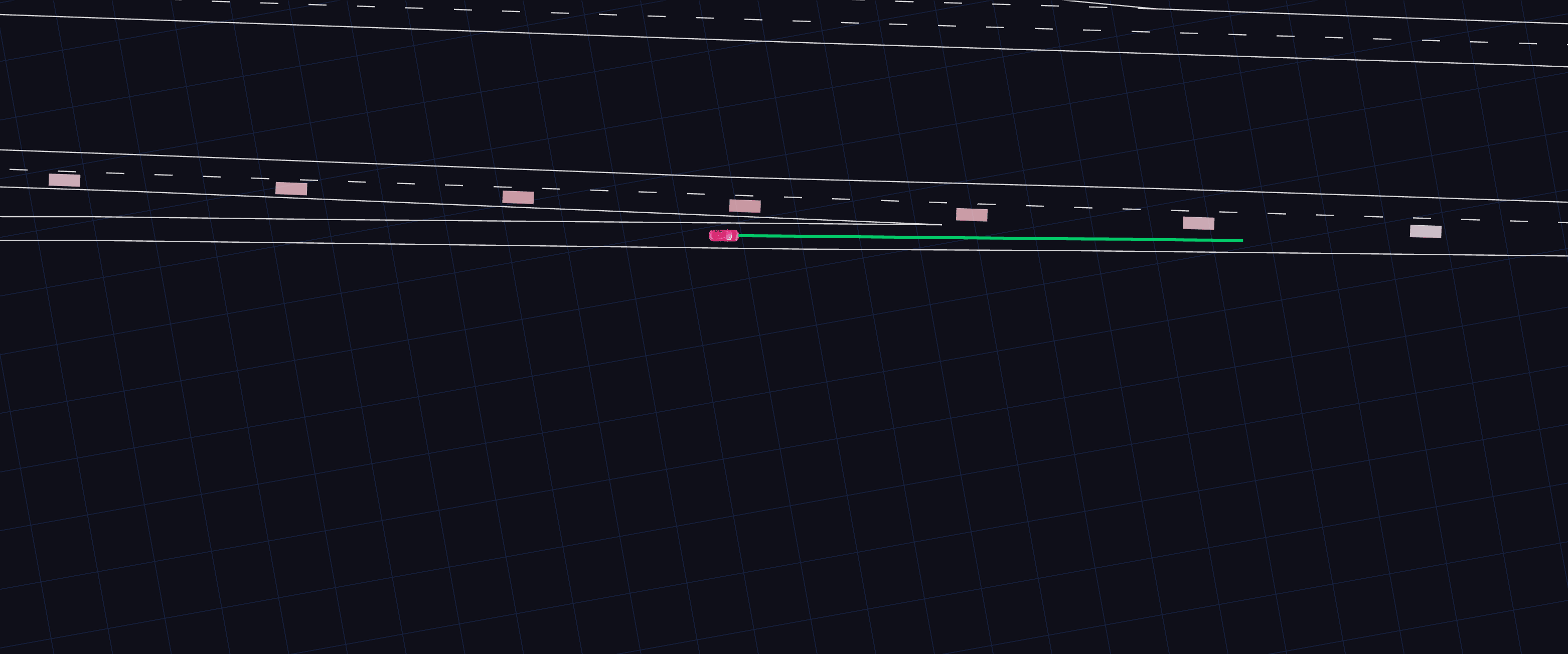} &
        \includegraphics[trim=1100 500 0 100,clip,width=0.6\columnwidth]{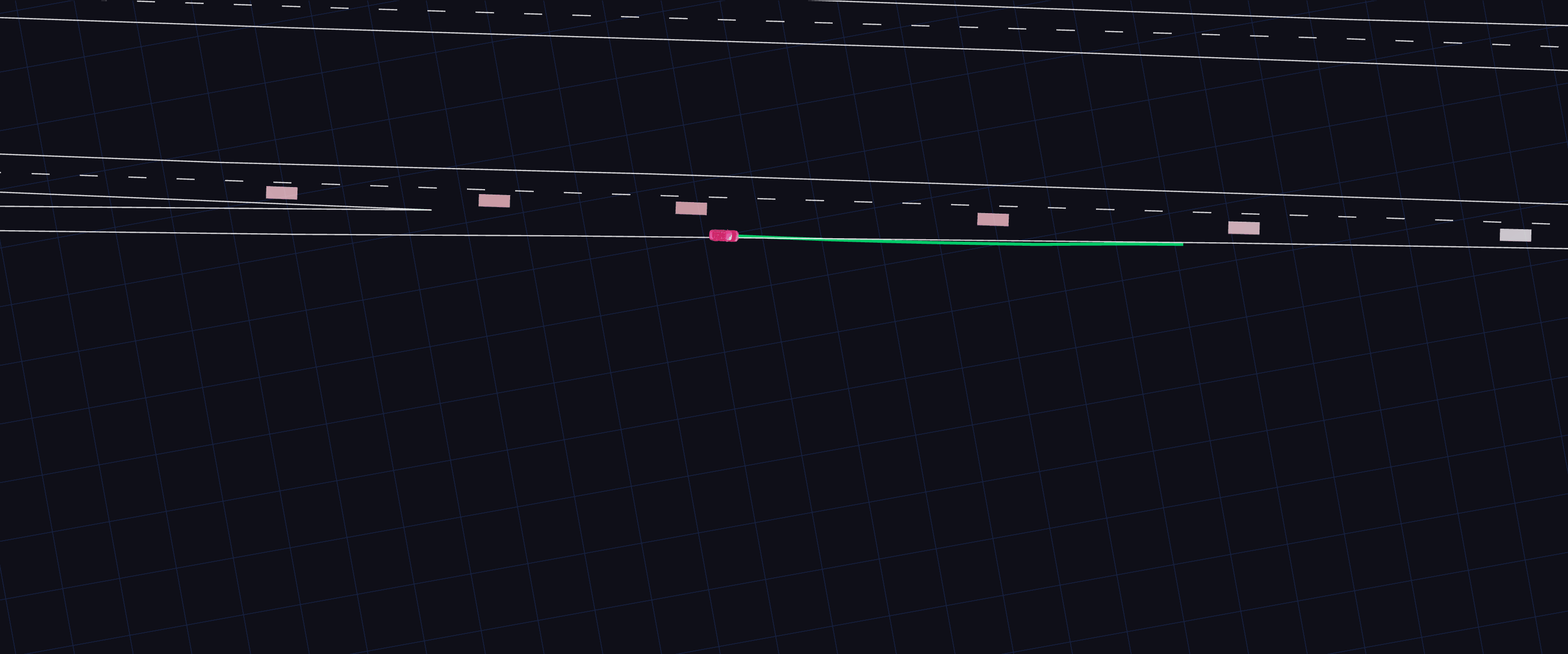} &
        \includegraphics[trim=1100 500 0 100,clip,width=0.6\columnwidth]{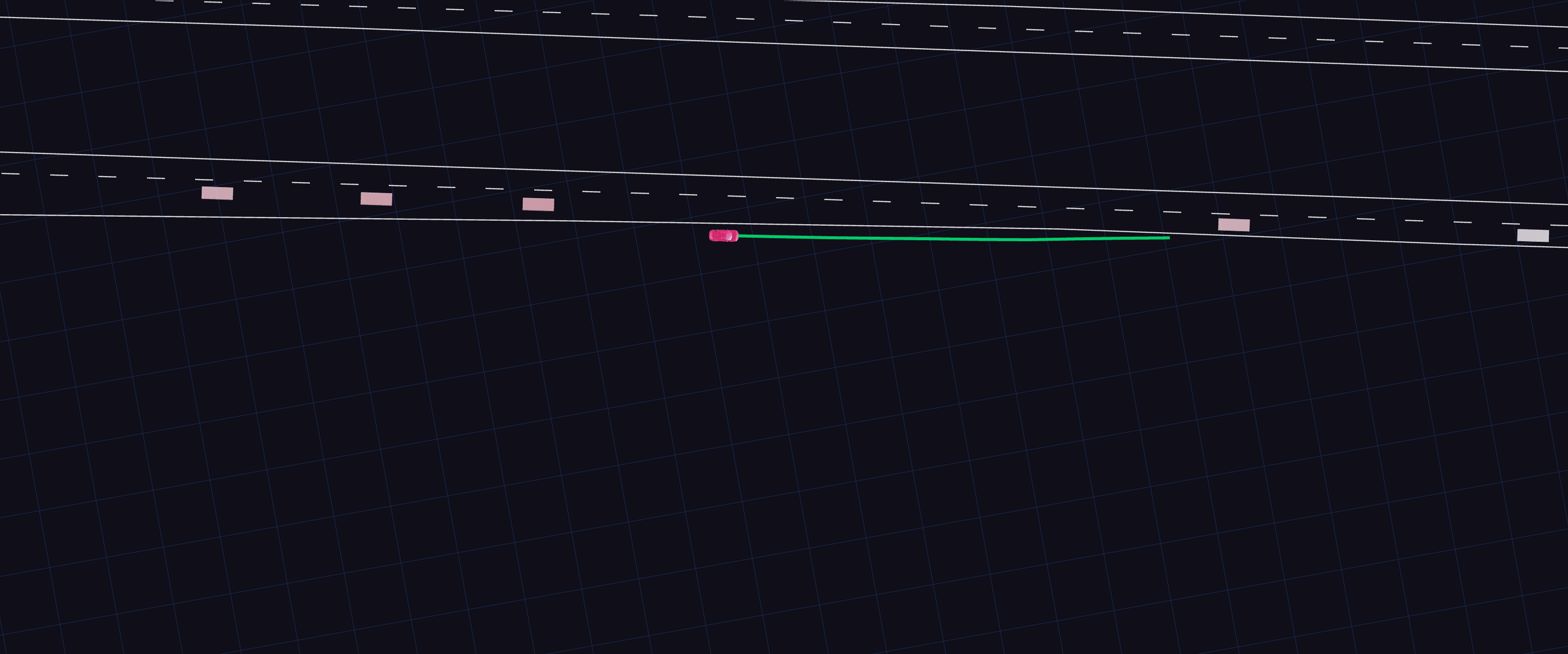} \\
    \end{tabularx}
    \caption{\textbf{Safety Scenario Comparison 2} Here we have a scenario where the vehicle must merge into the 
    desired lane from an on-ramp. Note how while $\ourmodel{}$ and P3 are both able to navigate into the traffic, 
    OccFlow fails to adequately rank the trajectories resulting in a collision. PlanT suffers from out of 
    distribution maps and therefore begins drifting off the road, and finally NMP acts aggressively 
    resulting in another collision.}
    \vspace{-10pt}
    \label{fig:scene_2_qual}
\end{figure*}

\end{document}